\documentclass[a4paper,fleqn]{cas-sc}

\usepackage[authoryear]{natbib}
\usepackage{enumitem}
\usepackage{algorithm}
\usepackage{algpseudocode}

\def\tsc#1{\csdef{#1}{\textsc{\lowercase{#1}}\xspace}}
\tsc{WGM}
\tsc{QE}
\begin{document}
\let\WriteBookmarks\relax
\def\floatpagepagefraction{1}
\def\textpagefraction{.001}

\shorttitle{\textsc{AutoSurrogate}}    

\shortauthors{Liu and Wang}  

\title [mode = title]{\textsc{AutoSurrogate}: An LLM-Driven Multi-Agent Framework for Autonomous Construction of Deep Learning Surrogate Models in Subsurface Flow}  


%\tnotetext[1]{This work is supported by xxx} 


\author[1]{Jiale Liu}[orcid=0009-0007-5881-3118]


\ead{Jiale.Liu@ed.ac.uk}

\affiliation[1]{organization={School of Physics and Astronomy, The University of Edinburgh},
                city={Edinburgh},
                postcode={EH9 3FD}, 
                country={United Kingdom}}

\author[2,3]{Nanzhe Wang}[orcid=0000-0002-5177-946X]
% % Footnote of the second author

\cormark[1]

% Email id of the second author
\ead{Nanzhe.Wang@hw.ac.uk}


\affiliation[2]{organization={Institute of GeoEnergy Engineering, School of Energy, Geoscience, Infrastructure and Society, Heriot-Watt University},
                city={Edinburgh},
                postcode={EH14 4AS}, 
                country={United Kingdom}}
\affiliation[3]{organization={Subsurface Energy Transition and Innovation Centre, Heriot-Watt University},
                city={Edinburgh},
                postcode={EH14 4AS}, 
                country={United Kingdom}}

\cortext[1]{Corresponding author}

% % Footnote text


\begin{abstract}
High-fidelity numerical simulation of subsurface flow is computationally intensive, especially for many-query tasks such as uncertainty quantification and data assimilation. Deep learning (DL) surrogates can significantly accelerate forward simulations, yet constructing them requires substantial machine learning (ML) expertise -- from architecture design to hyperparameter tuning -- that most domain scientists do not possess. Furthermore, the process is predominantly manual and relies heavily on heuristic choices. This expertise gap remains a key barrier to the broader adoption of DL surrogate techniques. For this reason, we present \textsc{AutoSurrogate}, a large-language-model-driven multi-agent framework that enables practitioners without ML expertise to build high-quality surrogates for subsurface flow problems through natural-language instructions. Given simulation data and optional preferences, four specialized agents collaboratively execute data profiling, architecture selection from a model zoo, Bayesian hyperparameter optimization, model training, and quality assessment against user-specified thresholds. The system also handles common failure modes autonomously, including restarting training with adjusted configurations when numerical instabilities occur and switching to alternative architectures when predictive accuracy falls short of targets. In our setting, a single natural-language sentence can be sufficient to produce a deployment-ready surrogate model, with minimum human intervention required at any intermediate stage. We demonstrate the utility of \textsc{AutoSurrogate} on a 3D geological carbon storage modeling task, mapping permeability fields to pressure and CO$_2$ saturation fields over 31 timesteps. Without any manual tuning, \textsc{AutoSurrogate} is able to outperform expert-designed baselines and domain-agnostic AutoML methods, demonstrating strong potential for practical deployment.
\end{abstract}


% % Research highlights
% \begin{highlights}
% \item 
% \item 
% \item 
% \end{highlights}

% Keywords

\maketitle

% Main text
\section{Introduction}
% background and our goal
Subsurface flow modeling is critical for the success of various subsurface energy projects, such as geological carbon storage, geothermal energy extraction, and underground hydrogen storage~\citep{boot2014carbon,rohit2023tracing,hellerschmied2024hydrogen}. These subsurface flow and energy systems inherently involve large uncertainty, multiple physical processes, and massive spatial scales, which make the high-fidelity numerical simulation of these systems highly computationally expensive and time-consuming. This issue would be magnified for tasks like uncertainty quantification, data assimilation, and production optimization, which require a large number of forward evaluations~\citep{babaei2015robust,tangDeeplearningbasedSurrogateFlow2021,wang2025deep,han2026recurrent}. Surrogate models offer an effective solution to bypass this computational bottleneck by approximating the high-fidelity simulation with efficient representations and offering orders-of-magnitude speedups. Deep learning (DL) surrogate models have emerged as a promising choice due to their powerful capabilities to handle high dimensional and nonlinear problems. Our goal in this work is to design an autonomous deep learning surrogate modeling framework (\textsc{AutoSurrogate}) by exploiting the power of large language models (LLMs) and AI agents. \textsc{AutoSurrogate} can reduce the technical barriers of deep learning surrogate modeling and enable domain scientists to easily construct, train, and deploy high-quality surrogates for subsurface flow through natural-language instructions. 

% Surrogate models
Various surrogate model methods have been widely utilized to speedup the numerical simulation, including intrusive and non-intrusive methods~\citep{babaei2015robust,asher2015review}. Intrusive methods, such as Galerkin projection~\citep{tadmor2010mean, hesthaven2016certified,semaan2016reduced}, trajectory piecewise linearization (TPWL)~\citep{gao2025reduced} require direct modification of the governing equations and access to the underlying numerical solver, making them difficult to implement in practice, particularly for complex subsurface flow systems. Non-intrusive methods, such as Gaussian process regression~\citep{gadd2019surrogate, allgeier2023surrogate}, non-intrusive polynomial chaos expansion~\citep{elsheikh2014efficient,meng2018uncertainty}, radial basis functions~\citep{buhmann2000radial}, support vector machine~\citep{he2019data}, learn input–output mappings directly from data, offering greater flexibility and ease of implementation without requiring access to the underlying solver. However, these methods typically suffer from the curse of dimensionality, with computational cost and model complexity growing rapidly as the input dimensionality increases, limiting their applicability to high-dimensional subsurface systems.

% deep learning surrogates
DL surrogates can handle high-dimensional problems more effectively and has been widely investigated for subsurface flow modeling in recent years. Various network architectures have been employed for surrogate modeling, including Encoder-decoder architecture~\citep{mo2019deep,wang2021theory}, hybrid convolution recurrent network~\citep{ma2022efficient,wang2023deep}, recurrent Residual U-Net~\citep{tang2020deep}, Transformer~\citep{fu2025deep,han2026recurrent}, Fourier neural operators (FNO)~\citep{wen2022u,pawar2026accelerated}, graph neural networks (GNN)~\citep{tang2024graph,tang2025graph}, etc. Moreover, the domain knowledge of subsurface flow can also be incorporated into the training processes of DL surrogates to improve the model performance and reduce the data volume dependency, such as physics-informed or theory-guided DL surrogates~\citep{zhu2019physics,karumuri2020simulator,wang2021efficient,wang2022surrogate}. DL surrogates have also been utilized for different tasks in subsurface flow, including data assimilation~\citep{tang2020deep,ma2022efficient,wang2025deep}, uncertainty quantification~\citep{mo2019deep,wang2021efficient,karumuri2020simulator}, and optimization~\citep{tang2024graph,wang2023deep}, as well as other areas~\citep{zhuDigitalTwinSurrogate2026, wirthDatadrivenSurrogateMaterial2026}.

\begin{figure}[pos=htbp]
    \centering
    \includegraphics[width=\linewidth]{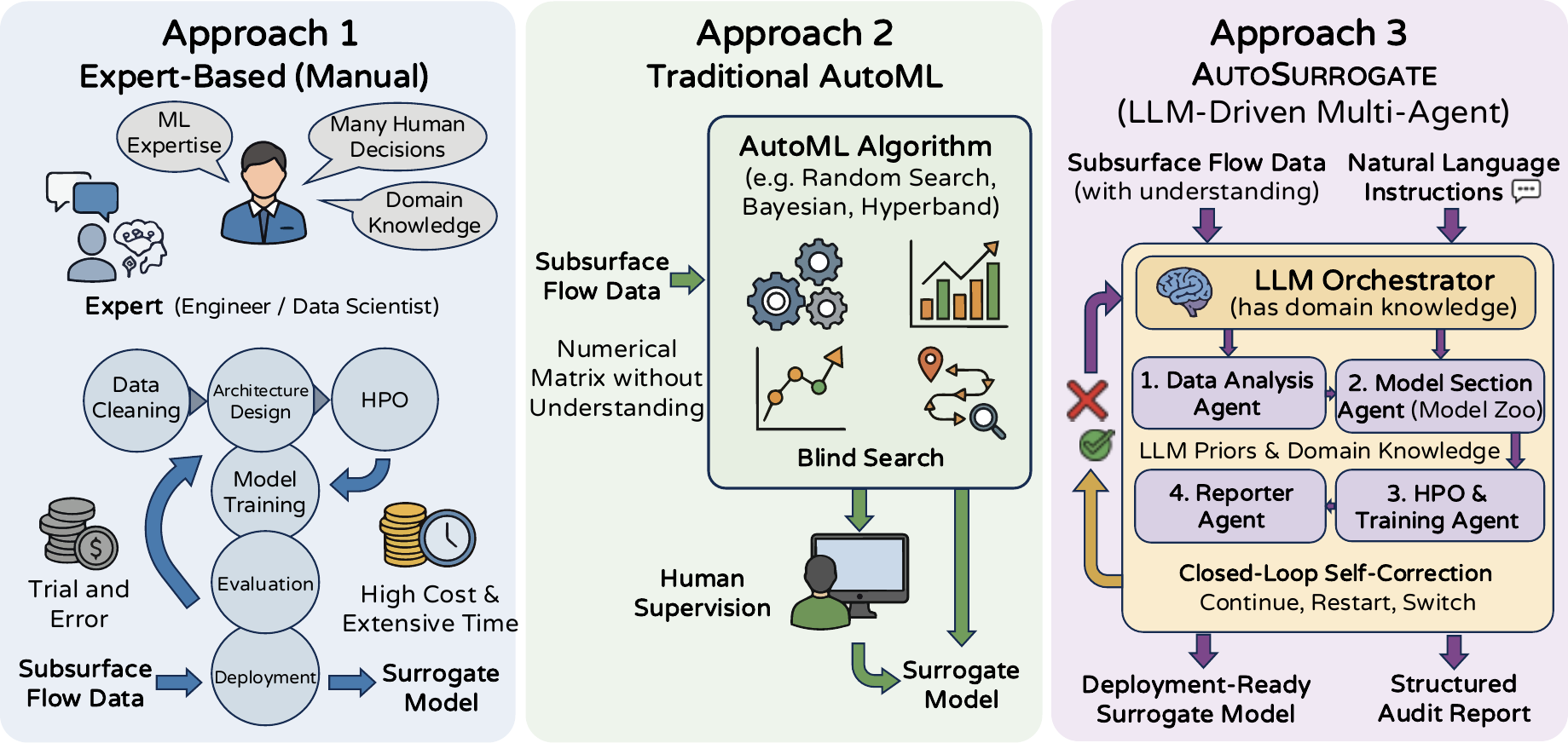}
    \caption{Three approaches to constructing deep learning surrogates for subsurface flow. Approach 1: manual, expert-driven pipeline relying on trial-and-error design and tuning. Approach 2: domain-agnostic AutoML, which performs brute-force search without exploiting physical priors. Approach 3: the proposed \textsc{AutoSurrogate}, an LLM-driven multi-agent framework that takes a dataset and a natural-language instruction and autonomously delivers a deployment-ready surrogate through physics-aware reasoning and closed-loop self-correction.}
    \label{fig:three approaches}
\end{figure}

% Challenges
Despite the great potential and broad exploration of DL surrogates, constructing high-quality DL surrogates remains mostly a manual and heuristic process, including data preprocessing, architecture design and selection, hyperparameter tuning, training implementation and monitoring, model evaluation, and model deployment~\citep{guLargeLanguageModels2024}. This process requires substantial ML and DL expertise, which is often beyond the reach of most domain scientists or engineers~\citep{yangHyperparameterOptimizationMachine2020}. Even for practitioners with ML expertise, the model design, training, and hyperparameter tuning tasks are not efficient processes, replying heavily on trial-and-error and iterative refinement, as demonstrated in Approach 1 of \autoref{fig:three approaches}. This gap leads to inefficient and labor-intensive surrogate construction, and constitutes a primary barrier preventing the adoption of DL surrogates in the subsurface modeling community.

While traditional Automated Machine Learning (AutoML) tools exist~\citep{akibaOptunaNextgenerationHyperparameter2019, yangHyperparameterOptimizationMachine2020}, they are fundamentally ``domain-agnostic.'' They treat subsurface flow data as purely abstract numerical matrices, relying on blind, brute-force searches for model architectures, as shown in Approach 2 of \autoref{fig:three approaches}. They fail to understand the underlying physical laws (e.g., multiphase flow, spatiotemporal evolution of pressure and saturation), resulting in highly inefficient search spaces and sub-optimal models for complex PDE-based tasks. Moreover, training processes of DL surrogates could be fragile especially for highly nonlinear subsurface flow datasets, which frequently suffer from numerical instabilities, such as gradient explosions, loss oscillations, or convergence failures~\citep{cuomoScientificMachineLearning2022}. Currently, resolving these issues requires a human-in-the-loop approach, where engineers must manually diagnose training logs and adjust configurations, further limiting the efficiency and reproducibility of surrogate model development. These challenges highlight the need for more automated and robust modeling frameworks.

% LLM and AI agents
The rapid breakthrough of large language models (LLMs)~\citep{naveed2025comprehensive} and autonomous AI agents~\citep{masterman2024landscape,sapkota2025ai} presents promising opportunities to achieve intelligent automation and revolutionize surrogate modeling. LLMs (e.g., GPT-4, Claude, Llama, DeepSeek) have achieved widespread adoption across diverse sectors, demonstrating their remarkable capacity for analysis, reasoning, and knowledge synthesis~\citep{achiam2023gpt,touvron2023llama,guo2025deepseek}.
LLMs have also catalyzed the development of AI agents, which have garnered increasing attention in recent years~\citep{moritz2025coordinated,gao2024empowering}. AI agents are autonomous systems capable of sensing input data, processing goals through reasoning, planning potential solutions, and interacting with environments to accomplish assigned tasks~\citep{fang2025comprehensive}. The foundation model (typically an LLM) serves as the cognitive core of the AI agent, interpreting input data, analyzing objectives, generating plans, and producing responses or actions.
AI agents have demonstrated versatility across numerous domains, with successful applications in healthcare~\citep{moritz2025coordinated}, software and hardware development~\citep{wang2024openhands, liuVeriSureContractAwareMultiAgent2026}, industrial systems~\citep{liuInsightXAgentLMMBased2026, liuAeroGPTLeveragingLargeScale2026, xieRapidGenerationMethod2025} etc., establishing a strong foundation for their potential application to subsurface flow modeling challenges.

% our contribution
In this work, we propose \textsc{AutoSurrogate}, an LLM-driven multi-agent framework designed to autonomously construct, train, and deploy deep learning surrogate models for subsurface flow modeling. Given a dataset and optional human instructions, the framework executes an end-to-end workflow that covers data inspection, architecture selection, hyperparameter optimization, training, evaluation, and artifact reporting, as shown in Approach 3 of \autoref{fig:three approaches}. Rather than treating surrogate construction as a monolithic prompting task, \textsc{AutoSurrogate} decomposes the workflow into a sequence of specialized agents that reason at a high level and invoke deterministic tools for execution. 
\textsc{AutoSurrogate} is built around three principles that differ from existing methods. First, it actively utilizes domain knowledge and reasoning rather than conducting random searches: the central LLM reasons over the physical semantics of the task and selects among a set of surrogate architectures tailored to structured spatiotemporal flow data. Second, it provides training process monitoring and closed-loop feedback: training failures and unstable runs trigger autonomous corrective actions, including tighter stability settings, continued training, and model switching. Third, it ensures tractability of all steps: all intermediate decisions, metrics and checkpoints are stored as structured artifacts that can be inspected after the run.

The contributions of this work are threefold. First, by constructing a novel end-to-end natural language-driven multi-agent system, we could reduce the technical barrier of DL surrogates and minimize the manual efforts required for constructing DL surrogates. By simply providing data and instructions, domain scientists or engineers can trigger a collaborative workflow among specialized AI agents to generate a deployment-ready surrogate, effectively democratizing advanced DL techniques for the subsurface research community. Second, unlike traditional black-box AutoML, \textsc{AutoSurrogate} leverages the vast prior knowledge embedded in LLMs and their reasoning ability to interpret the physical context of the surrogate modeling task and intelligently explore the search space to prioritize physically appropriate architectures. Third, the designed closed-loop monitoring and feedback mechanism is able to address the fragility of DL surrogate training, continuously detecting numerical instabilities or performance plateaus. Upon diagnosing a failure, the system autonomously applies corrective actions, such as dynamically adjusting learning rates, modifying regularization, or swapping model architectures. This mechanism can ensure robust, human-free training on complex multiphase flow datasets. With these contributions, the proposed framework establishes a new paradigm for surrogate modeling, shifting from manual and heuristic workflows to automated, scalable, and intelligent model construction.

This paper proceeds as follows. In \autoref{sec:method}, we formulated the surrogate modeling task, introduced the \textsc{AutoSurrogate} framework, and elaborated on the key components of the framework. In \autoref{sec:case}, the \textsc{AutoSurrogate} framework is tested and validated with a geological carbon storage (GCS) case. At last, discussions and conclusions are provided in \autoref{sec:conclusion}.

\section{Methodology}
\label{sec:method}
In this section, we first introduce the task of surrogate modeling for subsurface flow problems; then we present the proposed \textsc{AutoSurrogate} framework, which integrates data profiling, architecture selection, hyperparameter optimization, model training, closed-loop self-correction, and reporting into a multi-agent system, as shown in \autoref{fig:overall}.

% ------------------------------------------------------------
\subsection{Surrogate Modeling Task Formulation}
% ------------------------------------------------------------
Given a set of model parameters $\mathbf{m}$ (e.g., static geological properties such as permeability and porosity fields), the high-fidelity flow simulation can be formulated as follows:
\begin{equation}
  \mathbf{d} = f(\mathbf{m}),
  \label{eq:sim}
\end{equation}
where $f$ denotes the forward flow simulation, and $\mathbf{d}$ denotes the quantities of interest (QoIs), which could be dynamic reservoir state properties, such as pressure and saturation fields, or well-level responses, such as production rates and bottom-hole pressures. Due to the high computational cost of evaluating $f$, surrogate models are constructed to approximate the forward mapping, i.e.,
\begin{equation}
  \hat{\mathbf{d}} = \hat{f}_{\theta}(\mathbf{m}),
  \label{eq:surr}
\end{equation}
where $\hat{f}_{\theta}$ denotes the surrogate model with parameters $\theta$, which is a computationally efficient approximation of the original simulator, and $\hat{\mathbf{d}}$ denotes the predictions of QoIs. 

To train the surrogate, the training dataset ($\mathcal{D}$) needs to be constructed first, and a series of input–output pairs, i.e., $\{\mathbf{m}^{(i)}, \mathbf{d}^{(i)}\}_{i=1}^{N}$ could be generated by running the forward simulator under diverse parameter configurations. Specifically, the model parameters $\mathbf{m}$ (e.g., permeability fields) are sampled from prescribed prior distributions to capture variability in the parameter space. For each sampled realization, the forward simulator $f$ is evaluated to obtain the corresponding responses $\mathbf{d}$. Then the goal of surrogate modeling is to learn the mapping defined in Eq.~\ref{eq:surr} from the dataset $\mathcal{D} = \{\mathbf{m}^{(i)}, \mathbf{d}^{(i)}\}_{i=1}^{N}$. In this work, a geological carbon storage problem is considered for surrogate modeling, where the QoIs are the spatial variation and temporal evolution of pressure fields, $\mathbf{P}^{(i)}$, and CO$_2$ saturation fields, $\mathbf{S}^{(i)}$, corresponding to a given permeability field $\mathbf{K}^{(i)}$. Accordingly, the outputs are defined as $\mathbf{d}^{(i)}=[\mathbf{P}^{(i)},\, \mathbf{S}^{(i)}]$, and the dataset can be written as $\mathcal{D} = \{\mathbf{K}^{(i)},\, \mathbf{P}^{(i)},\, \mathbf{S}^{(i)}\}_{i=1}^{N}$. Separate surrogates need to be constructed for pressure and CO$_2$ saturation (denoted as $\hat{f}_{\theta_p}$ and $\hat{f}_{\theta_s}$), respectively. 

The task becomes more challenging for problems involving spatial heterogeneity and temporal dynamics, as the surrogate model must capture both the complex spatial structure of the geological system and the temporal evolution of fluid flow processes. Moreover, in multiphase flow systems, pressure and saturation fields exhibit qualitatively distinct behaviors. In particular, pressure fields are typically globally smooth and governed by quasi-elliptic behavior, whereas saturation fields often develop sharp displacement fronts driven by hyperbolic transport.These contrasting characteristics impose different requirements on surrogate model design, including network architecture and loss functions. As a result, generic regression models are often insufficient to capture such heterogeneous physical behaviors or to accommodate the distinct spatial and temporal dynamics inherent in multiphase flow systems.

The manual construction of tailored surrogate models that account for the underlying spatio-temporal structure and physical characteristics involves a sequence of interdependent decisions, including selecting appropriate neural network architectures for spatio-temporal data, exploring a hyperparameter space, and determining whether underperforming models should be refined, reconfigured, or replaced. Each decision requires expert judgment, where suboptimal choices at any stage can propagate and negatively impact subsequent steps. \textsc{AutoSurrogate} decomposes the surrogate construction process into a sequence of autonomous reasoning tasks, each handled by a specialized LLM-driven agent equipped with a well-defined set of tools.

\subsection{Framework Overview}
% ------------------------------------------------------------

\textsc{AutoSurrogate} is organized as a stage-wise framework in which four specialized agents collaborate with each other to produce a high-quality, deployment-ready surrogate model from simulation dataset following optional natural-language instructions. In \textsc{AutoSurrogate}, each agent is backed by an LLM that is responsible for high-level decision-making, such as selecting an appropriate architecture or diagnosing a training failure, while all computationally critical operations are carried out by predefined tools. This separation between semantic reasoning and deterministic execution prevents the imprecise nature of LLM outputs from introducing errors into numerical computations.

Specifically, each agent operates in an iterative reason--act cycle~\citep{yaoReActSynergizingReasoning2023}. Given the current workflow state, the LLM first generates a natural-language reasoning trace that explains its assessment of the situation and its intended next step. It then issues a structured call to one of its permitted tools, observes the tool's output, and updates its internal reasoning accordingly. This cycle is repeated until the agent determines that its objective has been achieved.

To enable effective communication and collaboration among agents, we establish a typed shared memory object, which functions as a global blackboard accessible to all agents throughout the workflow. Rather than passing large data tensors or model checkpoints through natural-language messages, agents write structured outputs, including data profiles, model selections, hyperparameter configurations, training histories, and error logs, to this shared memory, and retrieve the outputs of preceding stages from it. This design decouples inter-agent communication from agent reasoning, which can preserve the expressiveness of natural-language coordination and maintain the reliability and consistency of structured state management.

\begin{figure}
    \centering
    \includegraphics[width=\linewidth]{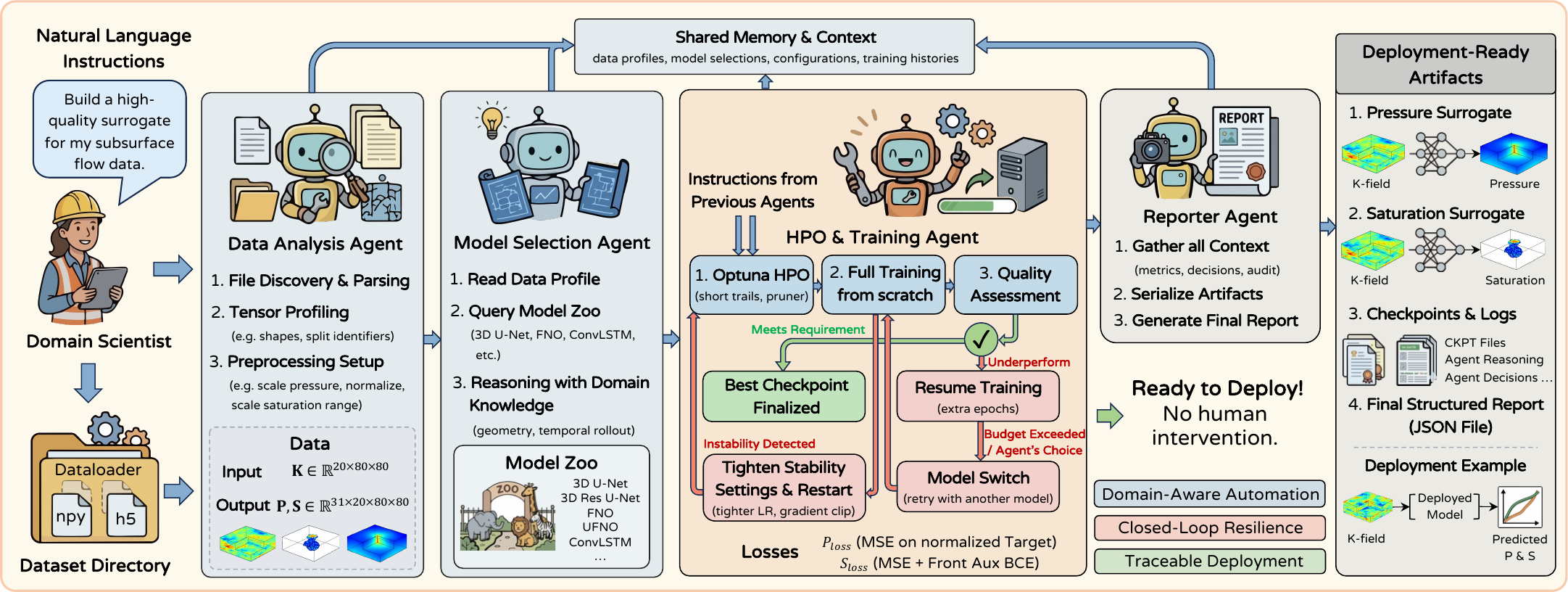}
    \caption{Overall architecture of the \textsc{AutoSurrogate} framework. Given simulation data and a natural-language instruction, four specialized agents collaborate through a shared memory context to autonomously produce deployment-ready surrogate models. The HPO \& Training Agent incorporates a closed-loop self-correction mechanism that handles training instabilities and suboptimal convergence through continuation, stability-constrained restart, or architecture switching.}
    \label{fig:overall}
\end{figure}

The workflow proceeds through four sequential stages, each governed by a dedicated agent. The \textit{Data Analysis Agent} (\autoref{sec:data_profiling}) inspects the raw simulation dataset, validates its structural consistency, and produces a semantic task representation that is shared with all downstream agents. The \textit{Model Selection Agent} (\autoref{sec:arch_select}) reasons over this representation to select an initial surrogate architecture for each prediction task from a model zoo (denoted as $\mathcal{M}_{\mathrm{zoo}}$), which is a collection of different neural network structures. The \textit{HPO \& Training Agent} (\autoref{sec:hpo} and \autoref{sec:selfcorrect}) executes a two-level optimization loop that includes hyperparameter search, full model training, performance evaluation, and autonomous recovery from failure. Finally, the \textit{Reporter Agent} (\autoref{sec:report}) consolidates all decisions, metrics, and artifacts into a structured report to support reproducibility and post-hoc analysis. The end-to-end procedure is summarized in \autoref{alg:pipeline}, and the overall architecture of the multi-agent framework is illustrated in \autoref{fig:overall}.

\begin{algorithm}[t]
\caption{The \textsc{AutoSurrogate} pipeline.}
\label{alg:pipeline}
\begin{algorithmic}[1]
\Require Dataset $\mathcal{D} = \{\mathbf{K}^{(i)},\, \mathbf{P}^{(i)},\, \mathbf{S}^{(i)}\}_{i=1}^{N}$; optional instruction $\mathcal{I}$
\Ensure Surrogates $\hat{f}_{\theta_p},\, \hat{f}_{\theta_s}$; structured report $\mathcal{R}$
\Statex \hspace{-\algorithmicindent}\textit{Stage 1 --- Data Analysis}
\State $\mathcal{P} \gets \textsc{ProfileData}(\mathcal{D})$ \Comment{shapes, splits, value ranges}
\State $\mathcal{C} \gets \textsc{ConfigurePreprocessing}(\mathcal{P}, \mathcal{I})$
\Statex \hspace{-\algorithmicindent}\textit{Stage 2 --- Model Selection}
\For{each target $t \in \{\text{pressure},\, \text{saturation}\}$}
    \State $\mathcal{A}_t \gets \textsc{ReasonAndSelect}(\mathcal{P},\, \mathcal{M}_{\mathrm{zoo}},\, \mathcal{I})$
    \If{$\textsc{EstimateMemory}(\mathcal{A}_t,\, \mathcal{P}) > B_{\mathrm{GPU}}$}
        \State $\mathcal{A}_t \gets \textsc{SelectAlternative}(\mathcal{M}_{\mathrm{zoo}} \setminus \{\mathcal{A}_t\})$
    \EndIf
\EndFor
\Statex \hspace{-\algorithmicindent}\textit{Stage 3 --- HPO, Training \& Self-Correction}
\For{each target $t \in \{\text{pressure},\, \text{saturation}\}$}
    \State $\boldsymbol{\lambda}_t^* \gets \textsc{BayesianHPO}(\mathcal{A}_t,\, \mathcal{D}_{\mathrm{train}},\, \mathcal{D}_{\mathrm{val}})$
    \State $\hat{f}_{\theta_t} \gets \textsc{Train}(\mathcal{A}_t,\, \boldsymbol{\lambda}_t^*,\, \mathcal{D}_{\mathrm{train}})$
    \While{$\lnot\,\textsc{MeetsQuality}(\hat{f}_{\theta_t})$ \textbf{and} budget remains}
        \State $\hat{f}_{\theta_t} \gets \textsc{SelfCorrect}(\hat{f}_{\theta_t},\, \mathcal{A}_t,\, \mathcal{M}_{\mathrm{zoo}})$ \Comment{\autoref{alg:selfcorrect}}
    \EndWhile
\EndFor
\Statex \hspace{-\algorithmicindent}\textit{Stage 4 --- Reporting}
\State $\mathcal{R} \gets \textsc{Consolidate}(\text{all decisions, metrics, checkpoints})$
\State \Return $\{\hat{f}_{\theta_p},\, \hat{f}_{\theta_s}\},\, \mathcal{R}$
\end{algorithmic}
\end{algorithm}

% ------------------------------------------------------------
\subsection{Physics-Aware Data Profiling}
\label{sec:data_profiling}
% ------------------------------------------------------------

At the initial stage, the \textit{Data Analysis Agent} performs a structured inspection of the raw dataset and constructs a semantically rich task representation that is shared with all downstream agents.

Specifically, the agent identifies the dataset composition, validates sample count consistency across input and output files, and resolves train/validation/test splits. It then encodes the task in physically meaningful terms, including the spatial resolution $(D_x \times D_y \times D_z)$, the number of output timesteps~$T$, the training set size $N_{\mathrm{train}}$, as well as the types and approximate value ranges of the output variables. These attributes are recorded explicitly as part of the data profile. This representation serves as the semantic interface between the data and the reasoning agents, which means that rather than reasoning over specific numerical tensors and matrices, downstream agents can reason over structured descriptions, such as ``a 31-step temporal rollout from a static $80 \times 80 \times 20$ permeability field with 700 training samples.'' This treatment enables more interpretable and context-aware decision-making in subsequent stages. 

Preprocessing strategies are also determined through agentic reasoning rather than hard-coded rules. Pressure outputs are scaled by $10^5$ to bring them into a numerically tractable range and then normalized using z-score statistics (mean, $\mu_p$, and standard deviation, $\sigma_p$) computed over the dataset-level. In contrast, saturation outputs, which are physically bounded in $[0,1]$, are retained in their native range to preserve the interpretability of displacement fronts. All of these preprocessing decisions are logged to the shared context so that training and evaluation modules operate on the same configuration, preventing inconsistencies between training and inference.

% ------------------------------------------------------------
\subsection{Task-Specific Architecture Selection}
\label{sec:arch_select}
% ------------------------------------------------------------

A fundamental limitation of conventional AutoML approaches for surrogate modeling is their treatment of architecture selection as a domain-agnostic search problem. By exploring candidate architectures through black-box optimization over an unconstrained search space, these methods fail to leverage the substantial prior knowledge available about the underlying physical structure of the problem, which an experienced ML practitioner would routinely exploit to dramatically narrow the search space. \textsc{AutoSurrogate} addresses this by repositioning architecture selection as an LLM-driven reasoning task, in which a language model acts as a proxy for domain expertise, applying prior knowledge about the relationship between the physical problem context and model inductive biases to identify appropriate candidate architectures.

%  3D U-Net, 3D Residual U-Net CNN-Transformer的引用

This reasoning operates over a model zoo containing eight surrogate architecture families: 3D U-Net~\citep{wang2025deep}, 3D Residual U-Net~\citep{jiang2021deep}, Recurrent Residual U-Net~\citep{tangDeeplearningbasedSurrogateFlow2021}, Encoder--Decoder ConvLSTM~\citep{fengEncoderdecoderConvLSTMSurrogate2024}, CNN-Transformer~\citep{feng2025hybrid}, U-DeepONet~\citep{diabUDeepONetUNetEnhanced2024}, Fourier Neural Operator (FNO)~\citep{liFourierNeuralOperator2020}, and U-FNO~\citep{wen2022u}. Each model entry includes not only its constructor parameters and associated hyperparameter search space, but also a natural-language description of its inductive biases and the types of spatiotemporal structure it is best suited to capture. The searchable hyperparameters exposed by the model zoo are summarized in \autoref{tab:search_space}: a common set of optimization and width hyperparameters is shared across all architectures, while a small number of architecture-specific knobs are additionally exposed where relevant. The concrete search range for each hyperparameter is not fixed a priori, but is instead determined by the \textit{HPO \& Training Agent} based on the data profile and the selected architecture. Given the data profile generated in the previous stage and the full model zoo, the \textit{Model Selection Agent} reasons over the correspondence between task characteristics and architectural strengths to identify suitable candidate architectures.

\begin{table}[pos=htbp]
\centering
\caption{Searchable hyperparameters exposed by the \textsc{AutoSurrogate} model zoo. The common set applies to all eight architectures in the zoo, whereas architecture-specific hyperparameters are additionally exposed only for the indicated model family. Concrete search ranges are not fixed a priori and are instead determined by the \textit{HPO \& Training Agent} based on the task profile and the selected architecture.}
\label{tab:search_space}
\small
\begin{tabular}{l l l}
\toprule
Category & Hyperparameter & Applies to \\
\midrule
\multirow{3}{*}{Optimization} & Learning rate & All \\
 & Weight decay & All \\
 & Batch size & All \\
\midrule
Capacity & Base channel width & All \\
\midrule
\multirow{2}{*}{Architecture-specific} & Transformer layers & CNN-Transformer \\
 & Transformer heads & CNN-Transformer \\
\bottomrule
\end{tabular}
\end{table}

This reasoning process mirrors the kind of expert deliberation that a human ML practitioner would apply. For example, given the multi-step temporal rollout requirement, the agent can recognize that architectures with explicit recurrent or temporal-attention mechanisms are preferable to purely convolutional architectures. Given the distinct spatial characteristics of pressure (smooth, globally correlated) versus saturation (sharp fronts, locally dominant), the agent can select different architectures for the two quantities, which is advantageous over methods that apply a single model architecture uniformly. For pressure, architectures with strong multi-scale spatial representations such as residual U-Nets may be preferred; for saturation, architectures with auxiliary front-aware mechanisms have a higher probability of being selected.

Semantic reasoning alone, however, is not sufficient to guarantee operational feasibility. An architecture that is physically appropriate for the studied problem may exceed available GPU memory at the target resolution. To address this, the \textit{Model Selection Agent} can invoke a memory estimation tool that performs a forward pass on a randomly initialized model at the target tensor dimensions, returning an estimate of the parameter count and peak activation memory. Architectures that exceed the available memory budget are excluded before a selection is committed. This two-stage process, semantic prior followed by feasibility verification, ensures that the selected architecture is both physically motivated and practically executable.

Importantly, the architecture selected at this stage is treated as an initial hypothesis rather than an irrevocable commitment. As detailed in the following sections, the training loop is explicitly permitted to revise this choice if convergence or accuracy targets are not met. This design reflects a broader principle in \textsc{AutoSurrogate}: no single agent makes a globally final decision; instead, decisions are revisable based on evidence accumulated during execution.

% ------------------------------------------------------------
\subsection{Task-Aware Hyperparameter Optimization and Training}
\label{sec:hpo}
% ------------------------------------------------------------

Given the selected architecture, the \textit{HPO \& Training Agent} conducts a two-level optimization procedure that balances exploration efficiency with training quality. In the first level, a short-budget hyperparameter search is performed using Optuna with a median-pruner stopping rule~\citep{akibaOptunaNextgenerationHyperparameter2019}. Each trial is sampled from the search space defined in the selected model's registry entry, covering parameters such as channel width, normalization type, learning rate, weight decay, and batch size, and is evaluated after a small number of epochs. The pruner terminates trials whose intermediate performance falls below the running median, concentrating the computational budget on the most promising configurations. Upon completion of the search, the configuration that achieved the best intermediate score is used to launch a full training run from scratch.

A distinguishing feature of the training procedure is that both the loss functions and the optimization targets are designed with awareness of the physical properties of each QoI. For pressure, the loss function of the surrogate model is defined with mean-squared error on the normalized target:
\begin{equation}
  \mathcal{L}_p = \mathrm{MSE}(\hat{P}^{\prime}, P^{\prime}),
  \qquad P^{\prime} = \frac{P / 10^5 - \mu_p}{\sigma_p + \epsilon},
  \label{eq:loss_p}
\end{equation}
where $\epsilon$ is a small positive constant (set to $10^{-8}$) added to the denominator to guard against division by zero when the pressure standard deviation $\sigma_p$ is near zero, and normalization ensures numerical stability given the large absolute magnitudes of reservoir pressure.

For saturation, whose primary challenge lies in accurately capturing the displacement front rather than the bulk field value, a composite loss is employed:
\begin{equation}
  \mathcal{L}_s = \mathrm{MSE}\!\left(\sigma(\hat{S}),\, S\right) +
  \lambda_{\mathrm{bce}}\,
  \mathrm{BCEWithLogits}\!\left(\hat{S},\, \mathbb{1}[S > \tau]\right),
  \label{eq:loss_s}
\end{equation}
where $S$ denotes the ground-truth CO$_2$ saturation field and $\hat{S}$ denotes the raw logit field output by the network; $\sigma(\cdot)$ is the element-wise sigmoid function that maps the logits back to the physically admissible range $[0,1]$ before computing the MSE term against $S$. The indicator $\mathbb{1}[S > \tau]$ produces a binary mask of CO$_2$-invaded cells defined by a small saturation threshold $\tau$ (set to $10^{-4}$), and $\mathrm{BCEWithLogits}(\cdot,\cdot)$ applies the numerically stable log-sum-exp form of the binary cross-entropy loss directly on the logits $\hat{S}$ to supervise this front-presence classification. The auxiliary binary cross-entropy term thereby encourages the model to correctly locate the saturation front~\citep{mo2019deep}. The weighting coefficient $\lambda_{\mathrm{bce}}$ balances the field-level MSE term and the front-aware classification term and is treated as a hyperparameter included in the Optuna search space. During evaluation, predictions are mapped back to original physical scales before computing test metrics, ensuring that $R^2$, RMSE, and relative $L_2$ error are reported in physically interpretable units.

Early stopping is employed to mitigate over-fitting in the training process and reduce resource consumption, which is determined based on the combined training-plus-validation loss,
\begin{equation}
  \text{score} = \mathcal{L}_{\text{train}} + \mathcal{L}_{\text{val}},
  \label{eq:stop}
\end{equation}
which helps avoid over-sensitivity to transient validation loss fluctuations that are common when training on the nonlinear response surfaces. Model checkpoints corresponding to both the lowest-loss epoch and the final epoch are retained throughout training, and per-epoch metrics are logged to support post-hoc diagnosis.

% ------------------------------------------------------------
\subsection{Autonomous Closed-Loop Self-Correction}
\label{sec:selfcorrect}
% ------------------------------------------------------------

A key capability of \textsc{AutoSurrogate} is its ability to autonomously recover from training failures and suboptimal outcomes without human intervention. In conventional surrogate modeling workflows, numerical instability or insufficient model accuracy typically require manual inspection of training logs followed by ad-hoc configuration adjustments, which is a process that is both time-consuming and difficult to systematize. \textsc{AutoSurrogate} replaces this human-in-the-loop intervention with a closed-loop control mechanism embedded in the \textit{HPO \& Training Agent}, which continuously monitors training dynamics and applies structured corrective actions according to a principled recovery strategy.

The control loop begins with an instability detection stage, in which the training process is evaluated against a set of failure criteria, including non-finite training or validation loss, exploding or vanishing gradient norms, and HPO runs in which no trial completes successfully. These failure modes are particularly common in DL surrogate training, where complex physical dynamics induce highly nonlinear loss landscapes that are sensitive to initialization and learning rate. By explicitly identifying such instabilities, rather than relying on generic convergence criteria, the recovery mechanism is able to prescribe targeted corrective actions.

Upon diagnosis, the agent selects a recovery action from:
\begin{enumerate}[leftmargin=*, label=(\roman*)]
  \item \textbf{Continuation}: If the model is numerically stable but has not yet converged to the target accuracy, training is resumed from the latest checkpoint for additional epochs and the model is reassessed.
  \item \textbf{Stability-constrained restart}: If numerical instability is detected, the hyperparameter search space is tightened by reducing the upper bound on admissible learning rates and strengthening gradient clipping constraints. A new HPO-plus-training cycle is then launched with the same architecture.
  \item \textbf{Architecture switching}: If the current architecture remains unsatisfactory after multiple retries, the agent selects the next most promising candidate from the model zoo. The selection follows the same semantic reasoning process as in the initial selection stage, but with the performance history of the current model incorporated as additional evidence.
  \item \textbf{Global-best fallback}: If the per-task retry budget is exhausted, the framework selects the best checkpoint from the entire search history, including all training runs, retries, and architectures, for deployment.
\end{enumerate}

Checkpoint management is fully integrated into the control loop. At each evaluation point, the framework maintains a global-best checkpoint that is updated whenever any training run, across all retries and architectures, surpasses all previous results. This design ensures that the final deployed surrogate represents the best model identified across the entire optimization trajectory, rather than merely the output of the last training run.

The complete self-correction procedure is summarized in \autoref{alg:selfcorrect}. Overall, this mechanism ensures that the quality of the surrogate is at least as good as the best result achieved within the allocated budget across the entire model registry, while the system remains fully unsupervised throughout. By embedding decision logic that would otherwise require human expertise directly into the agent control loop, \textsc{AutoSurrogate} enables DL surrogate modeling in settings where practitioner involvement at intermediate stages is not feasible.

\begin{algorithm}[t]
\caption{Closed-loop self-correction procedure.}
\label{alg:selfcorrect}
\begin{algorithmic}[1]
\Require Model $\hat{f}_{\theta}$, architecture $\mathcal{A}$, model zoo $\mathcal{M}_{\mathrm{zoo}}$, training history $\mathcal{H}$
\Ensure Improved model $\hat{f}_{\theta^*}$
\State $\hat{f}_{\theta^*} \gets \hat{f}_{\theta}$
\State $s \gets \textsc{Diagnose}(\mathcal{H})$ \Comment{inspect loss curve, gradient norms}
\If{$s = \textsc{Converging}$} \Comment{stable but below accuracy target}
    \State $\hat{f}_{\theta} \gets \textsc{ResumeTraining}(\hat{f}_{\theta},\, E_{\mathrm{extra}})$
\ElsIf{$s = \textsc{Unstable}$} \Comment{NaN loss or gradient explosion}
    \State Tighten learning-rate bounds; strengthen gradient clipping
    \State $\boldsymbol{\lambda}^{\prime} \gets \textsc{BayesianHPO}(\mathcal{A},\, \text{constrained space})$
    \State $\hat{f}_{\theta} \gets \textsc{Train}(\mathcal{A},\, \boldsymbol{\lambda}^{\prime})$ \Comment{restart from scratch}
\ElsIf{$s = \textsc{Underperforming}$} \Comment{budget exceeded or persistent failure}
    \State $\mathcal{A} \gets \textsc{SelectNext}(\mathcal{M}_{\mathrm{zoo}},\, \mathcal{H})$ \Comment{switch architecture}
    \State $\boldsymbol{\lambda}^{\prime} \gets \textsc{BayesianHPO}(\mathcal{A})$
    \State $\hat{f}_{\theta} \gets \textsc{Train}(\mathcal{A},\, \boldsymbol{\lambda}^{\prime})$
\EndIf
\If{$\textsc{Quality}(\hat{f}_{\theta}) > \textsc{Quality}(\hat{f}_{\theta^*})$}
    \State $\hat{f}_{\theta^*} \gets \hat{f}_{\theta}$
\EndIf
\State \Return $\hat{f}_{\theta^*}$
\end{algorithmic}
\end{algorithm}

% ------------------------------------------------------------
\subsection{Traceability and Reporting}
\label{sec:report}
% ------------------------------------------------------------

Scientific credibility requires not only strong predictive performance of the surrogate model, but also transparency, auditability, and reproducibility in the model construction process. To meet these requirements, the \textit{Reporter Agent} consolidates the full execution history of the pipeline into a structured report that accompanies the deployed model artifacts.

The report captures the complete decision chain of the workflow, including the dataset configuration discovered during the data analysis stage, the preprocessing settings applied to each QoI, the architectural reasoning generated during the model selection stage, the hyperparameter search histories and corresponding optimal configurations, the per-epoch training records, the test-set evaluation metrics, the checkpoint selected for deployment, any error traces encountered during execution, and the audit log recording how each user-provided natural-language instruction influenced specific decisions. This reporting layer transforms \textsc{AutoSurrogate} from a black-box automation tool into a transparent and reproducible scientific workflow. By preserving the intermediate reasoning alongside the final artifacts, the framework enables practitioners to verify that the selected surrogate is appropriate for their application, diagnose the cause of any suboptimal outcomes, and reproduce or extend the optimization process on new data or with updated requirements.

\section{Experiments}
\label{sec:case}
In this section, the performance of the proposed \textsc{AutoSurrogate} framework for automatic surrogate modeling is evaluated on a synthetic 3D geological carbon storage case. We first describe the geological model and simulation setup, and then detail the experimental configuration including baseline architectures and training protocols. The quantitative comparison against eight baseline models and three AutoML methods is presented, followed by an analysis of search efficiency. We further provide two case studies that examine the distinctive capabilities of the framework: interpretable decision-making tracing and autonomous self-correction. Finally, representative field visualizations are presented to qualitatively assess prediction quality.
\subsection{Case Setup}
\begin{figure}[pos=htbp]
    \centering
    \includegraphics[width=\linewidth]{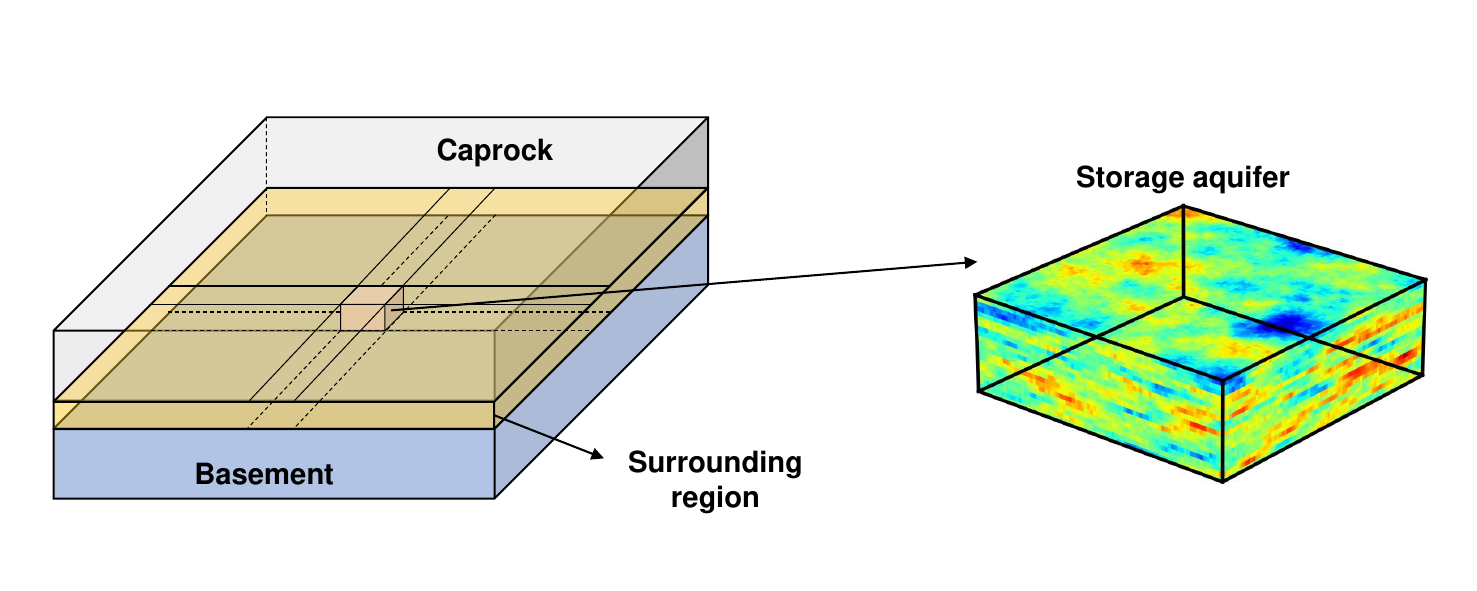}
    \caption{Schematic of the geological model for CO$_2$ storage, showing the full domain and the central storage aquifer.}
    \label{fig:geomodel}
\end{figure}

A 3D geological carbon storage case is considered in this work. The full physical domain of the geological model consists of a storage aquifer, a surrounding region, a caprock and a basement, as demonstrated in \autoref{fig:geomodel}, similar geomodel setup has been utilized in \citet{han2025accelerated}. The size of the full domain is 120 km $\times$ 120 km $\times$ 2.5 km, and is discretized into 100 $\times$ 100 $\times$ 30 non-uniform grid systems. The central storage aquifer, which serves as the CO$_2$ storage target zone, is the primary region of interest for modeling and analysis, even though it represents only a small portion of the full domain (12 km $\times$ 12 km $\times$ 100 m). Therefore, the storage aquifer is discretized into a fine-scale grid consisting of $80 \times 80 \times 20$ cells. While the surrounding region, the caprock, and the basement are discretized with a much coarser grid. 

Following the setup in~\citet{han2025accelerated}, the geological properties for the surrounding region, the caprock, and the basement are set to be homogeneous, as presented in \autoref{tab:case parameters}. For the storage aquifer, we consider heterogeneous geological fields, including permeability and porosity. The permeability of the storage aquifer is assumed to follow a log-normal distribution, with the mean of log-permeability ($\log k$) being 2.5 and the standard deviation of $\log k$ being 1.5. The permeability is also considered anisotropic, with the ratio between vertical permeability ($k_v$) and horizontal permeability ($k$) being 0.1. 

The correlation length of log-permeability ($\log k$) fields is set to be 3600 m in x and y directions ($l_x=l_y=3600$ m) and 15 m in z direction ($l_z=15$ m). The standard Gaussian realizations of $\log k$ are generated with the Stanford Geostatistical Modeling Software package, SGeMS~\citep{remy2009applied}, which have zero mean and unit variance, and can be parameterized with principal component analysis (PCA) to efficiently generate a large number of realizations. With the PCA model constructed, the new realizations could be generated by sampling from the low-dimensional Gaussian space and the standard Gaussian realizations can be further rescaled to the target mean and standard deviation values. 

The porosity of the storage aquifer is assumed to be dependent on the permeability, and the porosity-permeability relationship is formulated as follows:
\begin{equation}
  \phi =d \cdot \log k +e,
  \label{eq:poro_k}
\end{equation}
where $d$ and $e$ are set to be 0.03 and 0.08 in this work, respectively. To avoid nonphysical and extreme values of permeability and porosity, we also setup cutoff thresholds for them. For permeability, the cutoff values are $10^{-4}$ mD and $10^4$ md, and for porosity, the cutoff values are 0.05 and 0.3. 

One fully penetrated vertical injection well is placed in the center of the storage aquifer, which injects CO$_2$ into the aquifer at a rate of 1 Mt/year, and the total injection period is 30 years, which could be divided into 30 time steps. The relative permeability and capillary pressure curves utilized in this work are based on the Brooks-Corey model~\citep{saadatpoor2010new}, and the related coefficients of the model are summarized in \autoref{tab:case parameters}. The simulation of this case is implemented through the open-source multiphysics simulator, GEOS, which is designed for modeling geologic carbon storage processes~\citep{settgast2024geos}.

\begin{table*}[pos=htbp]
\centering
\caption{Parameters utilized for CO$_2$ storage simulation.}
\label{tab:case parameters}
\small
\begin{tabular}{c c}
\toprule
%\multirow{2}{*}{Method} & \multicolumn{3}{c}{Pressure} & \multicolumn{3}{c}{Saturation} \\
%\cmidrule(lr){2-4} \cmidrule(lr){5-7}
% & $R^2$ & RMSE & Rel$L_2$ & $R^2$ & RMSE & Rel$L_2$ \\

Rservoir parameters        & Values  \\
\midrule
Caprock thickness   & 1905 m  \\
Caprock porosity       & 0.08 \\
Caprock permeability (horizontal and vertical)             & 0.001 mD \\
Basement thickness   & 500 m \\
Basement porosity             & 0.09 \\
Basement permeability (horizontal and vertical)          & 2.3 mD \\
Surrounding region thickness             & 100 m  \\
Surrounding region porosity      & 0.2  \\
Surrounding region permeability (horizontal and vertical) & 20 mD  \\
Storage aquifer thickness  & 100 m  \\
\midrule
Relative permeability and capillary pressure parameters & Values  \\
\midrule
Irreducible water saturation, $S_{wi}$  & 0.22 \\
Residual CO$_2$ saturation, $S_{gr}$  & 0 \\
Water exponent for Corey model, $n_w$ & 9 \\
CO$_2$ exponent for Corey model, $n_g$ & 4\\
Relative permeability of CO$_2$ at $S_{wi}$, $k_{rg}(S_{wi})$ & 0.95\\
Capillary pressure exponent, $\lambda$ & 0.55\\

\bottomrule
\end{tabular}
\end{table*}

\subsection{Experiment Setup}
\label{sec:exp_setup}
To construct the training dataset for the case introduced above, 1000 standard Gaussian stochastic fields are generated with SGeMS and parameterized with PCA to reduce the dimension. The constructed PCA can then be utilized to generate more realizations easily. With PCA, 1000 standard Gaussian fields are regenerated, which are further rescaled to the target mean and standard deviation to obtain the log-permeability fields. The porosity fields can be obtained with Eq.~\ref{eq:poro_k}. For the 1000 geological realizations, the corresponding pressure and CO$_2$ saturation fields are solved with the numerical simulator GEOS, to construct the dataset $\mathcal{D} = \{\mathbf{K}^{(i)}, \mathbf{P}^{(i)},\mathbf{S}^{(i)}\}_{i=1}^{1000}$. Each sample takes the static 3D permeability field of the storage aquifer with dimension $80 \times 80 \times 20$ as input and produces pressure and CO$_2$ saturation fields over 31 timesteps (including initial state) as output, each of the same spatial resolution. The dataset is split sequentially into training (700 samples), validation (100 samples), and test (200 samples) sets. The preprocessing strategy follows the physics-aware scheme described in \autoref{sec:data_profiling}: pressure outputs are divided by $10^5$ to convert to bar scale and then z-score normalized ($\mu_p=208.14$, $\sigma_p=6.01$); saturation outputs remain in their native $[0,1]$ range without normalization. All evaluation metrics, including $R^2$, RMSE, and relative $L_2$ error, are computed on the 200-sample test set. For pressure, the metrics are reported on the un-normalized physical scale to ensure that the error magnitudes are directly interpretable in the context of reservoir pressure.

To establish a comprehensive baseline, eight 3D neural network architectures from the \textsc{AutoSurrogate} model zoo are independently trained for both pressure and saturation prediction, yielding 16 baseline models in total. The architectures span a broad range of design paradigms. The convolutional family includes UNet3D~\citep{wang2025deep}, a plain encoder--decoder architecture, and ResUNet3D~\citep{jiang2021deep}, which adds residual skip connections for improved gradient flow. The spectral family includes FNO3D~\citep{liFourierNeuralOperator2020}, a Fourier neural operator that learns mappings in the frequency domain, and UFNO3D~\citep{wen2022u}, which combines spectral processing with local U-Net pathways. UDeepONet3D~\citep{diabUDeepONetUNetEnhanced2024} adopts a spatial branch--temporal trunk decomposition for operator learning. For temporal modeling, CNNTransformer3D~\citep{feng2025hybrid} pairs a CNN encoder with a causal Transformer decoder, EDConvLSTM3D~\citep{fengEncoderdecoderConvLSTMSurrogate2024} employs an encoder--decoder with ConvLSTM bottleneck and CBAM attention, and RecurrentRUNet3D~\citep{tangDeeplearningbasedSurrogateFlow2021} augments the residual U-Net with a ConvLSTM recurrent bottleneck for capturing temporal dynamics. All baselines are trained with a common configuration: AdamW optimizer with learning rate $10^{-4}$ and weight decay $10^{-5}$, fixed learning rate schedule, batch size of 1, gradient clipping at max norm 1.0, and a maximum of 500 epochs with early stopping (patience of 20 epochs) based on the combined training and validation loss (Eq.~\ref{eq:stop}). Please do note that the 500-epoch budget is generously given to allow each baseline model to converge fully, thereby approximating its best achievable performance under fixed hyperparameters. Because no learning rate search or scheduling mechanism is applied, a sufficient training budget is necessary to ensure that no architecture is penalized by an inadequate training duration. Mixed precision is disabled to ensure numerical stability. The loss functions are identical to those used by \textsc{AutoSurrogate}: MSE for pressure (Eq.~\ref{eq:loss_p}) and MSE plus auxiliary BCE ($\lambda_{\mathrm{bce}}=0.01$, $\tau=10^{-4}$) for saturation (Eq.~\ref{eq:loss_s}). For each baseline, the checkpoint with the lowest validation loss is selected for test-set evaluation. Importantly, no hyperparameter optimization is applied to the baselines.

To further contextualize \textsc{AutoSurrogate}'s performance, three standard AutoML baselines are evaluated: Random Search, Optuna TPE (Tree-structured Parzen Estimator)~\citep{akibaOptunaNextgenerationHyperparameter2019}, and Hyperopt TPE~\citep{bergstraMakingScienceModel2013}. Each method is given the same budget and search space as \textsc{AutoSurrogate}, including 120 HPO trials drawn from the joint space of architecture (all 8 candidates) and hyperparameters (learning rate, base channels, normalization, batch size). Each trial trains for 5 epochs with 50 training batches and 20 validation batches per epoch; the best-performing configuration is then retrained for 30 epochs with early stopping. The search space, loss functions, and hardware are identical to those used by \textsc{AutoSurrogate}. This design ensures a fair comparison: all methods have access to the same model zoo and the same total computational budget, differing only in the search strategy.

Note that the two groups of baselines are intentionally configured to isolate different aspects of \textsc{AutoSurrogate}'s advantage. The eight model-zoo baselines are trained with a deliberately generous epoch budget and a well-tuned fixed configuration so that each architecture can reach its own best achievable performance, which approximates the quality that a careful human practitioner would obtain after extensive trial and error on this dataset and therefore serves as a strong upper bound on the accuracy of manually constructed surrogates. The AutoML baselines are instead allocated exactly the same model zoo, search space, per-trial training protocol, and overall computational budget as \textsc{AutoSurrogate}, so that any difference in final accuracy or wall-clock efficiency can be attributed to the search strategy rather than to asymmetric resources. Together, these two groups of comparisons allow us to separately assess whether \textsc{AutoSurrogate} can match or exceed the best hand-tuned surrogates and whether its LLM-driven reasoning yields a faster and more accurate search than conventional AutoML under an identical budget.

\subsection{Performance Evaluation}
\label{sec:perf_eval}

% 点比别人高
% 搜索比别人快/速度比别人快，even with reasoning

\autoref{tab:model_comparison} summarizes the test-set performance of all baseline architectures and two \textsc{AutoSurrogate} configurations: \textsc{AutoSurrogate}@1, which reports the result of a single autonomous pipeline run, and \textsc{AutoSurrogate}@3, which reports the best result obtained across three pipeline runs. For pressure prediction, \textsc{AutoSurrogate}@1 achieves an $R^2$ of 0.9959, which is competitive with the top-performing baselines (RecurrentRUNet3D at 0.9972 and ResUNet3D at 0.9960). \textsc{AutoSurrogate}@3 further improves to $R^2=0.9976$, surpassing all baselines. For CO$_2$ saturation prediction, both \textsc{AutoSurrogate} configurations achieve $R^2=0.9532$, outperforming the best baseline (RecurrentRUNet3D at 0.9359) by 1.73 percentage points. The relative $L_2$ error for saturation is reduced from 0.2508 to 0.2142, representing a 14.6\% improvement. Notably, \textsc{AutoSurrogate} autonomously selected different architectures for the two tasks: ResUNet3D for pressure and UNet3D for saturation (the latter determined through the self-correction mechanism described in \autoref{sec:self_correction_case}), validating the task-specific selection strategy of the framework.

\begin{table*}[pos=htbp]
\centering
\caption{Quantitative comparison of surrogate models for pressure and CO$_2$ saturation prediction. Best results are highlighted in \textbf{bold}.}
\label{tab:model_comparison}
\small
\begin{tabular}{l ccc ccc}
\toprule
\multirow{2}{*}{Method} & \multicolumn{3}{c}{Pressure} & \multicolumn{3}{c}{Saturation} \\
\cmidrule(lr){2-4} \cmidrule(lr){5-7}
 & $R^2$ & RMSE & Rel$L_2$ & $R^2$ & RMSE & Rel$L_2$ \\
\midrule
UDeepONet3D        & 0.8534 & 2.3697 & 0.0114 & 0.8249 & 0.0139 & 0.4144 \\
CNNTransformer3D   & 0.9861 & 0.7300 & 0.0035 & 0.9098 & 0.0100 & 0.2974 \\
EDConvLSTM3D       & 0.9887 & 0.6577 & 0.0032 & 0.9309 & 0.0088 & 0.2604 \\
UFNO3D             & 0.9911 & 0.5853 & 0.0028 & 0.9102 & 0.0100 & 0.2967 \\
RecurrentRUNet3D   & 0.9972 & 0.3290 & 0.0016 & 0.9359 & 0.0084 & 0.2508 \\
FNO3D              & 0.9893 & 0.6395 & 0.0031 & 0.8608 & 0.0124 & 0.3695 \\
ResUNet3D          & 0.9960 & 0.3873 & 0.0019 & 0.9301 & 0.0088 & 0.2619 \\
UNet3D             & 0.9899 & 0.6206 & 0.0030 & 0.9273 & 0.0090 & 0.2670 \\
\midrule
Random Search      & 0.9820 & 0.8300 & 0.0040 & 0.7593 & 0.0163 & 0.4860 \\
Optuna TPE         & 0.9804 & 0.8650 & 0.0042 & 0.8915 & 0.0110 & 0.3260 \\
Hyperopt TPE       & 0.9922 & 0.5460 & 0.0026 & 0.8848 & 0.0113 & 0.3360 \\
\midrule
\textsc{AutoSurrogate}@1  & 0.9959 & 0.3975 & 0.0019 & \textbf{0.9532} & \textbf{0.0072} & \textbf{0.2142} \\
\textsc{AutoSurrogate}@3  & \textbf{0.9976} & \textbf{0.3048} & \textbf{0.0015} & \textbf{0.9532} & \textbf{0.0072} & \textbf{0.2142} \\

\bottomrule
\end{tabular}
\end{table*}

% AutoSurrogate Saturation: with UNet3D, Pressure with ResUNet3D

\begin{figure*}[pos=htbp]
    \centering
    \includegraphics[width=\textwidth]{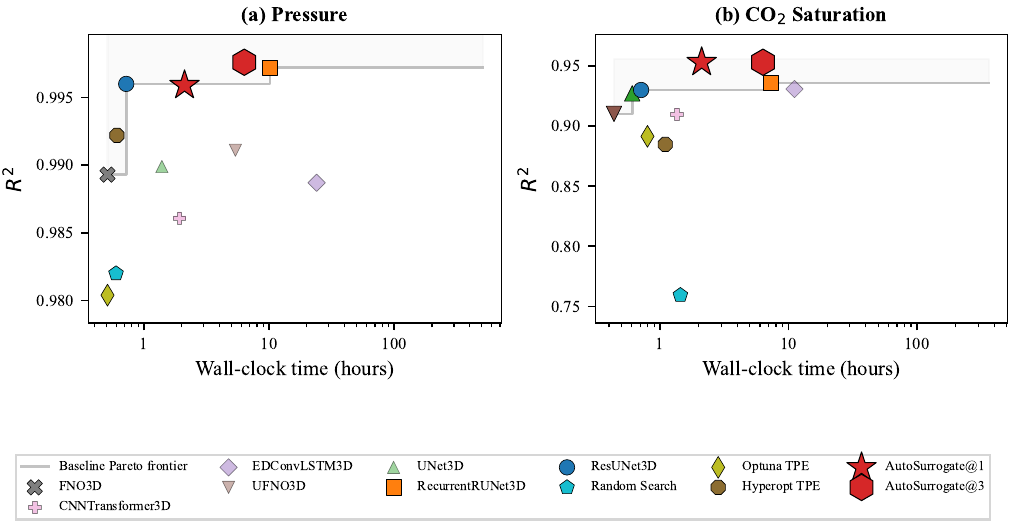}
    \caption{Performance--efficiency Pareto frontier for pressure and CO$_2$ saturation prediction. The step line connects Pareto-optimal baseline configurations. \textsc{AutoSurrogate}@1 (star) and @3 extend the frontier beyond all baselines. AutoML methods are fast but achieve substantially lower $R^2$, especially on saturation where they fall below hand-tuned baselines. UDeepONet3D and FNO3D (saturation) are omitted for clarity.}
    \label{fig:perf_efficiency}
\end{figure*}

\autoref{fig:perf_efficiency} visualizes the performance--efficiency tradeoff as a Pareto frontier, where the step line connects configurations for which no other model achieves both higher $R^2$ and lower wall-clock time; points below the frontier are dominated. On both tasks, \textsc{AutoSurrogate} extends the Pareto frontier beyond all baselines. For pressure, \textsc{AutoSurrogate}@1 achieves $R^2=0.9959$ in 2.11 hours, and \textsc{AutoSurrogate}@3 further improves to $R^2=0.9976$ in 6.34 hours, dominating RecurrentRUNet3D ($R^2=0.9972$, 10.1 hours) in both accuracy and speed. For saturation, \textsc{AutoSurrogate}@1 reaches $R^2=0.9532$ in 2.11 hours, dominating all baselines including RecurrentRUNet3D ($R^2=0.9359$, 7.3 hours) and EDConvLSTM3D ($R^2=0.9309$, 11.1 hours).

The performance variation across baseline architectures is substantial, with $R^2$ ranging from 0.8534 to 0.9972 for pressure and from 0.8249 to 0.9359 for saturation, underscoring that architecture choice is a primary determinant of surrogate quality. Among the baselines, RecurrentRUNet3D and ResUNet3D consistently rank highest, benefiting from residual connections and, in the former case, temporal recurrence through ConvLSTM. UDeepONet3D is the weakest performer on both tasks, suggesting that the branch-trunk decomposition is poorly suited for dense 3D spatiotemporal mapping at this resolution. FNO3D, despite its strong theoretical foundations for operator learning, underperforms on saturation ($R^2=0.8608$), indicating that purely spectral methods struggle to accurately represent the sharp displacement fronts characteristic of CO$_2$ plume evolution. This wide disparity in baseline performance demonstrates that the selection of an appropriate architecture is not a trivial decision and can lead to order-of-magnitude differences in prediction error, further motivating the need for an intelligent, physics-aware selection mechanism as provided by \textsc{AutoSurrogate}.

The AutoML baselines further underscore the difficulty of the saturation prediction task and the value of physics-aware reasoning. Despite having the same total trial budget (120 trials) and access to the same model zoo, all three AutoML methods perform substantially worse than \textsc{AutoSurrogate} on saturation: the best AutoML result (Optuna TPE, $R^2=0.8915$) falls 6.17 percentage points below \textsc{AutoSurrogate} ($R^2=0.9532$) and is below every hand-tuned baseline except UDeepONet3D and FNO3D. Random Search is catastrophically poor ($R^2=0.7593$), demonstrating that blind search over the joint architecture--hyperparameter space routinely selects configurations that fail to capture the sharp displacement fronts characteristic of CO$_2$ plume evolution. On pressure, the gap is smaller ($R^2$ of 0.9804--0.9922 vs.\ 0.9959), consistent with the observation that pressure fields are smooth and nearly linear, making them amenable to nearly any reasonable configuration. Notably, all three AutoML methods underperform the best hand-tuned baselines on both tasks, indicating that the standard HPO search paradigm---which treats architecture and hyperparameters as a flat combinatorial space without domain knowledge---is fundamentally limited for physics-governed surrogate modeling. These results confirm that the performance advantage of \textsc{AutoSurrogate} stems not from a larger search budget but from its ability to reason about physics--architecture compatibility and to recover from training failures through self-correction.

\subsection{Search Efficiency}
\label{sec:search_efficiency}

To isolate the efficiency of each search strategy from the cost of final model training, \autoref{tab:search_time} compares the search overhead of each method---defined as the time spent on hyperparameter optimization trials (5 epochs per trial for all methods) plus, for \textsc{AutoSurrogate}, the additional time for data profiling, model selection, and LLM reasoning. This is a fair comparison because all methods use the same per-trial training protocol (5 epochs, 50 training batches, 20 validation batches); the only difference is how each method decides \emph{which} configurations to evaluate.

For saturation, \textsc{AutoSurrogate}'s search overhead (32.1 minutes for HPO plus ${\sim}$10 minutes for LLM-driven profiling and model selection) is substantially lower than all three AutoML baselines (50.7--93.6 minutes), despite achieving dramatically higher final $R^2$ (0.9532 vs.\ 0.7593--0.8915). This efficiency gain arises because the \textit{Model Selection Agent} pre-filters the architecture space: rather than distributing 120 trials uniformly across all eight architectures, \textsc{AutoSurrogate} focuses its trials on two physically motivated candidates (ResUNet3D and UNet3D), avoiding expensive trials on architectures that are poorly suited for sharp-front saturation dynamics. For pressure, \textsc{AutoSurrogate} ran 120 trials across 8 HPO rounds, resulting in a search overhead of 55.5 minutes, which falls within the range of AutoML's budgets (45.6--85.7 minutes). Between 25\% and 53\% of AutoML trials produce non-finite gradient norms and are simply discarded without informing subsequent search decisions; \textsc{AutoSurrogate}'s self-correction mechanism, in contrast, diagnoses these failures and adapts the search space accordingly.

\begin{figure*}[pos=htbp]
    \centering
    \includegraphics[width=\textwidth]{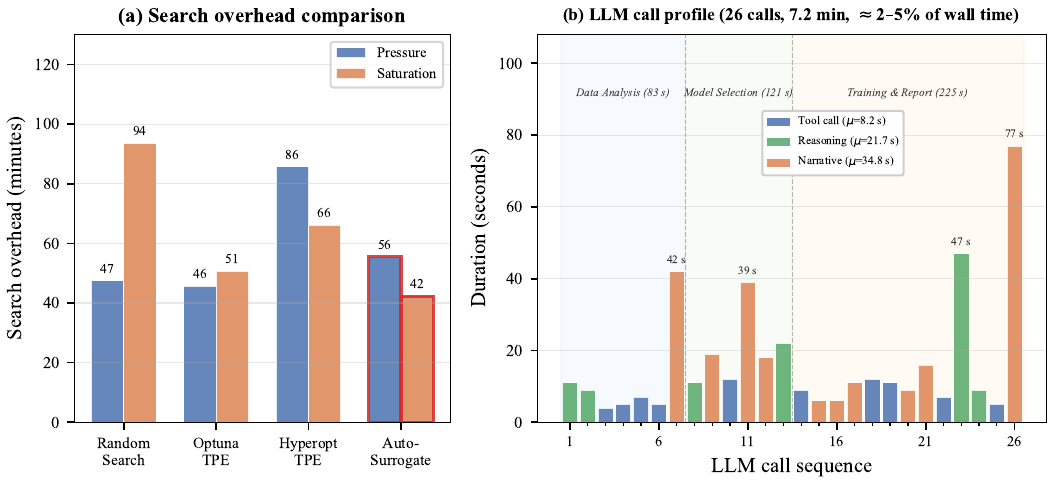}
    \caption{Search efficiency analysis. (a)~Search overhead comparison: time spent on HPO trials (and, for \textsc{AutoSurrogate}, data profiling + LLM reasoning). For saturation, \textsc{AutoSurrogate}'s LLM-guided search is faster than all AutoML methods because it focuses trials on two pre-selected architectures. (b)~LLM call profile for a representative pipeline run (26 calls, 7.2\,min total). Each bar represents one LLM invocation, colored by type: tool calls ($\mu{=}8.2$\,s), reasoning ($\mu{=}21.7$\,s), and narrative generation ($\mu{=}34.8$\,s). Background shading marks the three pipeline phases. Across five runs, total LLM inference time is 5.8--8.2\,min (${\sim}$2--5\% of wall time), confirming that the framework's ``intelligence'' is nearly free in computational terms.}
    \label{fig:search_efficiency}
\end{figure*}

\autoref{fig:search_efficiency} provides a comprehensive view of the efficiency tradeoff. Panel~(a) compares the search overhead across methods. Panel~(b) profiles the LLM invocations for a representative pipeline run (26 calls, 7.2\,min total), categorized into three types: \emph{tool calls} (data configuration, model estimation; $\mu{=}8.2$\,s), \emph{reasoning} (planning and diagnosis; $\mu{=}21.7$\,s), and \emph{narrative generation} (summaries and reports; $\mu{=}34.8$\,s). The calls naturally decompose into three pipeline phases---data analysis (83\,s), model selection (121\,s), and training/reporting (225\,s). Across five pipeline runs (116 LLM calls in total), the aggregate LLM inference time ranges from 5.8 to 8.2 minutes, representing only 2--5\% of pipeline wall time. All LLM inference is performed on a locally deployed model at negligible cost. In other words, the intelligence provided by the LLM is nearly ``free''.

\begin{table*}[pos=htbp]
\centering
\caption{Search efficiency comparison. ``Search overhead'' reports HPO trial time for AutoML methods; for \textsc{AutoSurrogate}, it additionally includes data profiling, model selection, and LLM reasoning. ``Best $R^2$'' is the test-set metric of the final deployed model after full training.}
\label{tab:search_time}
\small
\begin{tabular}{l cc cc}
\toprule
\multirow{2}{*}{Method} & \multicolumn{2}{c}{Search Overhead (min)} & \multicolumn{2}{c}{Best $R^2$} \\
\cmidrule(lr){2-3} \cmidrule(lr){4-5}
 & Pressure & Saturation & Pressure & Saturation \\
\midrule
All baselines (brute-force)  & \multicolumn{2}{c}{68.7 GPU-hours (16 models to convergence)} & 0.9972 & 0.9359 \\
\midrule
Random Search (120 trials)     & 47.4 & 93.6 & 0.9820 & 0.7593 \\
Optuna TPE (120 trials)        & \textbf{45.6} & 50.7 & 0.9804 & 0.8915 \\
Hyperopt TPE (120 trials)      & 85.7 & 66.1 & 0.9922 & 0.8848 \\
\midrule
\textsc{AutoSurrogate}  & 55.5 & \textbf{42.3} & \textbf{0.9972} & \textbf{0.9532} \\
\bottomrule
\end{tabular}
\end{table*}

\subsection{Case Studies}
\label{sec:case_studies}

In this subsection, we examine two distinctive capabilities of \textsc{AutoSurrogate} that differentiate it from conventional AutoML approaches: interpretable decision-making and autonomous self-correction. These case studies demonstrate that the framework not only achieves competitive accuracy and efficiency, but also provides full traceability of its reasoning process and exhibits robustness to training failures.

\subsubsection{Decision-Making Tracing}
\label{sec:decision_trace}

% 二、AutoSurrogate的决策过程追踪
% 这是本文区别于普通AutoML的关键卖点。

A key differentiator of \textsc{AutoSurrogate} from black-box AutoML is the full interpretability and traceability of its decision process. Every stage of the pipeline produces a structured reasoning trace that is preserved in the report for post-hoc inspection. To demonstrate how the framework operates in practice, \autoref{tab:pipeline_trace} presents the complete end-to-end pipeline trace for the saturation task, which exercises all major pipeline stages including data profiling, model selection, HPO, training, and self-correction through architecture switching.

\begin{table*}[pos=htbp]
\centering
\caption{End-to-end pipeline trace for the saturation task. Each row corresponds to one stage of the autonomous pipeline, showing the agent's action, its reasoning based on observed data, and the quantitative outcome. The trace demonstrates the full decision chain from raw data profiling through self-correction to target achievement.}
\label{tab:pipeline_trace}
\footnotesize
\begin{tabular}{p{1.9cm} p{2.5cm} p{7.0cm} p{2.8cm}}
\toprule
Stage & Agent Action & Observation / Reasoning & Outcome \\
\midrule
\textit{Data Profiling} & Extract dataset statistics via profiling tools & ``1000 samples, spatial grid $80 \times 80 \times 20$, 31 timesteps. Saturation range $[0, 0.466]$, mean $=0.005$ (very sparse: $>99\%$ of voxels near zero). Pressure normalized ($\mu{=}208.14$, $\sigma{=}6.01$); saturation kept in native $[0,1]$ range.'' & Profile stored in shared context \\
\midrule
\textit{Model Selection} & Estimate GPU memory for 8 candidate architectures; select based on task physics & ``Residual skip connections suit 3D spatiotemporal mapping. ResUNet3D at \texttt{base\_channels=16}: ${\sim}$4.6M params, ${\sim}$609\,MB GPU memory at batch 1---appropriate for 700 training samples. Supports BCE auxiliary loss for sparse front reconstruction.'' & Selected: ResUNet3D \\
\midrule
\textit{HPO Round\,1} & Bayesian HPO (TPE + median pruner, 15 trials, 5 epochs/trial) & Search space: \texttt{base\_channels} $\in \{8,16,32,64\}$, \texttt{lr} $\in [10^{-5}, 10^{-2}]$, \texttt{batch\_size} $\in \{1,2,4\}$, $\lambda_{\mathrm{bce}} \in [10^{-4}, 0.1]$. Best trial: \texttt{bc=16}, \texttt{lr}$=6.27 \times 10^{-4}$, $\lambda_{\mathrm{bce}}=4.6 \times 10^{-4}$, \texttt{bs=4}. 6 trials completed, 9 pruned. & Best val loss: $8.73 \times 10^{-4}$ \\
\midrule
\textit{Training Round\,1} & Train ResUNet3D with best HPO config (30 epochs, early stopping patience 15) & Training converges; test evaluation on 200-sample hold-out set. & $R^2=0.9479$ (below 0.95 target) \\
\midrule
\textit{Self-Correction: Continuation} & Resume training for 10 additional epochs from last checkpoint & ``$R^2$ unchanged at 0.9479 after additional epochs---learning has plateaued. The model's capacity or hyperparameter regime cannot close the remaining gap.'' & Plateau confirmed; no improvement \\
\midrule
\textit{Self-Correction: Architecture Switch} & Switch to UNet3D; launch fresh HPO & ``ResUNet3D has saturated. Switching to UNet3D: simpler encoder--decoder without residual blocks may benefit from a different loss landscape under aggressive HPO.'' & New HPO initiated \\
\midrule
\textit{HPO Round\,2} (UNet3D) & Bayesian HPO (15 trials, 5 epochs/trial) & Discovers dramatically different configuration: \texttt{bc=32} ($2\times$ wider), \texttt{lr}$=7.02 \times 10^{-3}$ ($11\times$ Round\,1), $\lambda_{\mathrm{bce}}=0.037$ ($80\times$ Round\,1), \texttt{bs=2}. Higher BCE weight forces sharper front reconstruction. & Best val loss: $2.05 \times 10^{-3}$ \\
\midrule
\textit{Training Round\,2} & Train UNet3D with new HPO config (30 epochs) & UNet3D with optimized hyperparameters surpasses all baselines including RecurrentRUNet3D ($R^2{=}0.9359$, 7.3\,h). & $R^2=\mathbf{0.9532}$ (target met) \\
\bottomrule
\end{tabular}
\end{table*}

Several observations from \autoref{tab:pipeline_trace} merit discussion. First, the \textit{Data Profiling} stage reveals that saturation fields are extremely sparse (mean $=0.005$), which directly informs both the model selection rationale (architectures supporting BCE auxiliary loss are preferred) and the HPO search space (the BCE weight $\lambda_{\mathrm{bce}}$ is included as a tunable hyperparameter). Second, the \textit{Model Selection} stage demonstrates physics-aware reasoning that goes beyond simple performance ranking: the agent explicitly considers the relationship between architecture capacity, GPU memory constraints, and the number of available training samples---a form of reasoning that traditional AutoML, which treats all prediction targets uniformly, cannot provide. Third, the HPO configurations discovered in Round~1 and Round~2 are strikingly different: the learning rate increases by $11\times$ and the BCE weight by $80\times$ when switching from ResUNet3D to UNet3D, indicating that optimal hyperparameters are strongly architecture-dependent and cannot be transferred between models. Finally, the self-correction sequence demonstrates that UNet3D---which ranks only fifth among the eight baselines when trained with default hyperparameters ($R^2=0.9273$)---surpasses all baselines when paired with HPO-discovered configurations, underscoring that architecture selection and hyperparameter optimization are jointly important.

\autoref{tab:hpo_comparison} compares the key hyperparameters discovered by \textsc{AutoSurrogate}'s Bayesian HPO against the hand-tuned baseline defaults. For pressure, the HPO discovered a learning rate of $4.12 \times 10^{-3}$, approximately 41 times higher than the baseline default of $10^{-4}$, and doubled the network width from 16 to 32 base channels. For saturation, the discovered learning rate was $7.02 \times 10^{-3}$ (70 times higher), and the BCE weight $\lambda_{\mathrm{bce}}$ was set to 0.037, approximately 3.7 times the baseline default of 0.01, suggesting that the default substantially underweights the importance of front sharpness in the loss landscape. These large deviations from conventional defaults demonstrate that the ``standard'' hyperparameters commonly adopted in the literature are far from optimal for this dataset, and that automated Bayesian search can uncover substantially better configurations.

\begin{table}[pos=htbp]
\centering
\caption{Key hyperparameters discovered by \textsc{AutoSurrogate} compared with hand-tuned baseline defaults.}
\label{tab:hpo_comparison}
\small
\begin{tabular}{l c c c}
\toprule
\multirow{2}{*}{Hyperparameter} & Baseline & \multicolumn{2}{c}{\textsc{AutoSurrogate}} \\
\cmidrule(lr){3-4}
 & Default & Pressure & Saturation \\
\midrule
Architecture       & (fixed)      & ResUNet3D  & UNet3D \\
Base channels      & 16           & 32         & 32 \\
Normalization      & Group        & Instance   & Instance \\
Learning rate      & $10^{-4}$    & $4.12 \times 10^{-3}$ & $7.02 \times 10^{-3}$ \\
Weight decay       & $10^{-5}$    & $3.0 \times 10^{-4}$  & $1.8 \times 10^{-5}$ \\
Batch size         & 1            & 4          & 2 \\
$\lambda_{\mathrm{bce}}$  & 0.01  & --         & 0.037 \\
\bottomrule
\end{tabular}
\end{table}

\subsubsection{Self-Correction Mechanism}
\label{sec:self_correction_case}

% 三、自纠正机制的验证
% 这是论文最有特色的部分，需要专门设计实验来展示它确实有效。具体可以包括：在训练过程中是否触发了self-correction、触发了哪种类型（continuation / stability restart / architecture switch）、纠正前后的指标变化。如果在当前case中没有自然触发instability，可以考虑人为构造一个更困难的设置（比如减少训练样本、加大permeability对比度）来激发失败模式，展示系统的恢复能力。

The self-correction mechanism described in \autoref{sec:selfcorrect} is validated through two representative pipeline runs that together exercise all four recovery strategies of \autoref{alg:selfcorrect}. The correction timelines are summarized in \autoref{tab:self_correction}. Note that we choose to specify different targets with natural language instructions.

\textit{Saturation task -- successful recovery through architecture switching.} In the saturation pipeline run, the user instruction specifies a target of $R^2>0.95$. The \textit{Model Selection Agent} initially selects ResUNet3D, and after HPO and 30 epochs of training, the model achieves $R^2=0.9479$, falling short of the target. The \textit{HPO \& Training Agent} first applies Strategy~(i), continuation, resuming training for 10 additional epochs, but the model plateaus with no improvement. The agent then escalates to Strategy~(iii), architecture switching, selecting UNet3D as the next candidate from the model registry. A fresh HPO search is launched for UNet3D, discovering a substantially different configuration: base channels of 32, a learning rate of $7.02 \times 10^{-3}$, and a BCE weight of 0.037. After training with these optimized hyperparameters, the model achieves $R^2=0.9532$, exceeding the target. Notably, UNet3D with default hyperparameters ranks only fifth among the eight baselines for saturation ($R^2=0.9273$), yet with HPO-optimized parameters it surpasses all baselines, including RecurrentRUNet3D ($R^2=0.9359$), which is the strongest baseline but requires 7.3 hours of training. This result demonstrates that architecture selection and hyperparameter optimization are jointly important: the right architecture with wrong hyperparameters can underperform, while a seemingly inferior architecture with optimized hyperparameters can excel.

\textit{Pressure task -- persevering global best checkpoint for final deployment.} In the pressure run, to achieve a performance that is as high as possible, after the initial ResUNet3D training achieves $R^2=0.9972$, the agent enters a sustained self-correction process spanning 10 rounds across six different architectures. During this process, eight numerical instability events (non-finite gradient norms) are detected and handled: the agent progressively tightens stability constraints (reducing the learning rate upper bound from default to $5 \times 10^{-4}$ and gradient clipping from 1.0 to 0.25) and cycles through UDeepONet3D, CNNTransformer3D, EDConvLSTM3D, UFNO3D, and RecurrentRUNet3D. Each architecture encounters either numerical instability or fails to surpass the global best. Upon exhausting the retry budget, the framework activates Strategy~(iv), global-best fallback, deploying the best checkpoint found across the entire search trajectory: the initial ResUNet3D model at $R^2=0.9972$.

Together, these two case studies validate the complete self-correction hierarchy. The self-correction mechanism is essential for practical deployment because real-world datasets, particularly those involving highly nonlinear multiphase flow, frequently trigger training failures that cannot be anticipated at design time. Unlike human-in-the-loop debugging, where corrections depend on individual expertise and are rarely systematized, \textsc{AutoSurrogate}'s self-correction actions are systematic, fully logged, and reproducible, enabling practitioners to audit every recovery decision after the run.

\begin{table*}[pos=htbp]
\centering
\caption{Self-correction timelines for two representative \textsc{AutoSurrogate} pipeline runs. Strategy labels refer to \autoref{alg:selfcorrect}: (i)~continuation, (ii)~stability-constrained restart, (iii)~architecture switching, (iv)~global-best fallback.}
\label{tab:self_correction}
\footnotesize
\begin{tabular}{l l l l c l}
\toprule
Task & Round & Strategy & Architecture & $R^2$ & Status \\
\midrule
\multicolumn{6}{l}{\textit{Saturation (target: make $R^2$ higher than 0.95)}} \\
\midrule
Sat. & 1   & Initial HPO + Train     & ResUNet3D & 0.9479 & Below target \\
Sat. & 1b  & (i) Continuation        & ResUNet3D & 0.9479 & Plateau \\
Sat. & 2   & (iii) Arch.\ switch + HPO & UNet3D  & --     & Training \\
Sat. & 2b  & (i) Continuation        & UNet3D    & --     & Improving \\
Sat. & Final & --                    & UNet3D    & \textbf{0.9532} & Target met \\
\midrule
\multicolumn{6}{l}{\textit{Pressure (target: make $R^2$ as high as possible)}} \\
\midrule
Pres. & 1    & Initial HPO + Train      & ResUNet3D      & 0.9972 & Below target \\
Pres. & 2    & (ii) Stability restart   & ResUNet3D      & 0.9972 & No improvement \\
Pres. & 3    & (ii) Tighten + restart   & ResUNet3D      & --     & Instability \\
Pres. & 4--6 & (iii) Arch.\ switches    & 3 architectures$^\dagger$  & --     & Instability \\
Pres. & 7    & (iii) Arch.\ switch      & UFNO3D         & --     & Runtime error \\
Pres. & 8--10 & (iii)+(ii) Tighten      & RecurrentRUNet3D & --   & Instability \\
Pres. & Final & (iv) Global-best fallback & ResUNet3D     & \textbf{0.9972} & Budget exhausted \\
\bottomrule
\multicolumn{6}{l}{\footnotesize $^\dagger$UDeepONet3D $\to$ CNNTransformer3D $\to$ EDConvLSTM3D.}
\end{tabular}
\end{table*}

% case study: use AS to HPO for a sturcture

% 六、可视化
% 实验部分应包含若干典型样本的场可视化：选取2-3个测试样本，展示ground truth与AutoSurrogate预测的pressure/saturation场在若干代表性时间步的对比，以及逐像素误差图。这对于subsurface领域的读者来说几乎是必备的。

\subsection{Prediction Visualization}
\label{sec:visualization}

To examine the field-level behavior of the models, we visualize two representative saturation cases (Samples~822 and~851) and two representative pressure cases (Samples~820 and~815) at the final time step (30 years). For each sample, the top row compares predictions from four selected baselines and \textsc{AutoSurrogate} against the ground truth, while the bottom row shows the corresponding absolute error maps.

As shown in \autoref{fig:saturation_vis}, for both Samples~822 and~851, \textsc{AutoSurrogate} accurately reproduces the overall plume extent, the dominant plume geometry, and regions of high saturation. In contrast, several baseline models, such as CNNTransformer3D and UDeepONet3D, exhibit noticeably larger errors. The residual errors of \textsc{AutoSurrogate} are primarily concentrated near the plume boundaries, where the saturation field changes rapidly. The discrepancies may also be attributed to limitations in the available training data.

The pressure fields in \autoref{fig:pressure_vis} evolve more smoothly than saturation fields. Most baseline models capture the global pressure distribution reasonably well. However, predictions from CNNTransformer3D exhibits obvious errors near the injector in Sample~820, and UDeepONet3D shows more spatially distributed errors across both samples. RecurrentRUNet3D and UFNO3D achieve strong performance for pressure prediction.\textsc{AutoSurrogate} also provides a close visual match to the ground truth across all cases. Overall, these visualizations demonstrate that \textsc{AutoSurrogate} can accurately capture both the plume geometry in saturation fields and the smooth global structure of pressure fields.

\newlength{\predvispanelheight}
\setlength{\predvispanelheight}{0.089\textwidth}
\newcommand{\predvispanel}[2]{%
  \begin{minipage}[t]{0.145\textwidth}
    \centering
    \includegraphics[height=\predvispanelheight]{#1}\\[1pt]
    \scriptsize #2
  \end{minipage}%
}
\newcommand{\predvispanelcropped}[2]{%
  \begin{minipage}[t]{0.145\textwidth}
    \centering
    \includegraphics[trim=31.44bp 109.44bp 24.72bp 36.48bp,clip,height=\predvispanelheight]{#1}\\[1pt]
    \scriptsize #2
  \end{minipage}%
}
\newcommand{\predviscolorbar}[2]{%
  \begin{minipage}[t]{0.055\textwidth}
    \centering
    \includegraphics[height=\predvispanelheight]{#1}\\[1pt]
    \scriptsize #2
  \end{minipage}%
}
\newcommand{\predvisblank}{%
  \begin{minipage}[t]{0.145\textwidth}
    \centering
    \rule{0pt}{\predvispanelheight}\\[1pt]
    \scriptsize \phantom{Ground Truth}
  \end{minipage}%
}
\newcommand{\predvissample}[1]{%
  \makebox[\textwidth][l]{\scriptsize\textbf{Sample #1}}\par\vspace{2pt}%
}

\begin{figure*}[pos=htbp]
\centering
\predvissample{822}
\predvispanel{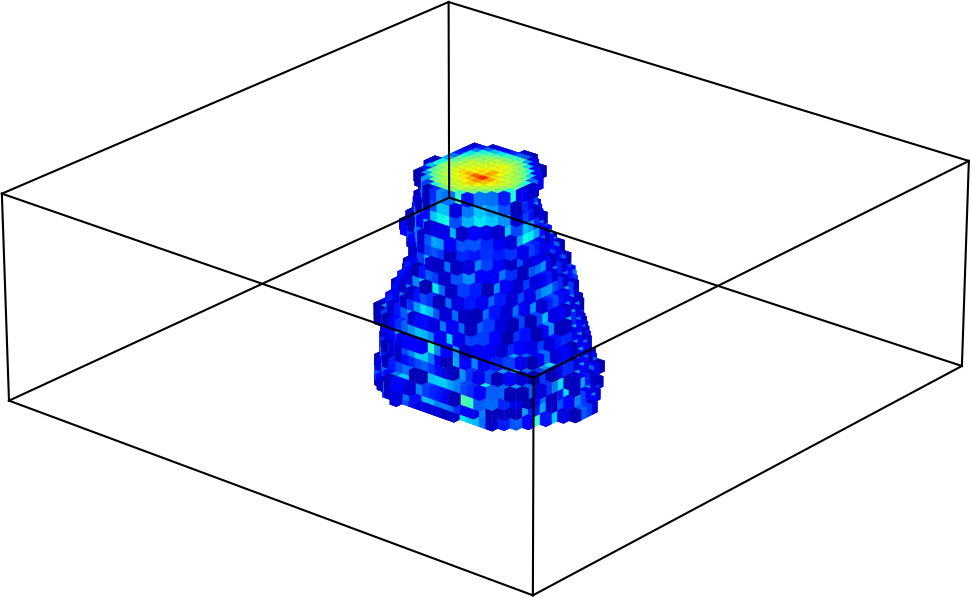}{RecurrentRUNet3D}\hfill
\predvispanel{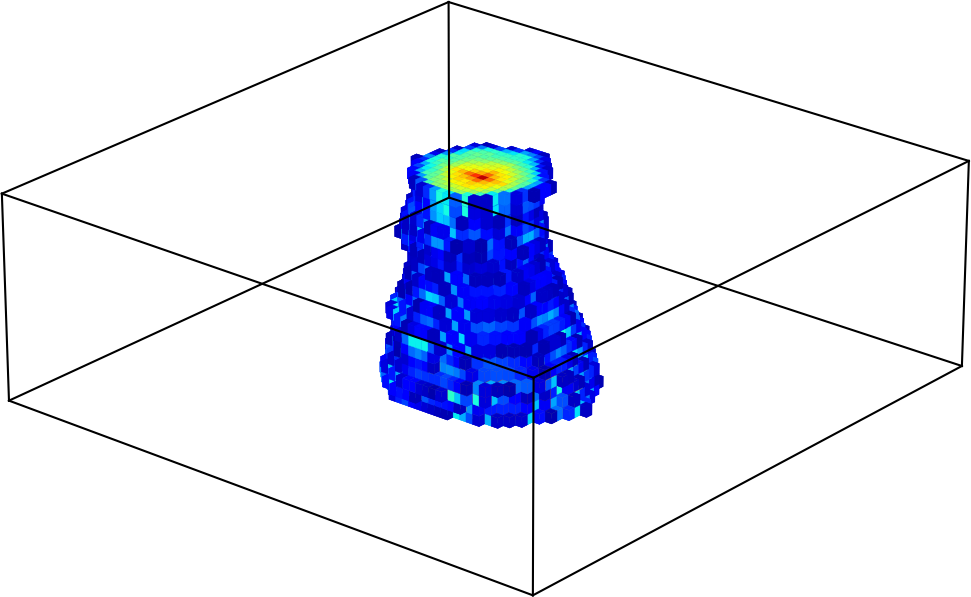}{UFNO3D}\hfill
\predvispanel{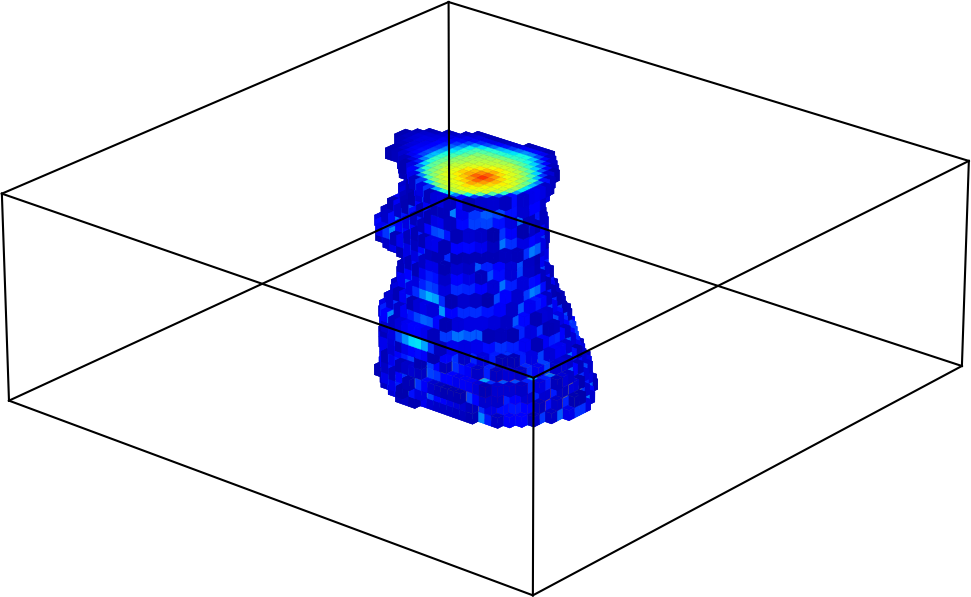}{CNNTransformer3D}\hfill
\predvispanel{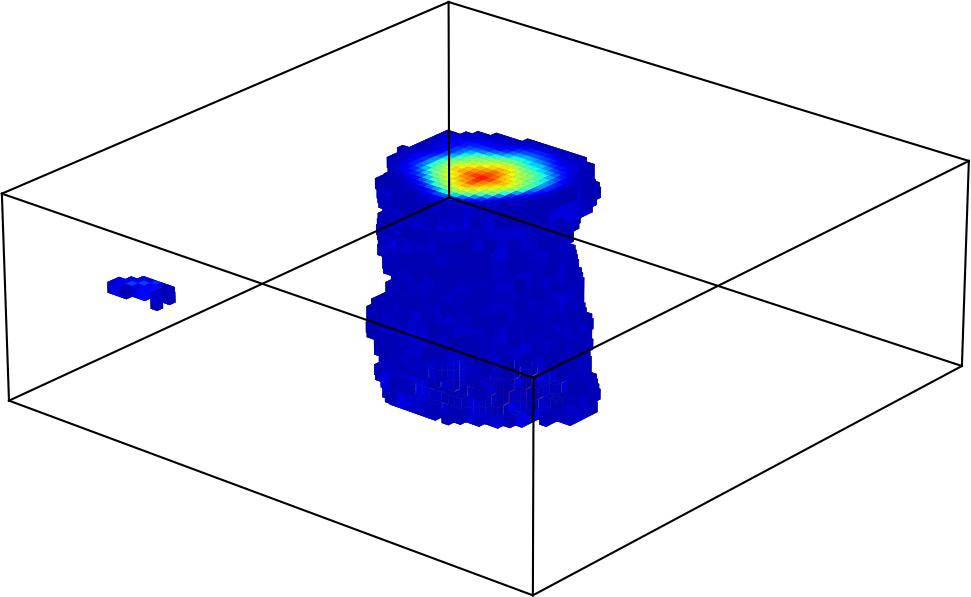}{UDeepONet3D}\hfill
\predvispanel{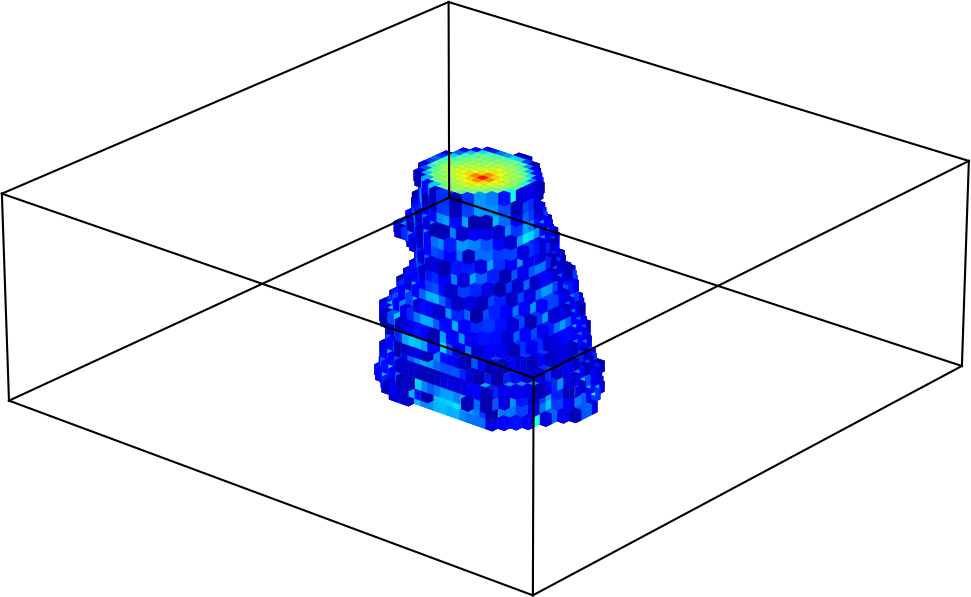}{\textsc{AutoSurrogate}}\hfill
\predvispanel{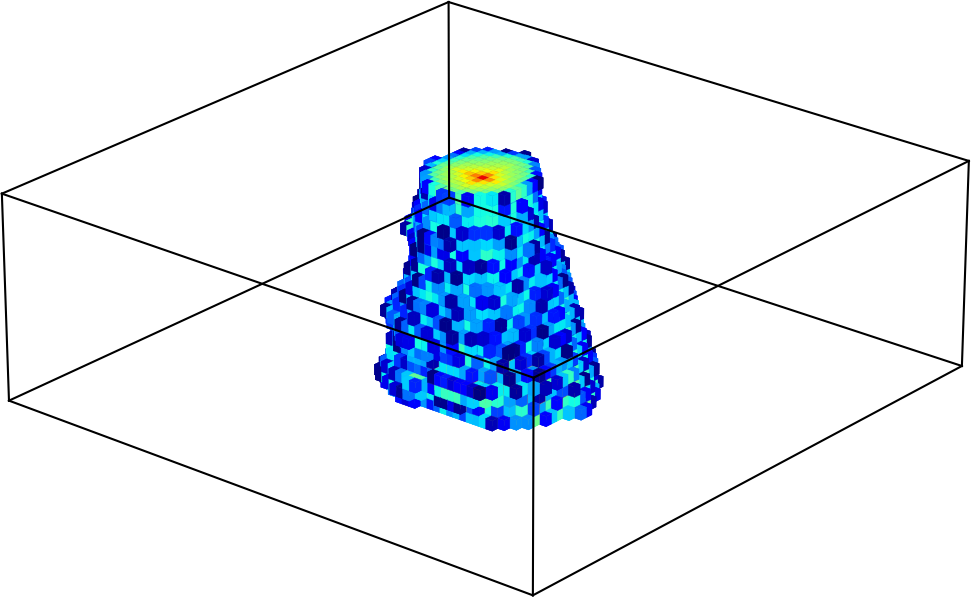}{Ground Truth}\hfill
\predviscolorbar{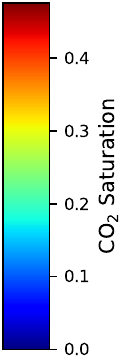}{$S$}

\vspace{4pt}

\predvispanel{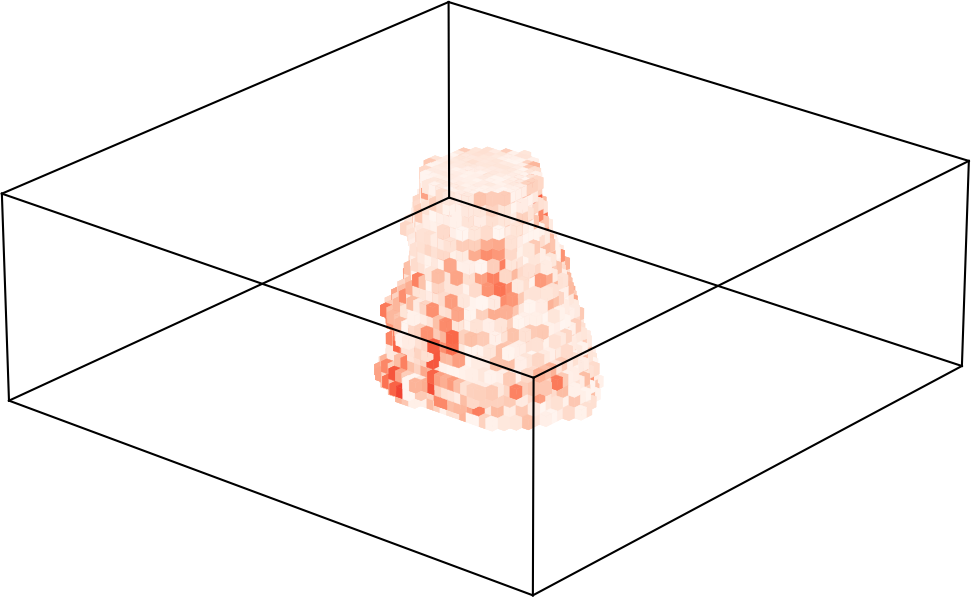}{RecurrentRUNet3D}\hfill
\predvispanel{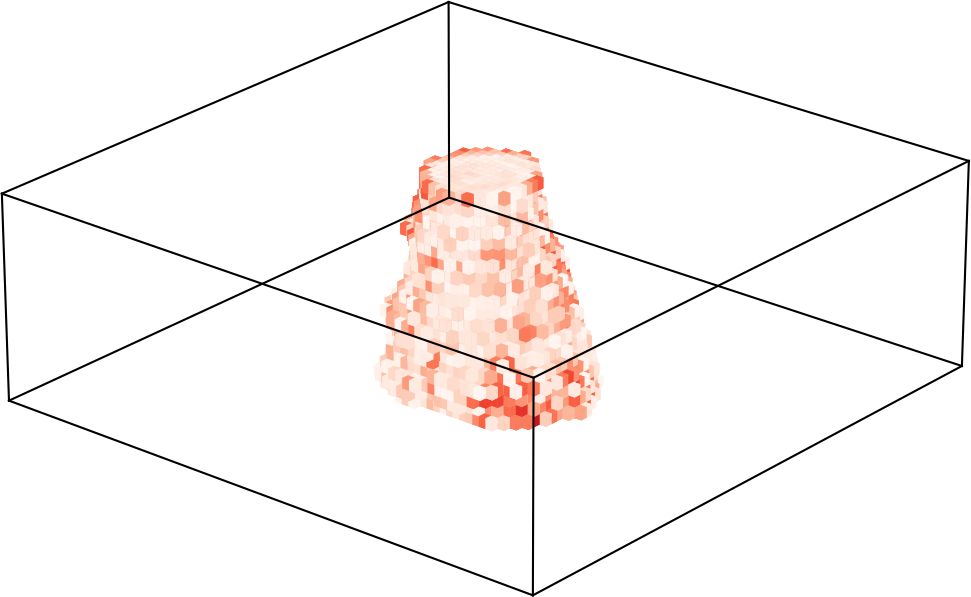}{UFNO3D}\hfill
\predvispanel{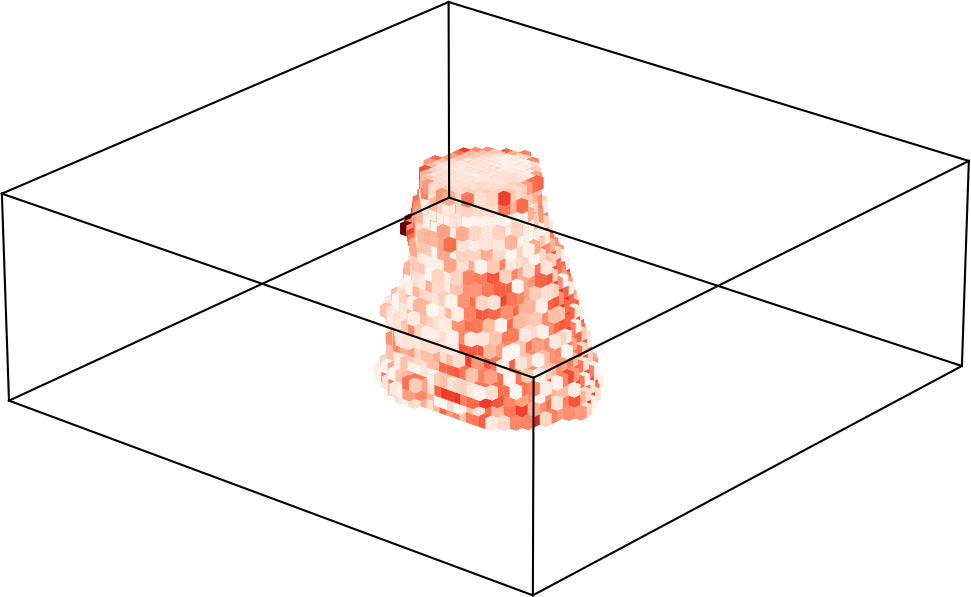}{CNNTransformer3D}\hfill
\predvispanel{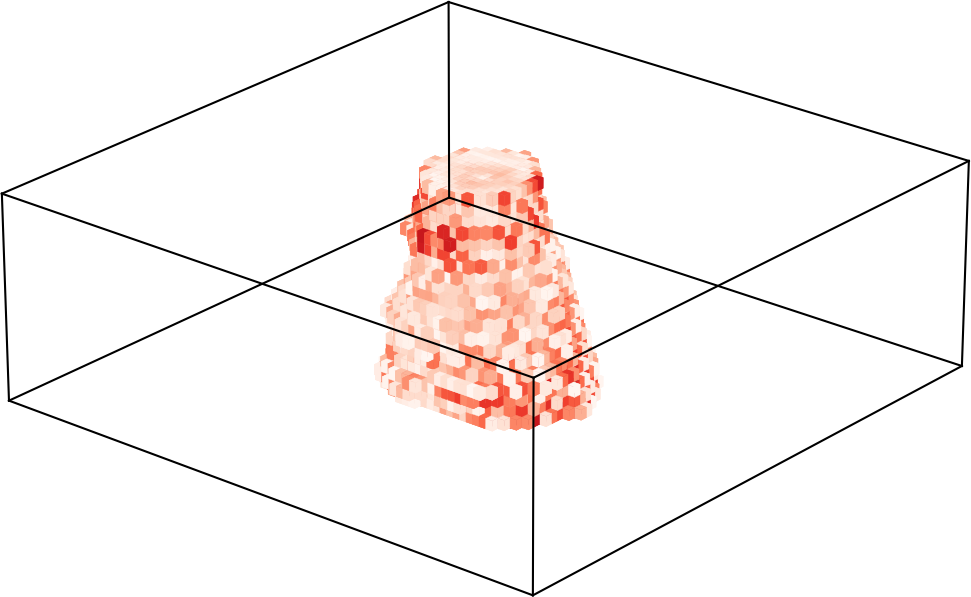}{UDeepONet3D}\hfill
\predvispanel{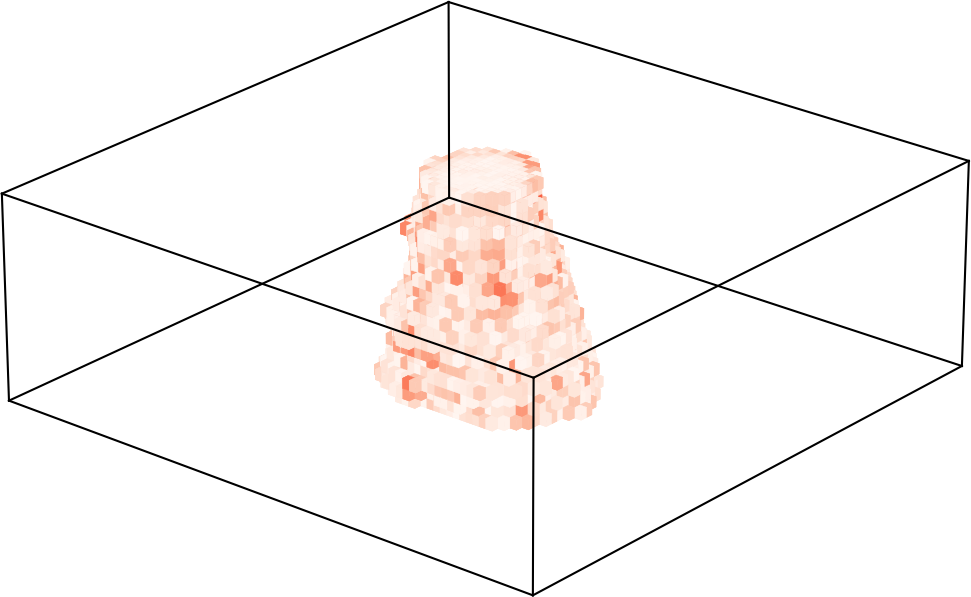}{\textsc{AutoSurrogate}}\hfill
\predvisblank\hfill
\predviscolorbar{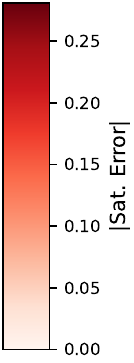}{$|\hat{S}-S|$}

\vspace{8pt}

\predvissample{851}
\predvispanelcropped{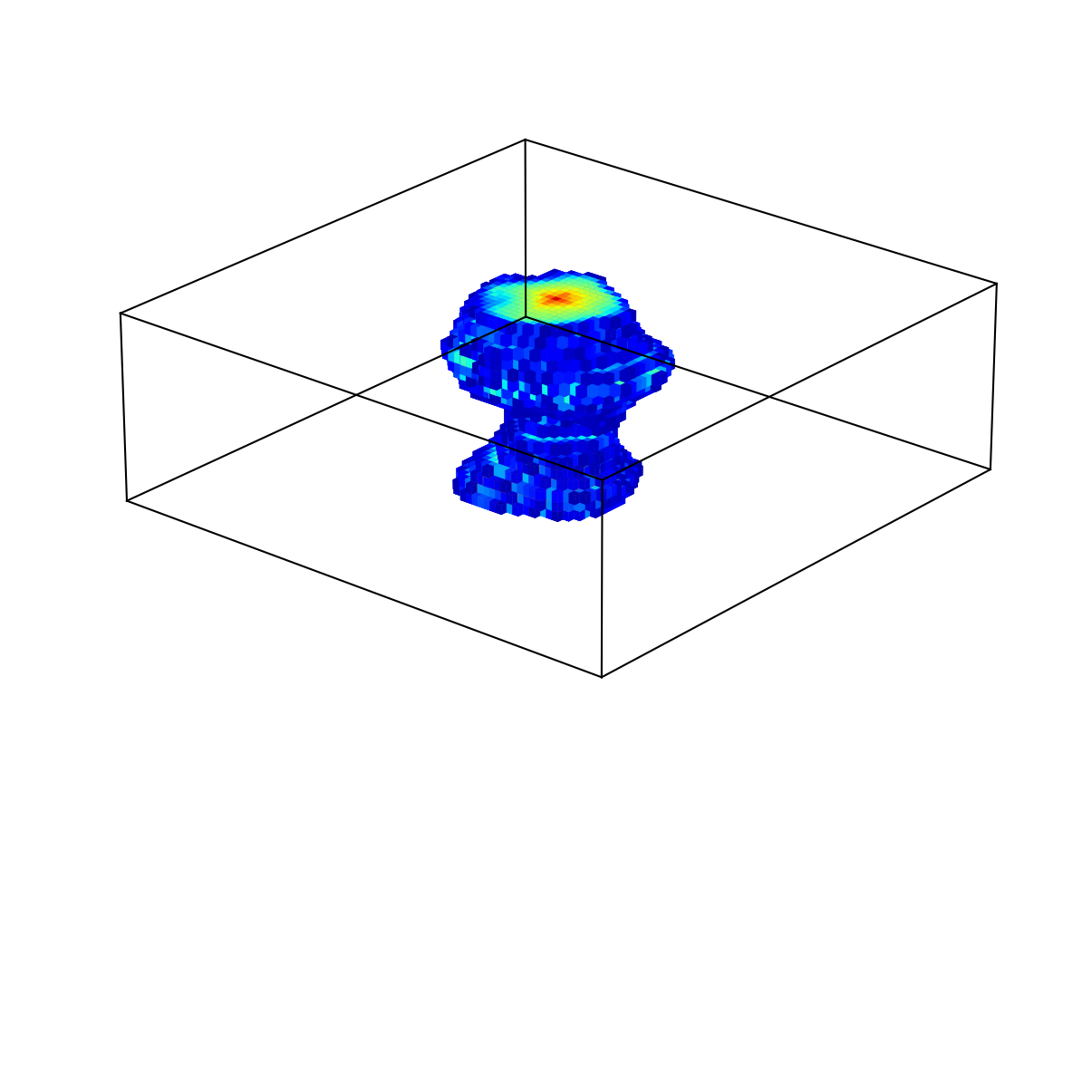}{RecurrentRUNet3D}\hfill
\predvispanelcropped{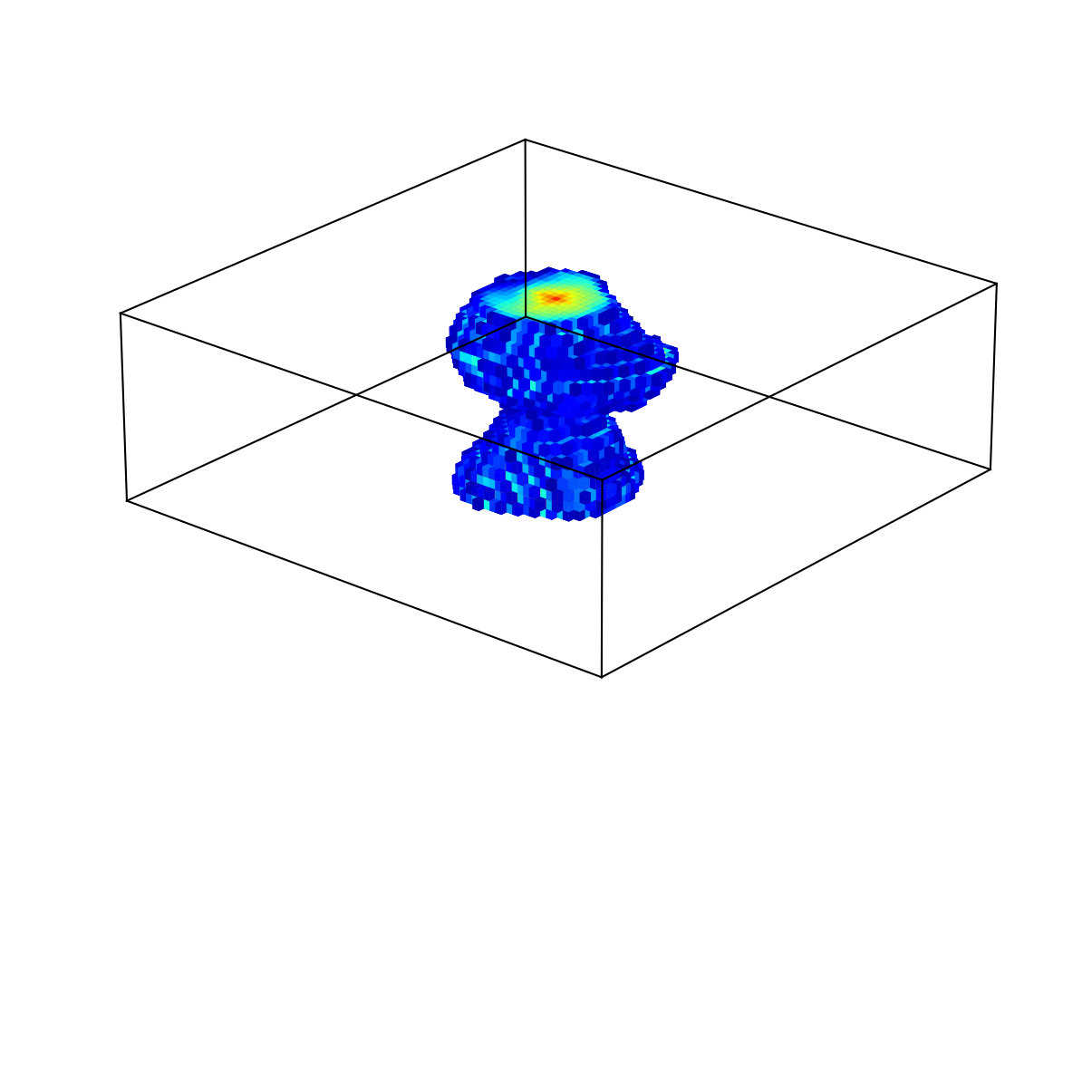}{UFNO3D}\hfill
\predvispanelcropped{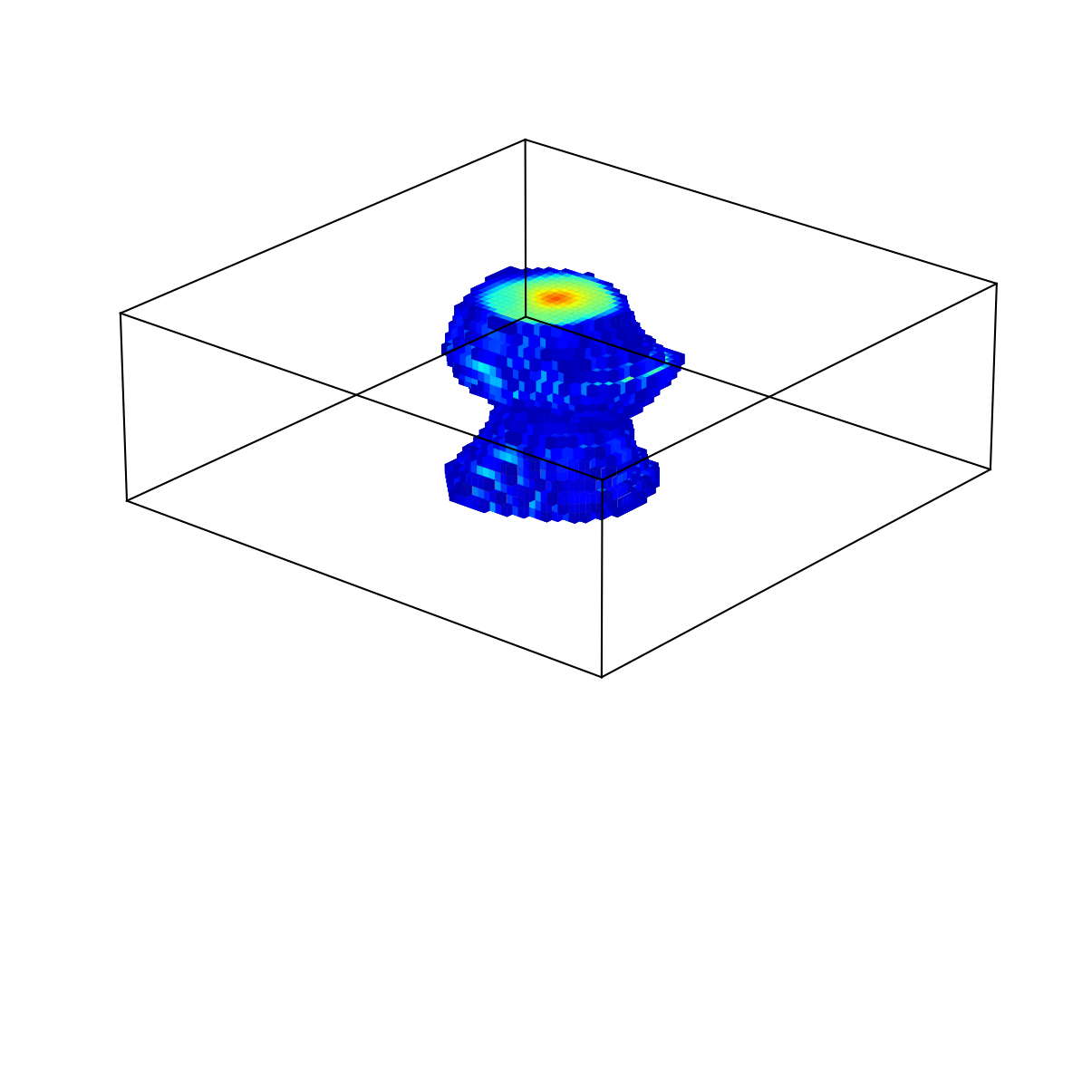}{CNNTransformer3D}\hfill
\predvispanelcropped{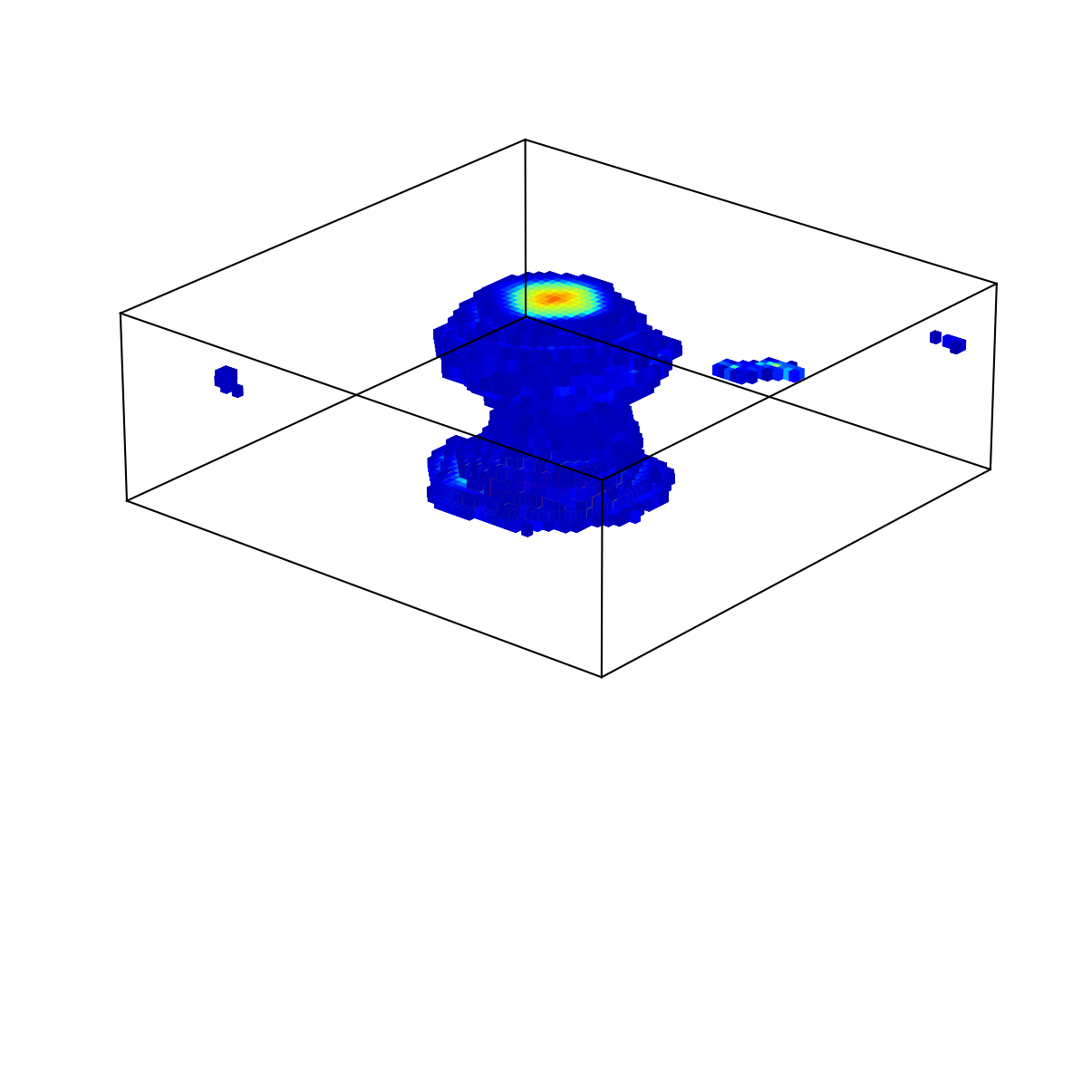}{UDeepONet3D}\hfill
\predvispanelcropped{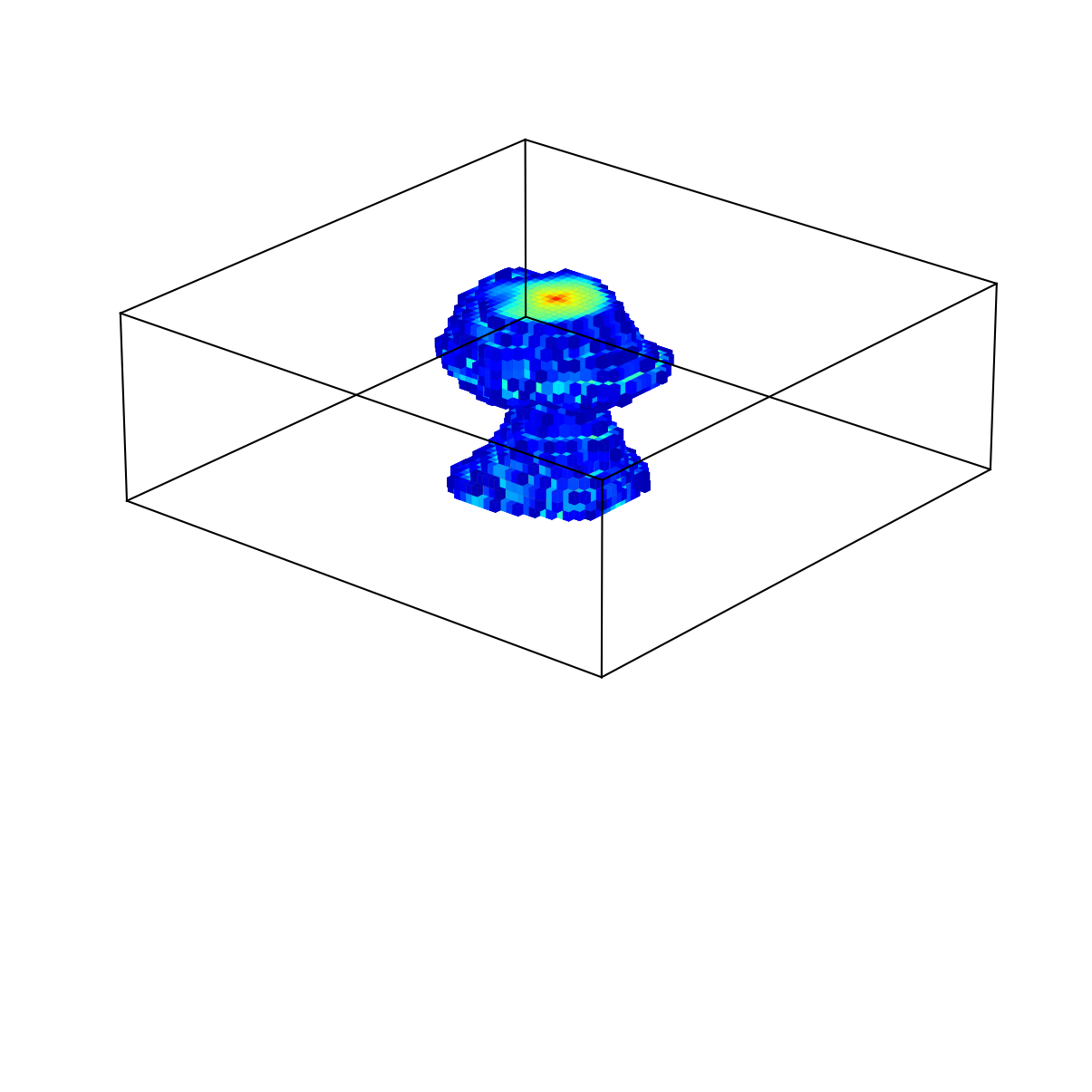}{\textsc{AutoSurrogate}}\hfill
\predvispanelcropped{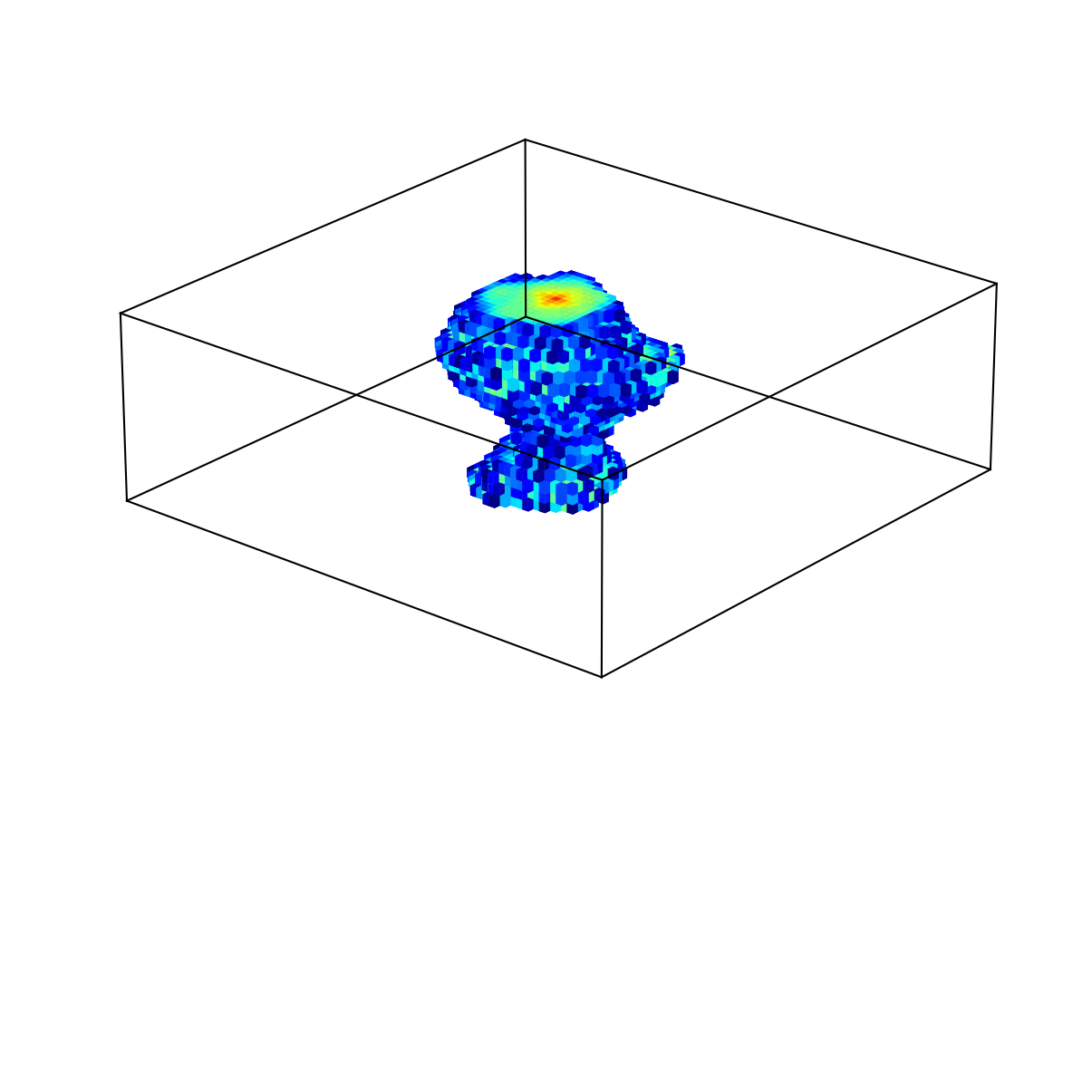}{Ground Truth}\hfill
\predviscolorbar{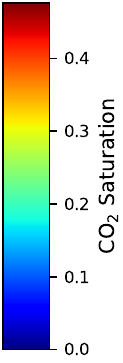}{$S$}

\vspace{4pt}

\predvispanelcropped{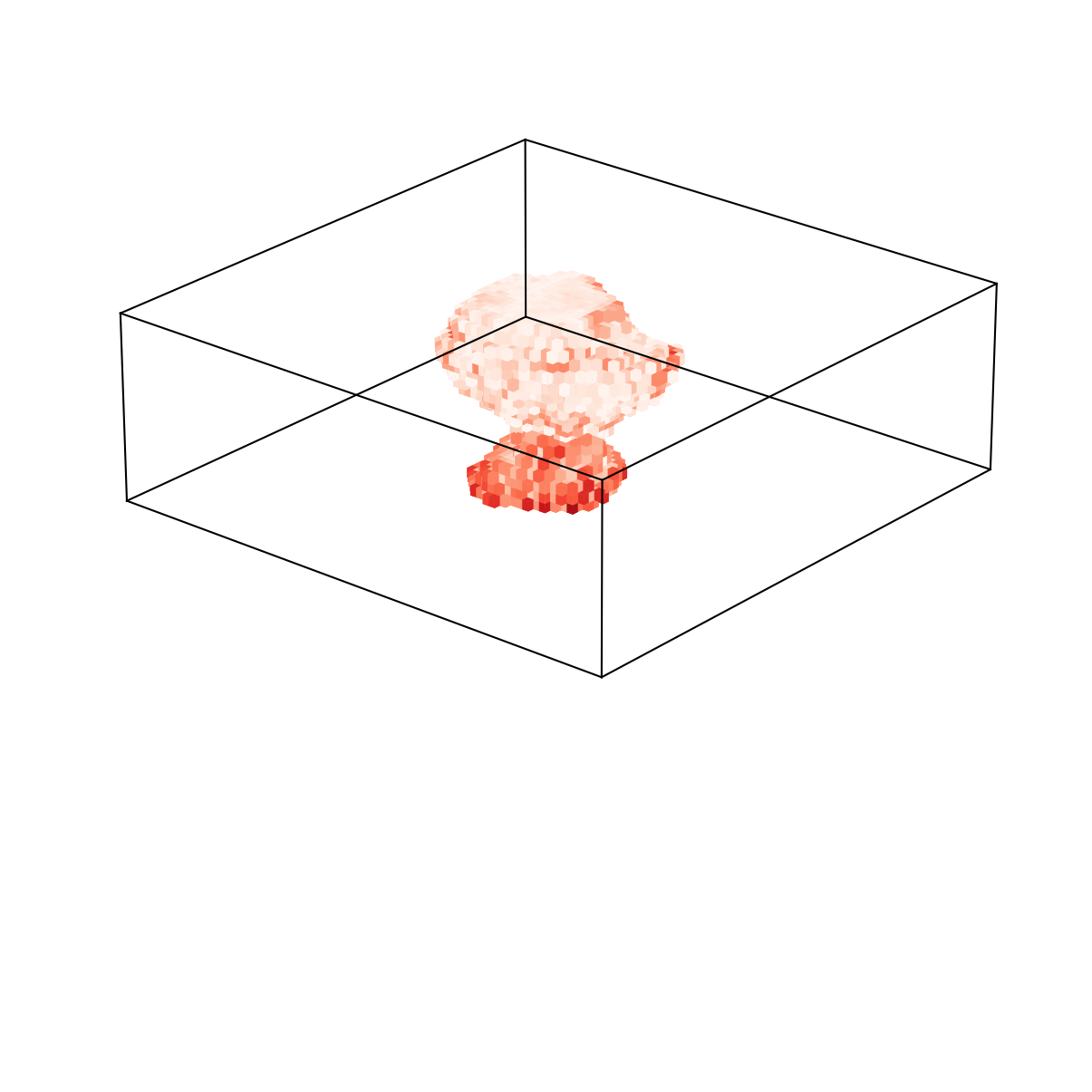}{RecurrentRUNet3D}\hfill
\predvispanelcropped{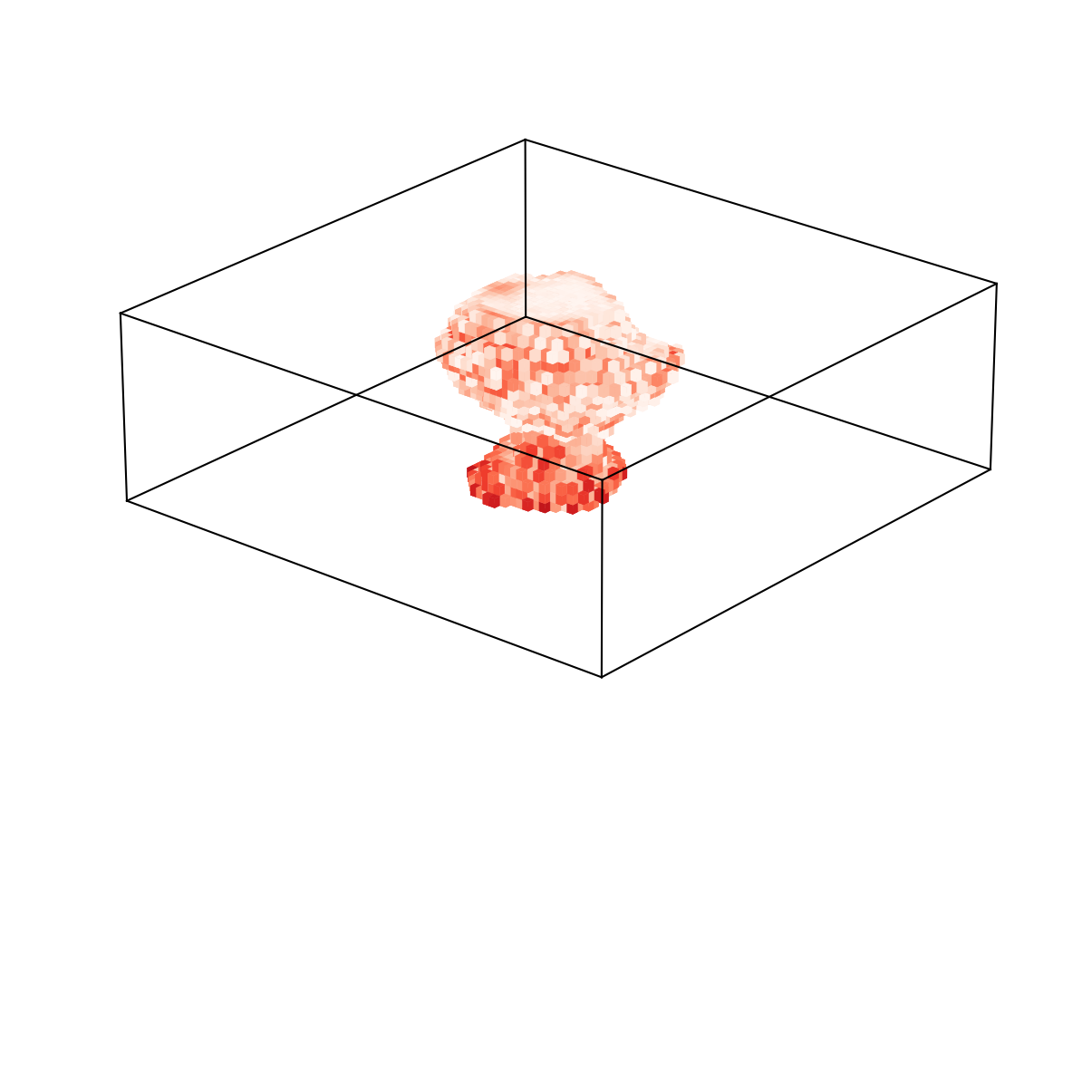}{UFNO3D}\hfill
\predvispanelcropped{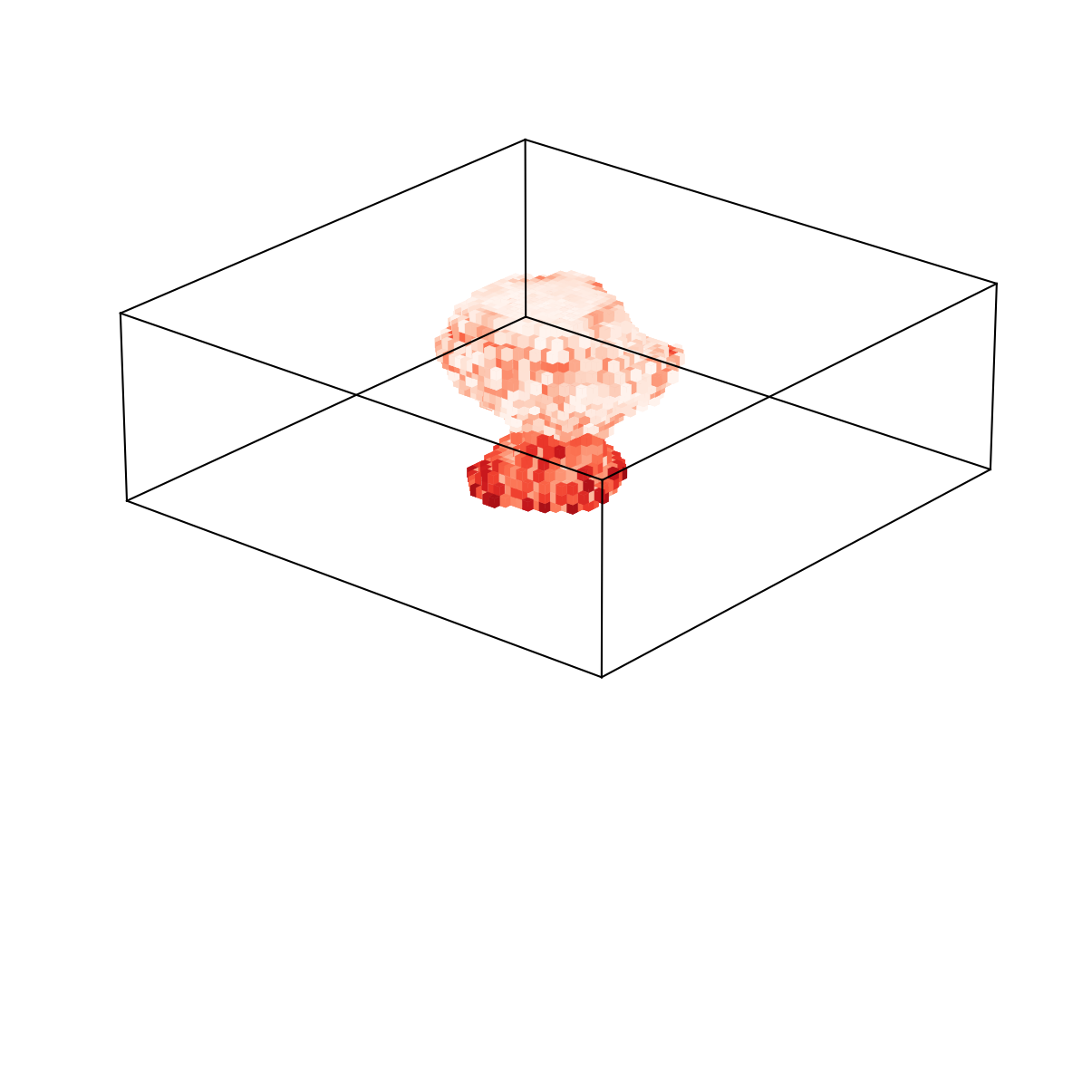}{CNNTransformer3D}\hfill
\predvispanelcropped{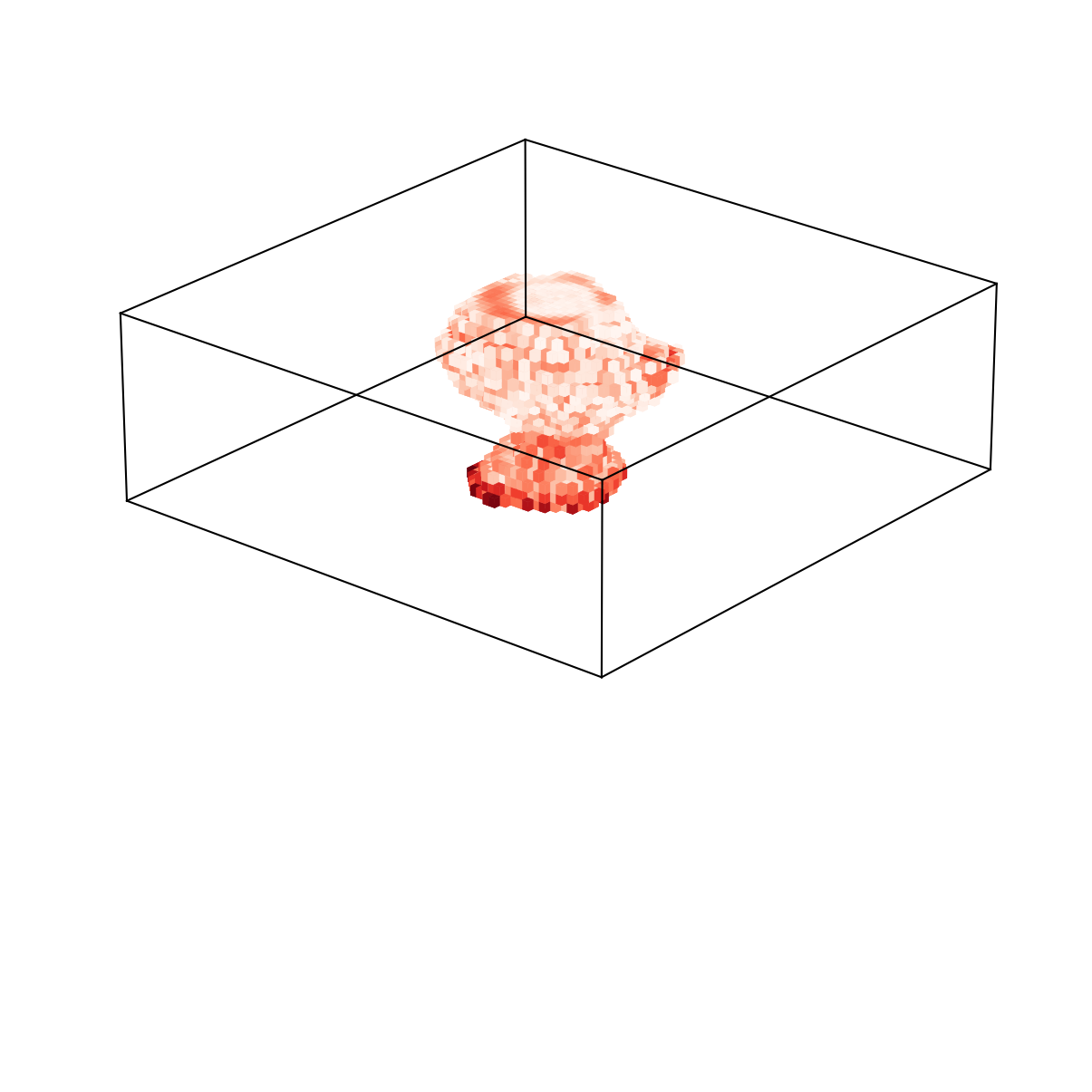}{UDeepONet3D}\hfill
\predvispanelcropped{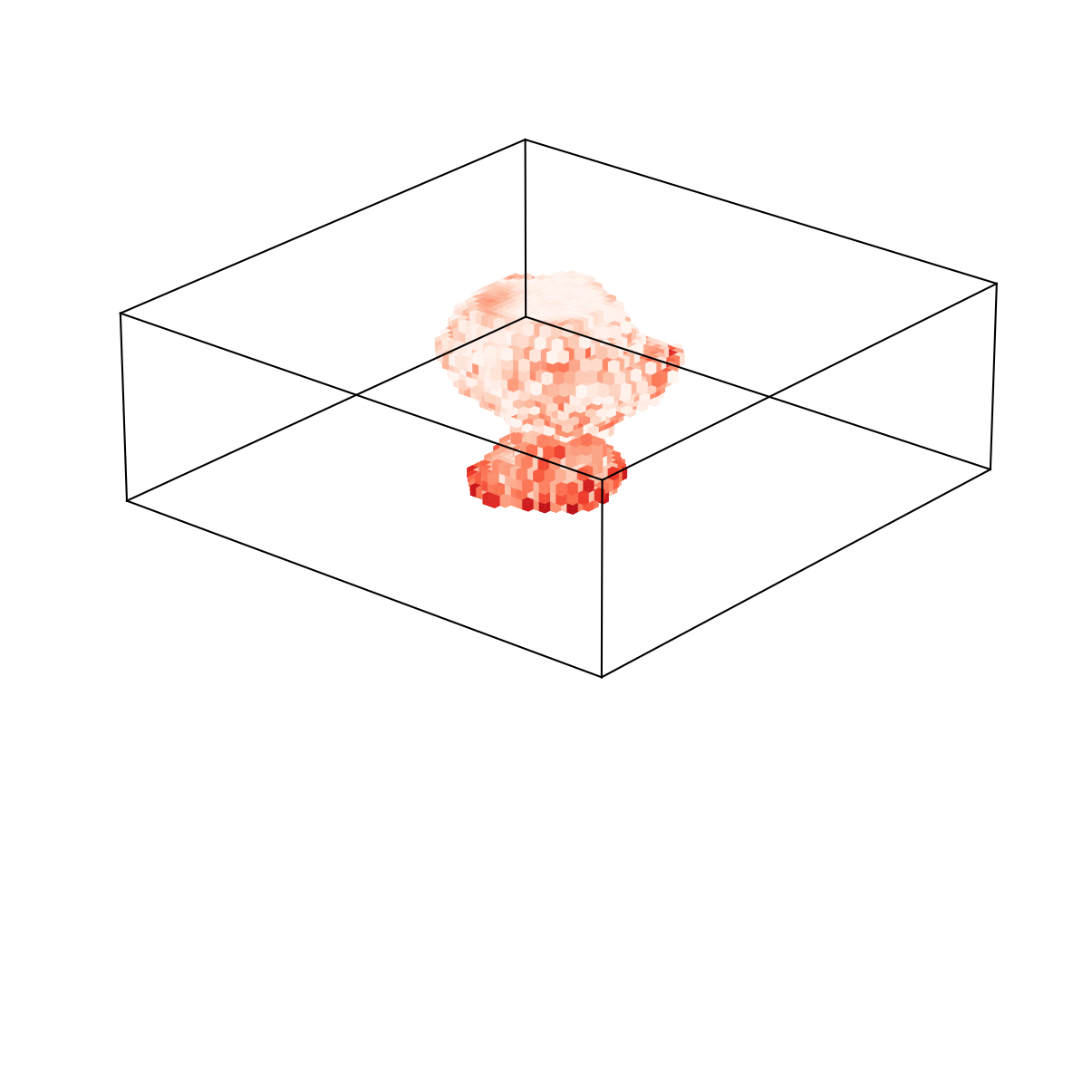}{\textsc{AutoSurrogate}}\hfill
\predvisblank\hfill
\predviscolorbar{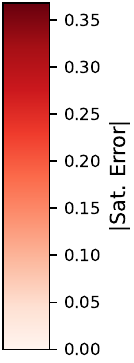}{$|\hat{S}-S|$}

\caption{CO$_2$ saturation predictions and absolute errors for Samples~822 and~851 at 30th year. Each sample uses two rows, and the rightmost column shows the shared colorbars.}
\label{fig:saturation_vis}
\end{figure*}

\begin{figure*}[pos=htbp]
\centering
\predvissample{820}
\predvispanel{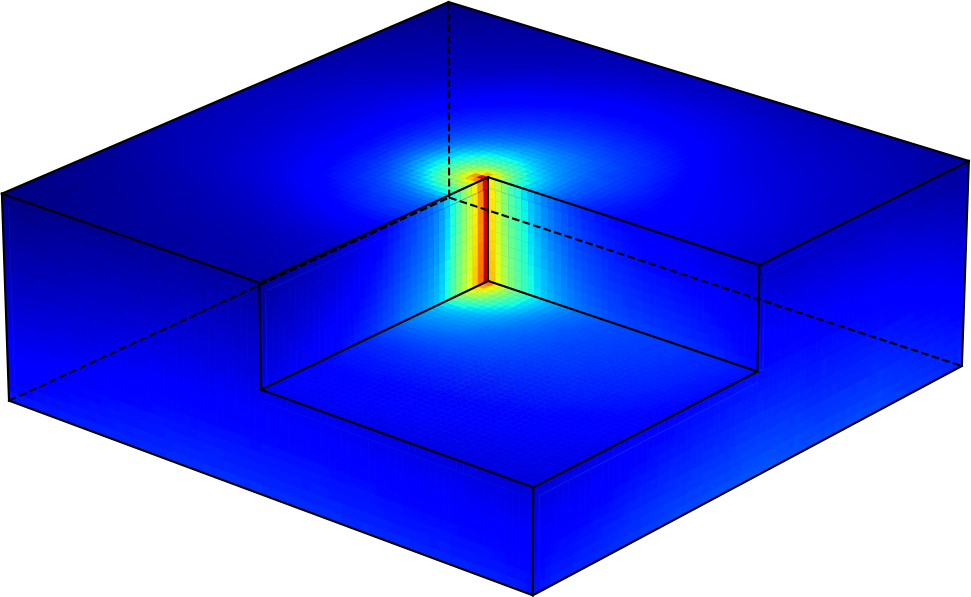}{RecurrentRUNet3D}\hfill
\predvispanel{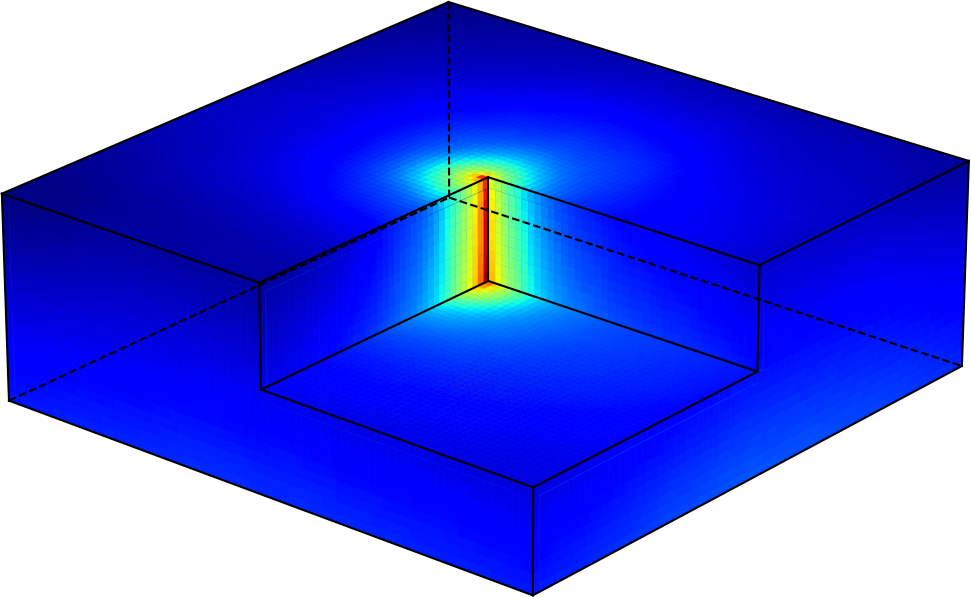}{UFNO3D}\hfill
\predvispanel{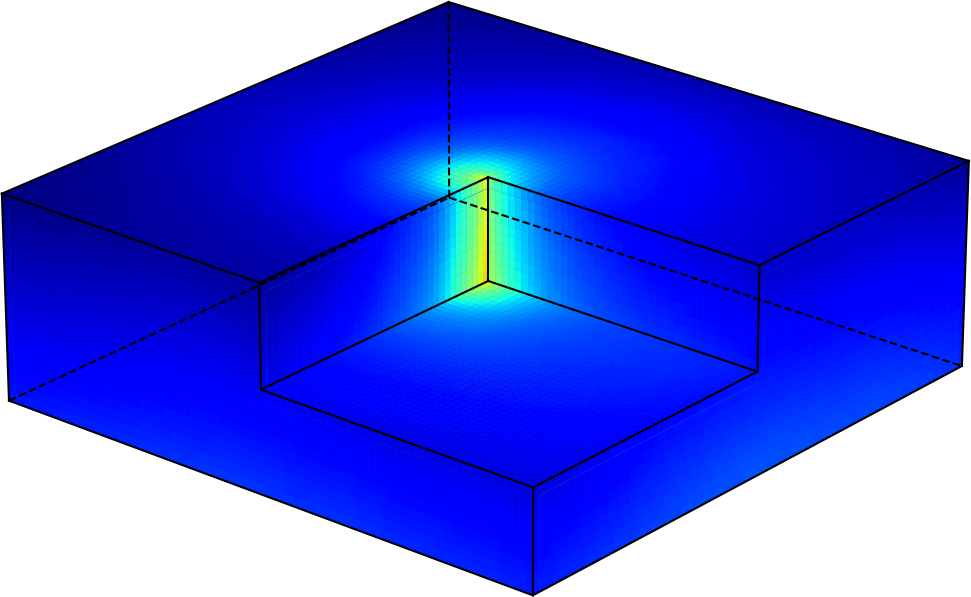}{CNNTransformer3D}\hfill
\predvispanel{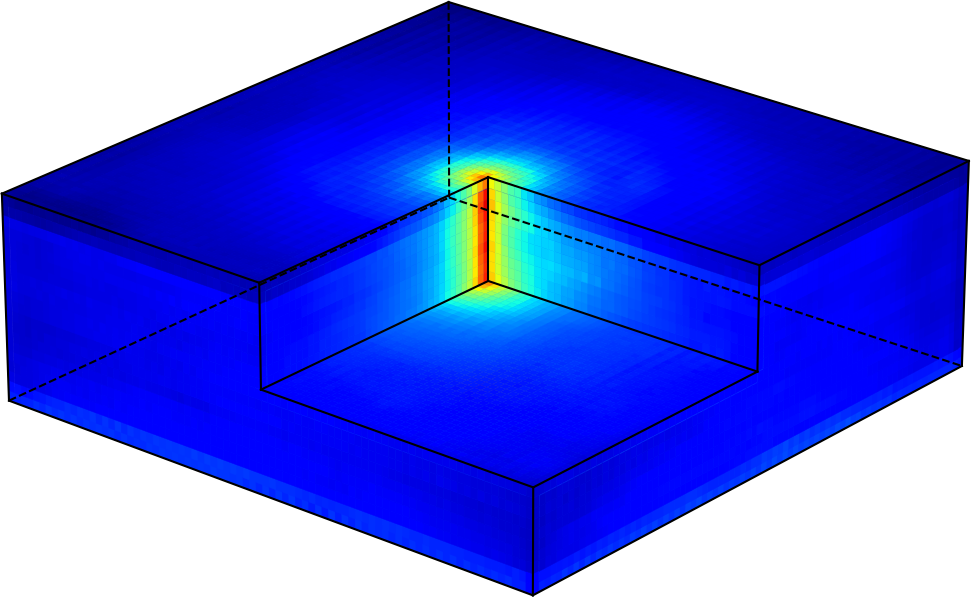}{UDeepONet3D}\hfill
\predvispanel{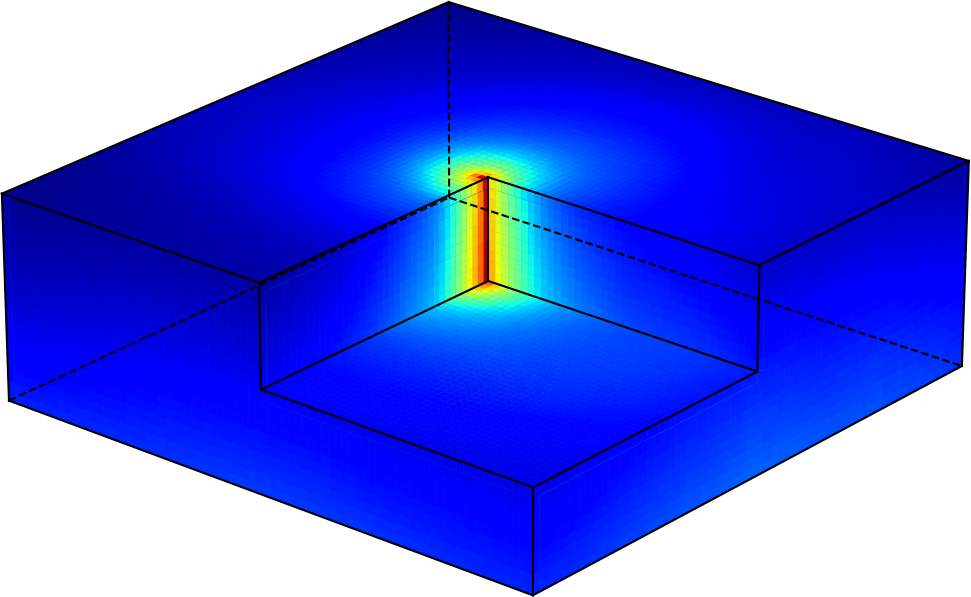}{\textsc{AutoSurrogate}}\hfill
\predvispanel{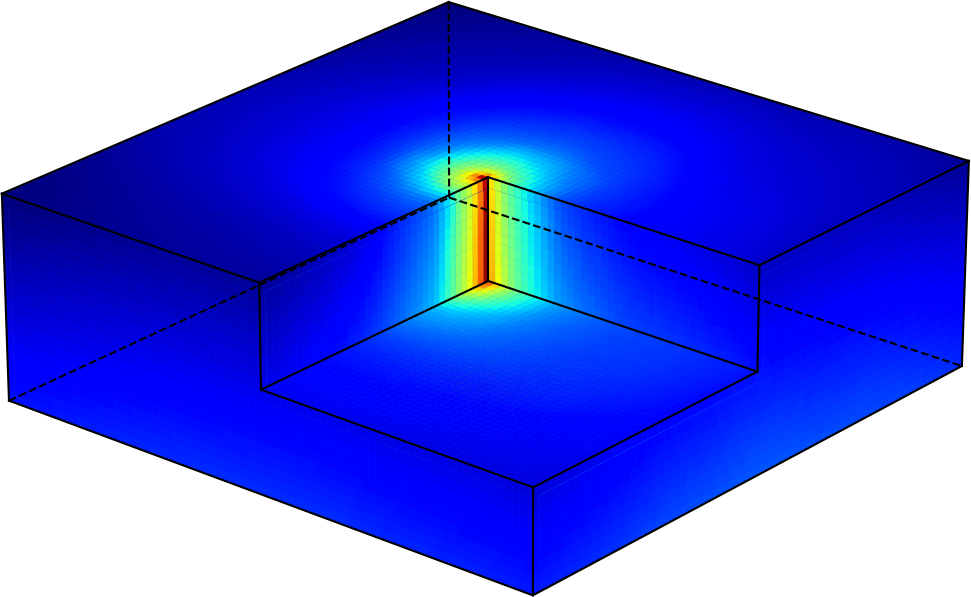}{Ground Truth}\hfill
\predviscolorbar{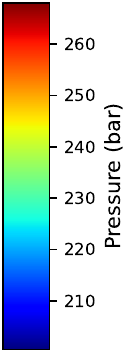}{$P$ [bar]}

\vspace{4pt}

\predvispanel{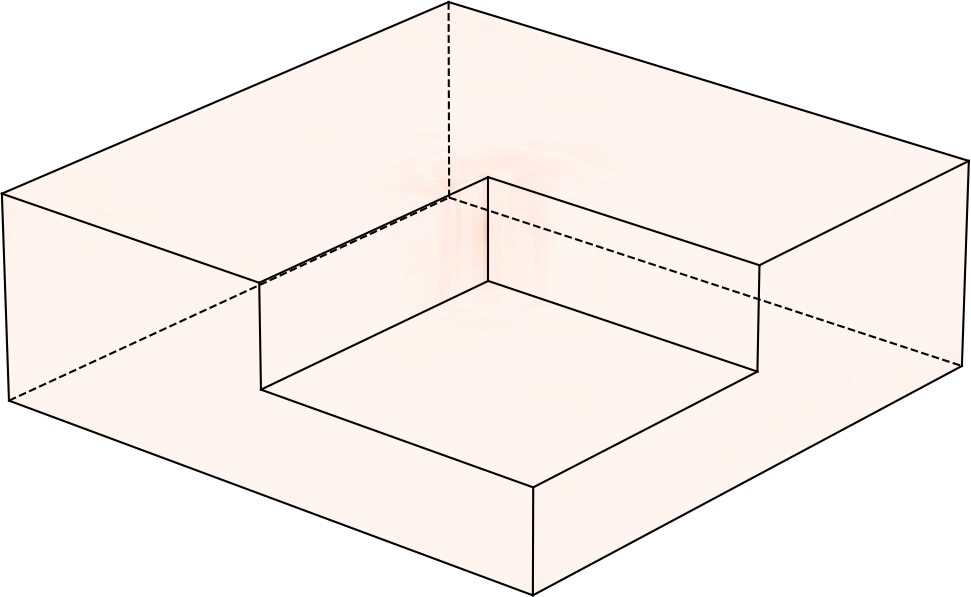}{RecurrentRUNet3D}\hfill
\predvispanel{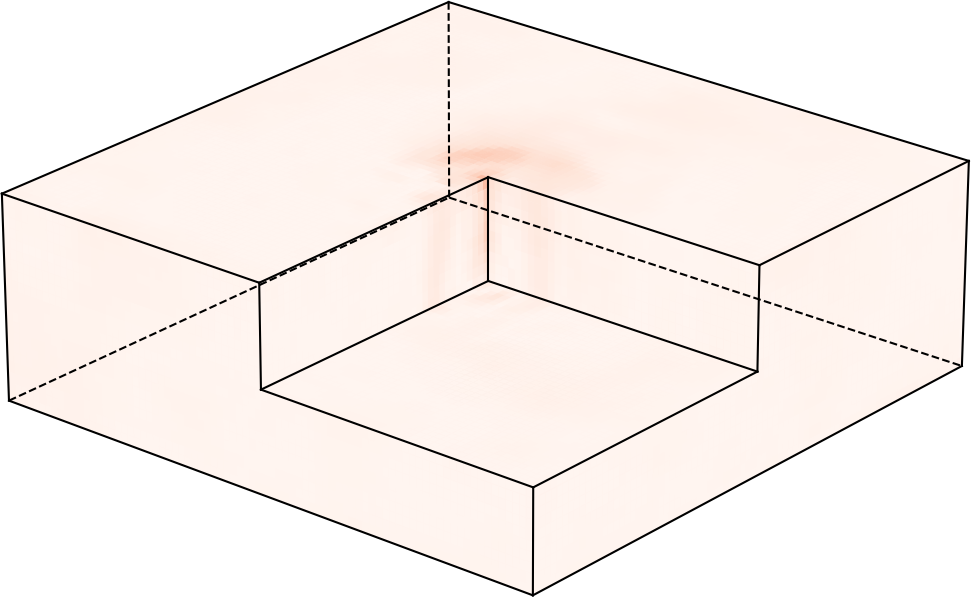}{UFNO3D}\hfill
\predvispanel{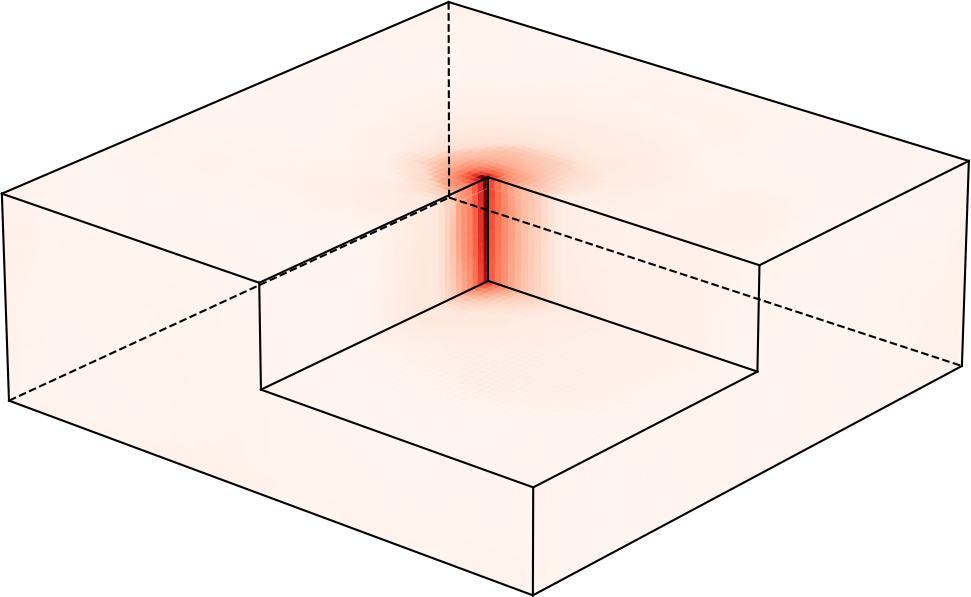}{CNNTransformer3D}\hfill
\predvispanel{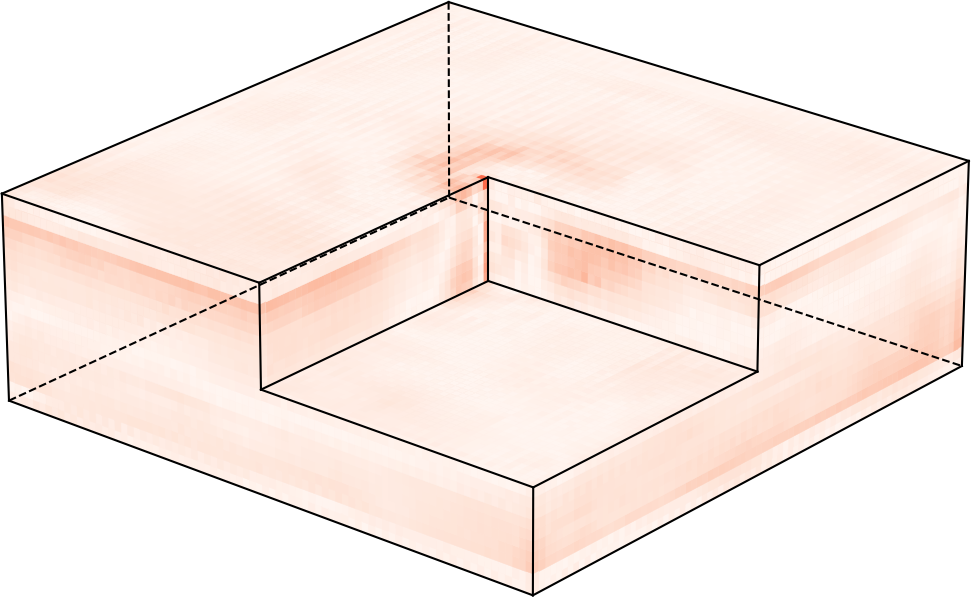}{UDeepONet3D}\hfill
\predvispanel{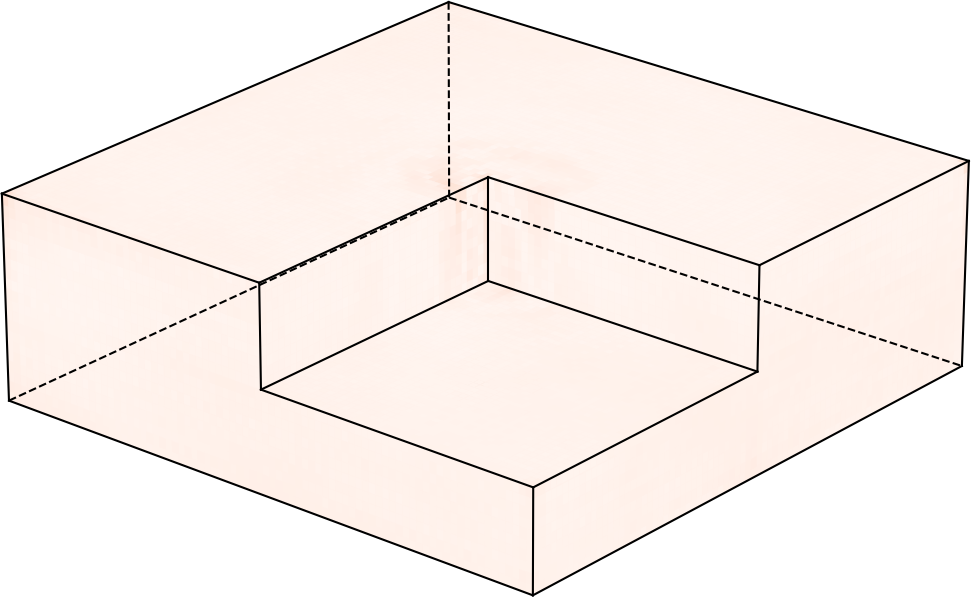}{\textsc{AutoSurrogate}}\hfill
\predvisblank\hfill
\predviscolorbar{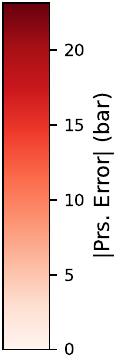}{$|\hat{P}-P|$ [bar]}

\vspace{8pt}

\predvissample{815}
\predvispanelcropped{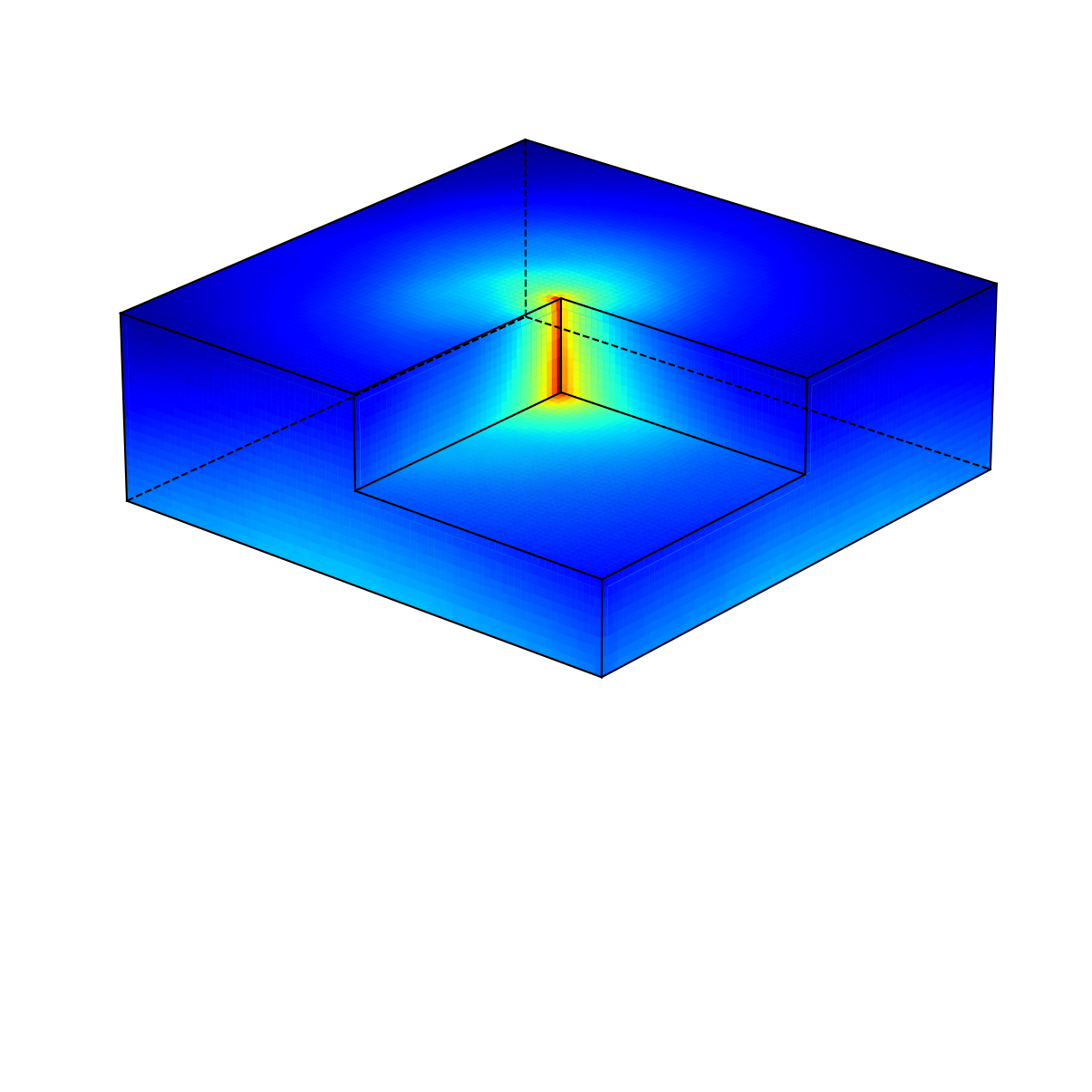}{RecurrentRUNet3D}\hfill
\predvispanelcropped{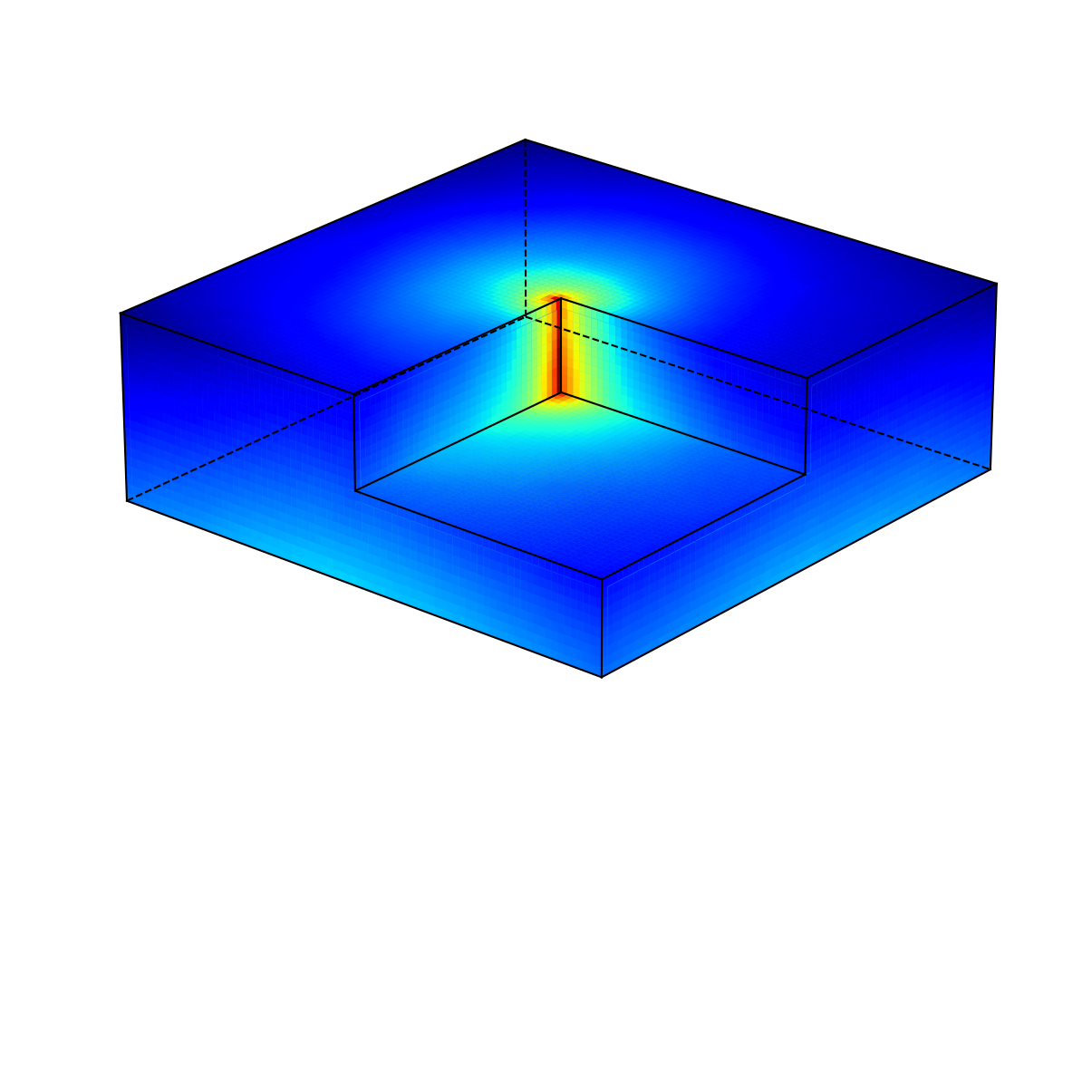}{UFNO3D}\hfill
\predvispanelcropped{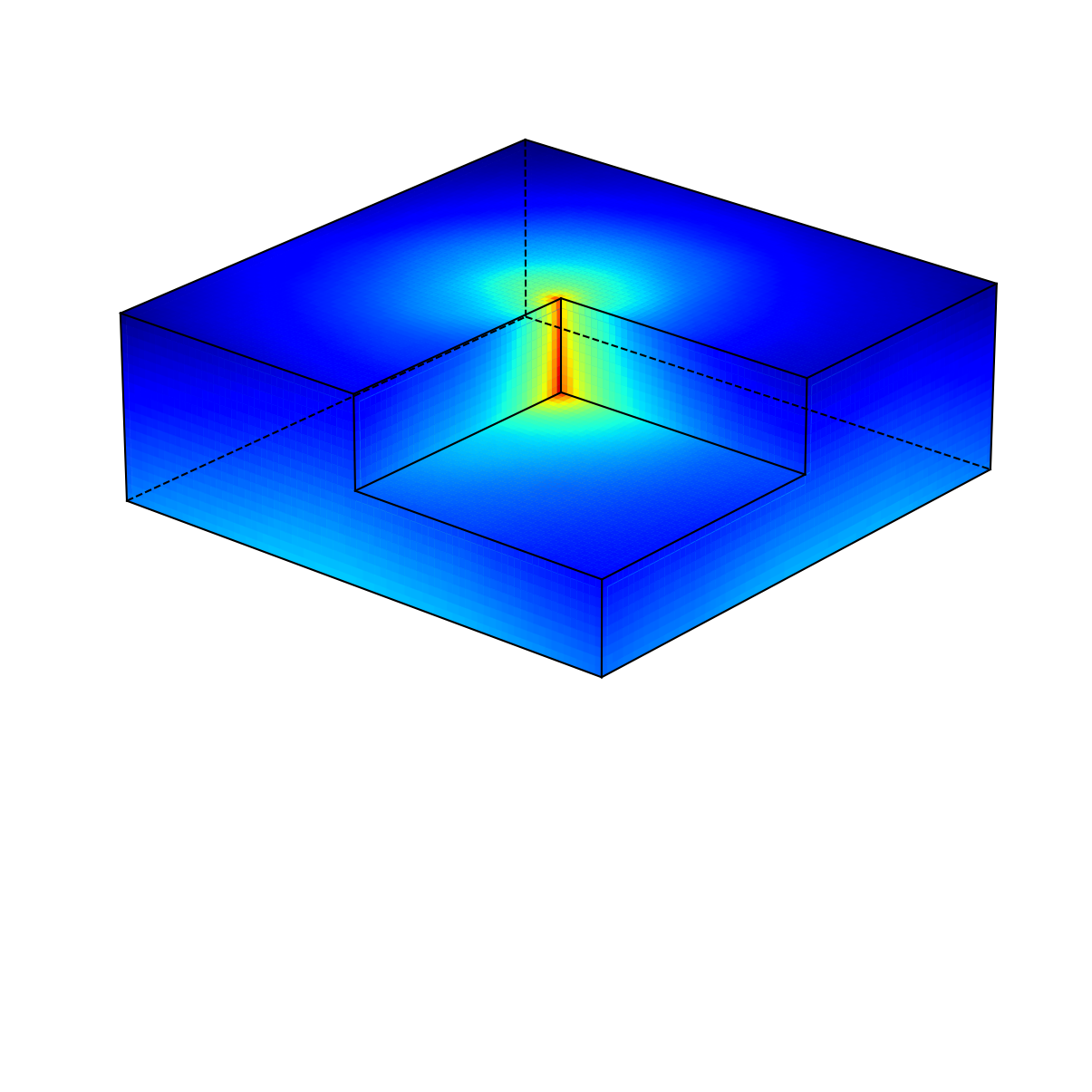}{CNNTransformer3D}\hfill
\predvispanelcropped{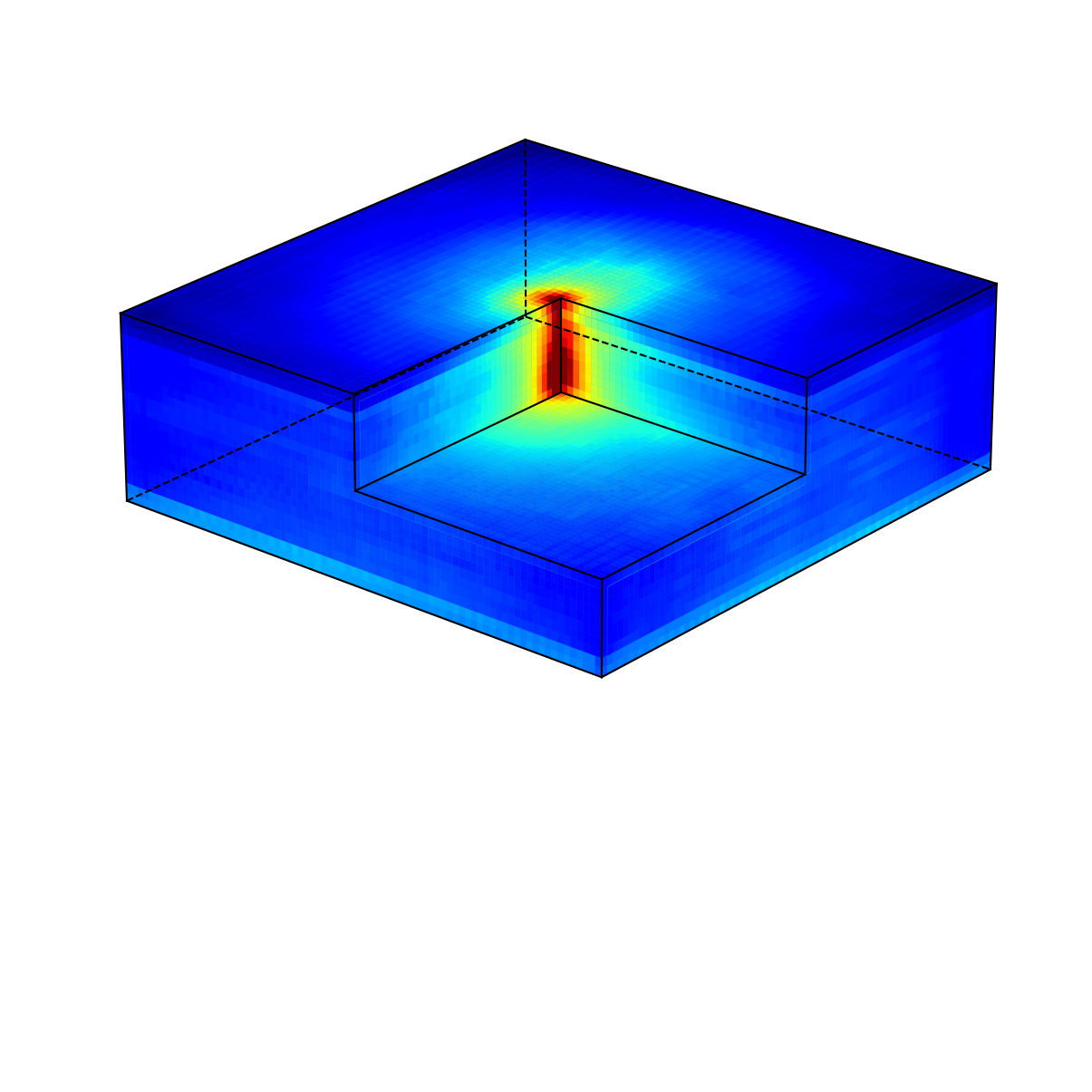}{UDeepONet3D}\hfill
\predvispanelcropped{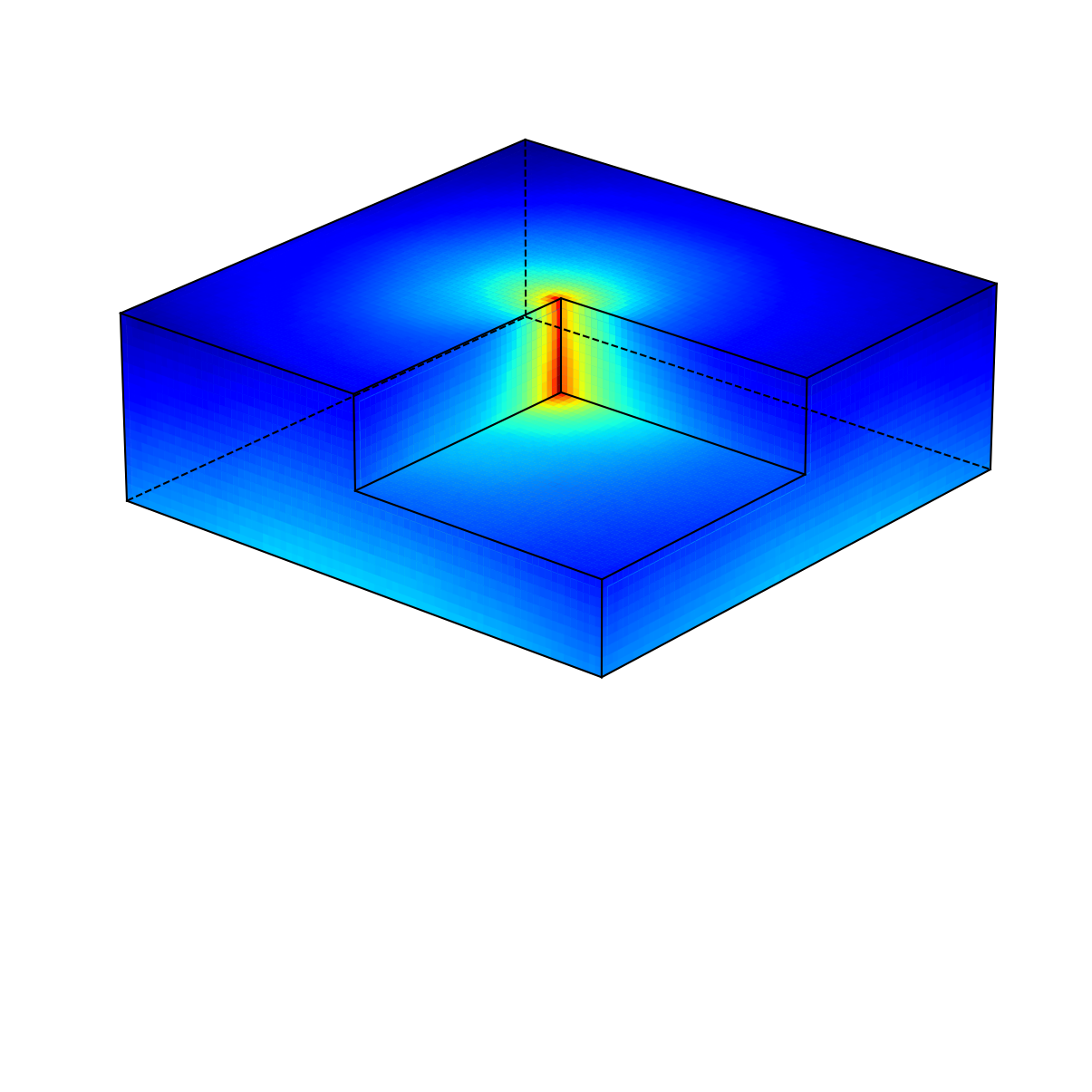}{\textsc{AutoSurrogate}}\hfill
\predvispanelcropped{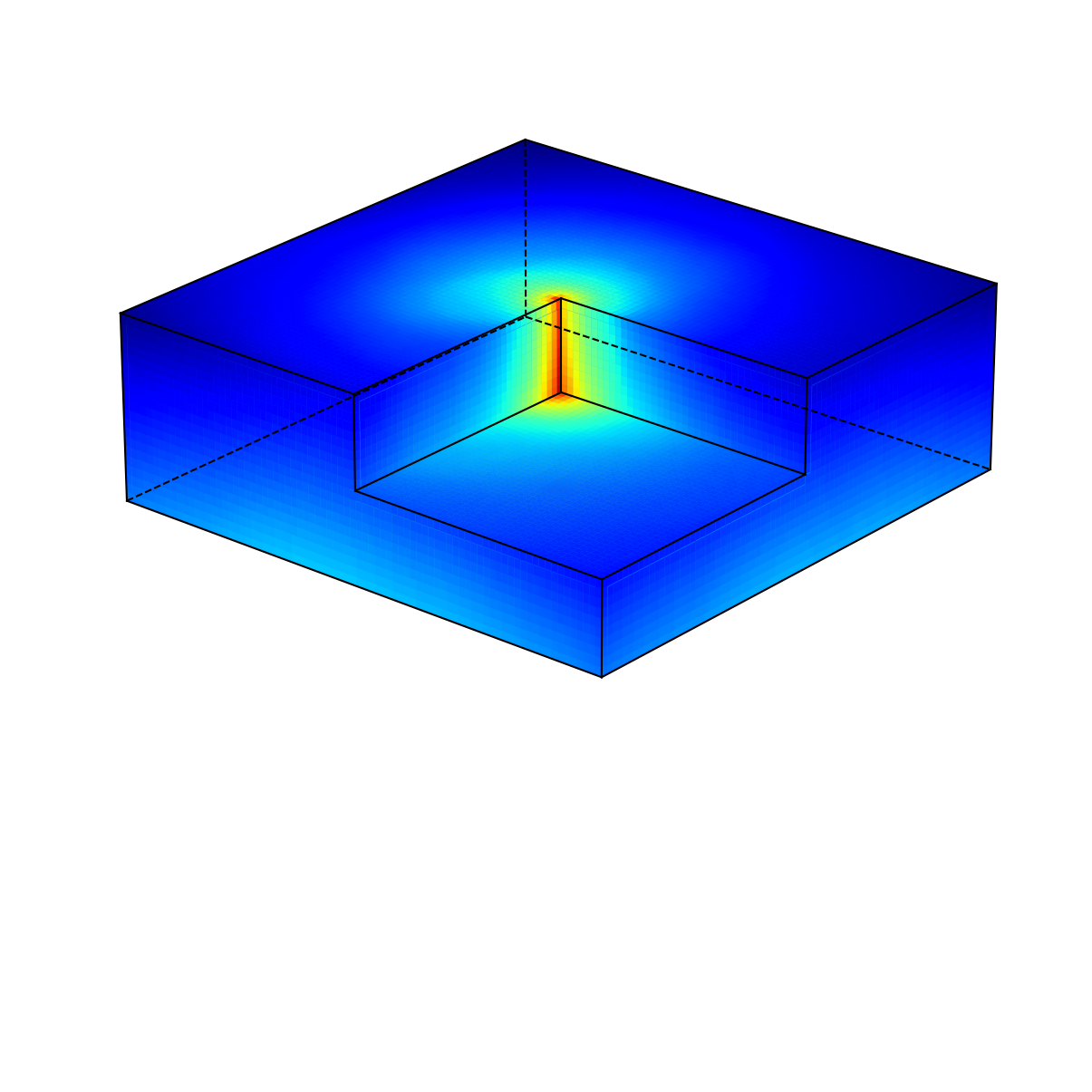}{Ground Truth}\hfill
\predviscolorbar{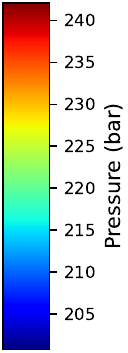}{$P$ [bar]}

\vspace{4pt}

\predvispanelcropped{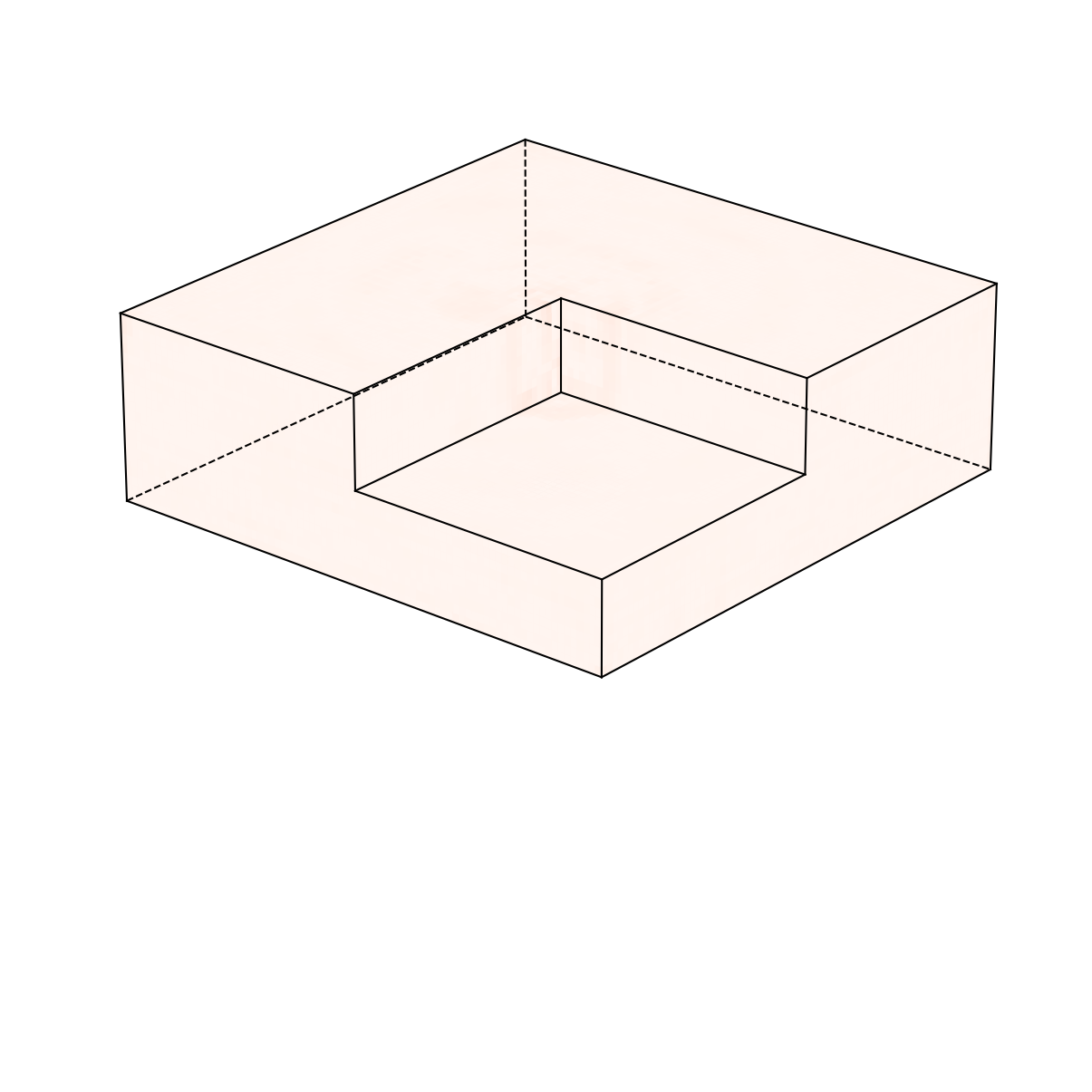}{RecurrentRUNet3D}\hfill
\predvispanelcropped{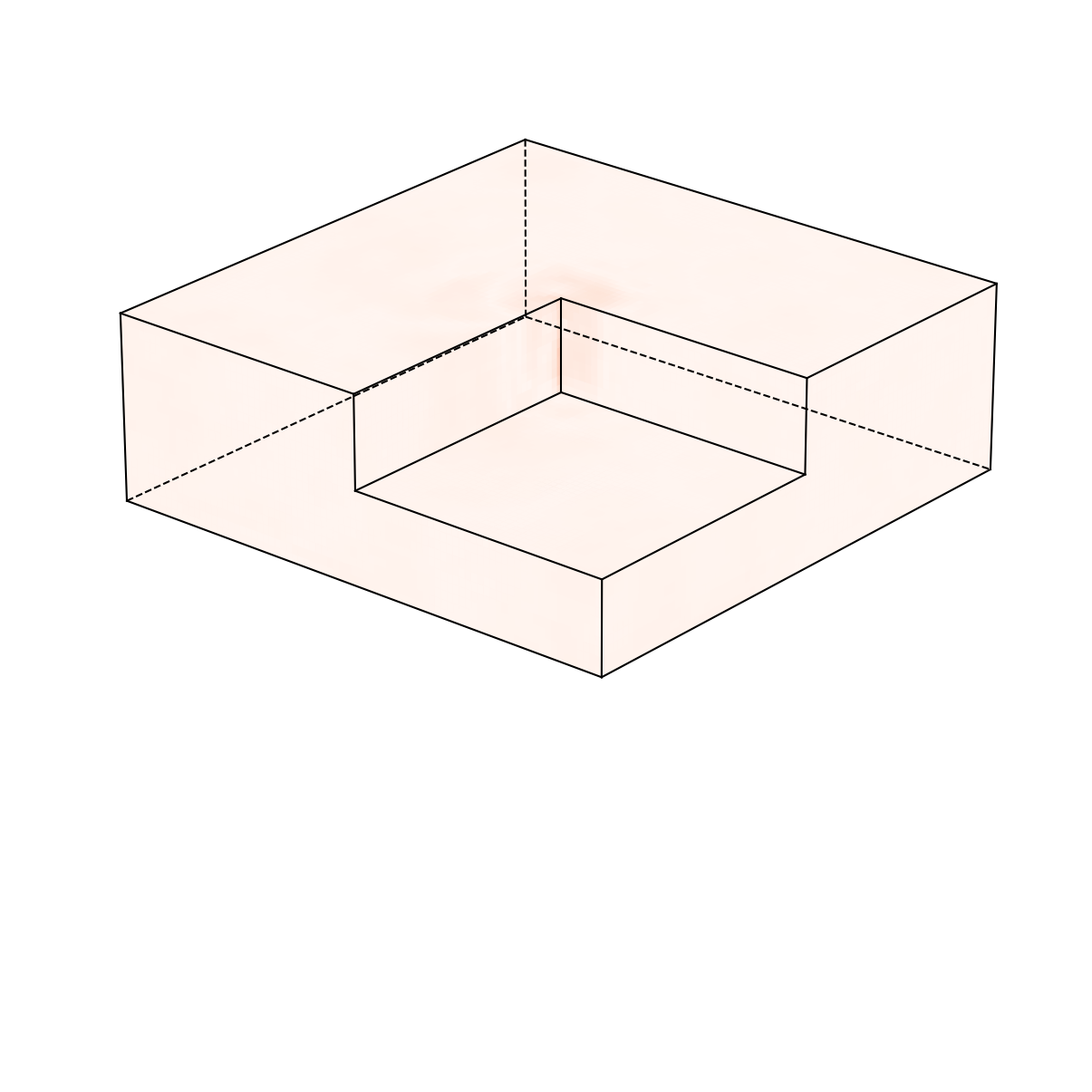}{UFNO3D}\hfill
\predvispanelcropped{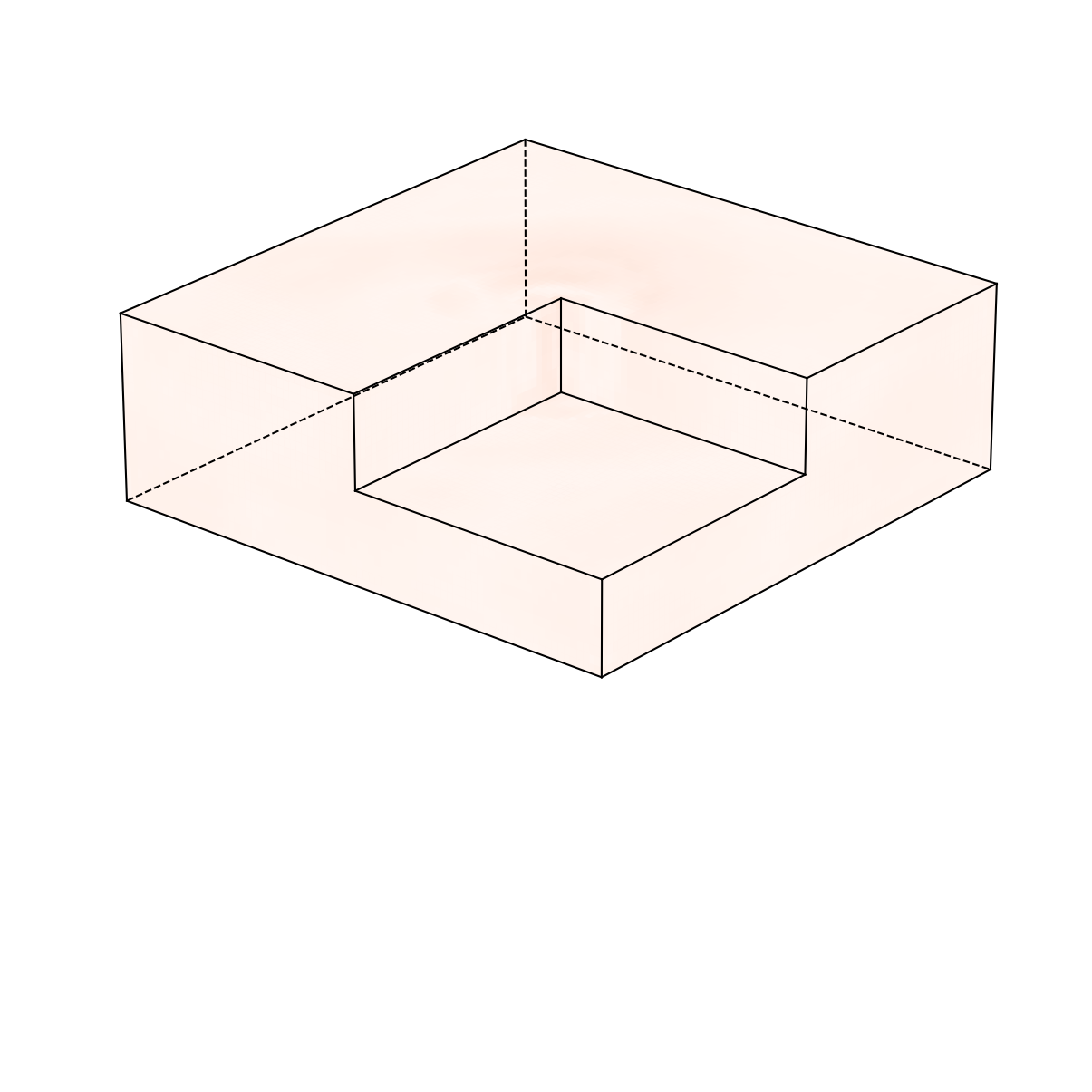}{CNNTransformer3D}\hfill
\predvispanelcropped{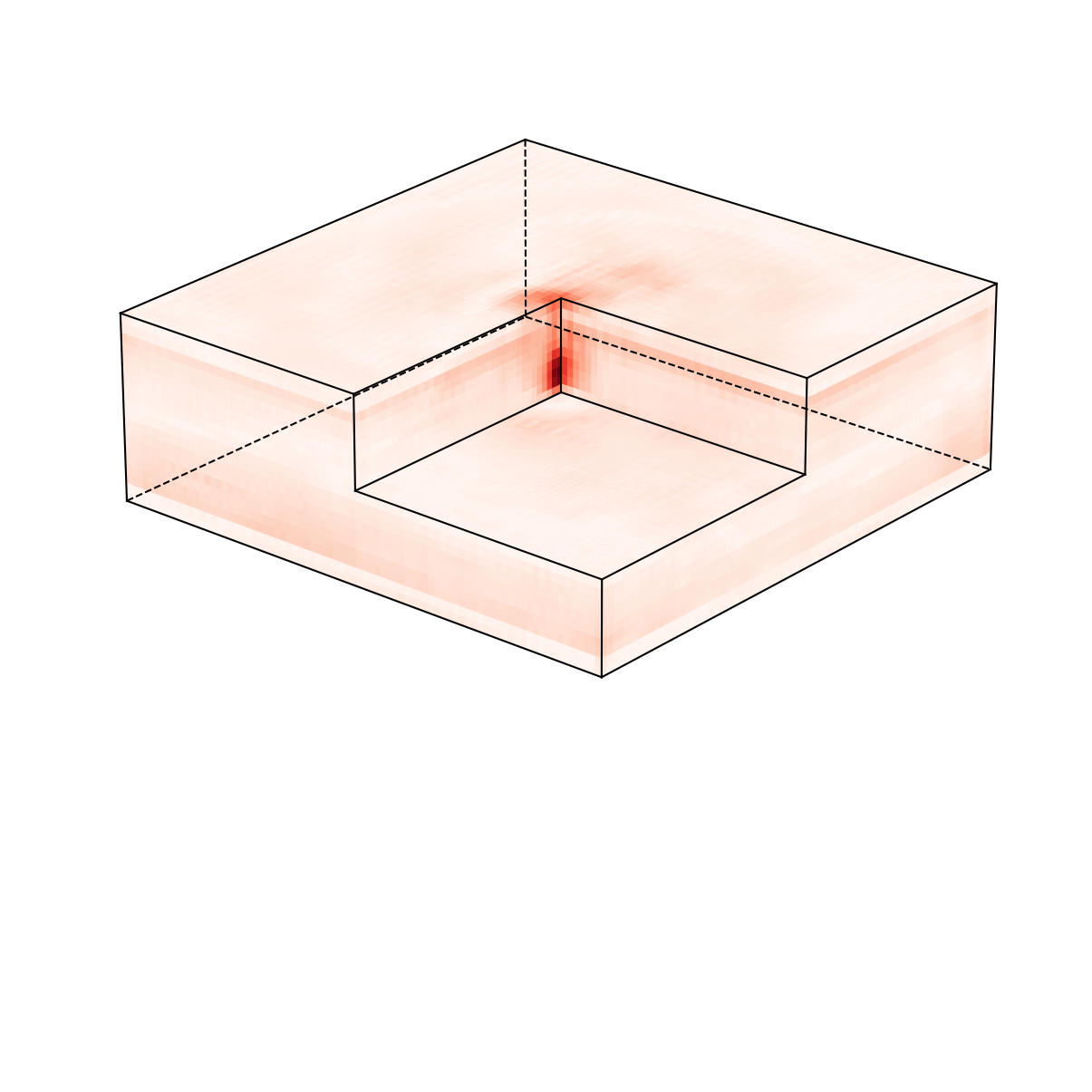}{UDeepONet3D}\hfill
\predvispanelcropped{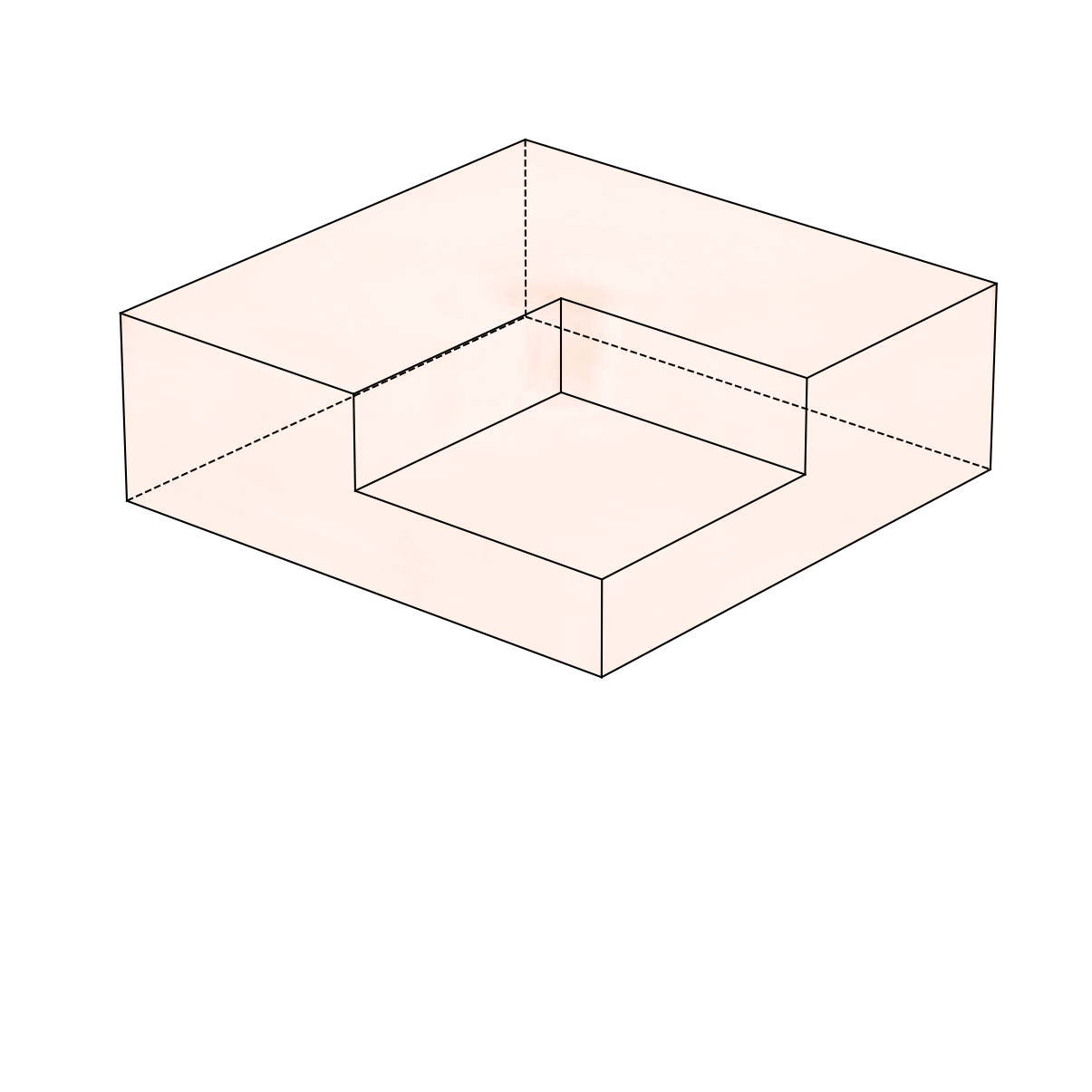}{\textsc{AutoSurrogate}}\hfill
\predvisblank\hfill
\predviscolorbar{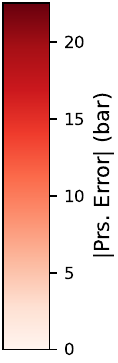}{$|\hat{P}-P|$ [bar]}

\caption{Pressure predictions and absolute errors for Samples~820 and~815 at 30th year. Each sample uses two rows, and the rightmost column shows the shared colorbars.}
\label{fig:pressure_vis}
\end{figure*}

\section{Conclusion}
\label{sec:conclusion}
This work introduced \textsc{AutoSurrogate}, an LLM-driven multi-agent framework for autonomous construction of deep learning surrogate models for subsurface flow problems. The framework takes simulation data and optional natural-language instructions as input, and then carries out data profiling, architecture selection, hyperparameter optimization, model training, self-correction, and reporting with minimal human intervention. By combining high-level reasoning with deterministic execution tools, \textsc{AutoSurrogate} substantially reduces the manual effort typically required to design, tune, and stabilize deep learning surrogate.

We evaluated the framework on a 3D geological carbon storage case, where the surrogate models learn to map permeability fields to pressure and CO$_2$ saturation fields over 31 timesteps. Across autonomous runs, \textsc{AutoSurrogate} achieves strong predictive accuracy for both pressure and saturation evolution, outperforming various baseline models proposed from prior studies. We further compare its performance against domain-agnostic AutoML methods, and show that \textsc{AutoSurrogate} consistently achieves superior results. \textsc{AutoSurrogate} also focuses the search space through task-aware reasoning and achieves better performance on the Pareto front, while maintaining search overhead comparable to that of AutoML methods. In addition, the closed-loop self-correction mechanism improves robustness by recovering from unstable or underperforming training runs.

Overall, the results indicate that autonomous, language-driven surrogate construction is a viable approach for realistic subsurface flow applications. \textsc{AutoSurrogate} provides a practical pathway toward lowering the expertise barrier that has limited the broader adoption of deep learning surrogates in geoscience and reservoir engineering.

\bibliographystyle{cas-model2-names}

\bibliography{cas-refs}

@article{gadd2019surrogate,
	title        = {A surrogate modelling approach based on nonlinear dimension reduction for uncertainty quantification in groundwater flow models},
	author       = {Gadd, C and Xing, Wei and Nezhad, M Mousavi and Shah, Akeel A},
	year         = 2019,
	journal      = {Transport in porous media},
	publisher    = {Springer},
	volume       = 126,
	number       = 1,
	pages        = {39--77}
}

@article{allgeier2023surrogate,
	title        = {Surrogate-Model Assisted Plausibility-Check, Calibration, and Posterior-Distribution Evaluation of Subsurface-Flow Models},
	author       = {Allgeier, Jonas and Cirpka, Olaf A},
	year         = 2023,
	journal      = {Water Resources Research},
	publisher    = {Wiley Online Library},
	volume       = 59,
	number       = 7,
	pages        = {e2023WR034453}
}

@article{elsheikh2014efficient,
	title        = {Efficient Bayesian inference of subsurface flow models using nested sampling and sparse polynomial chaos surrogates},
	author       = {Elsheikh, Ahmed H and Hoteit, Ibrahim and Wheeler, Mary F},
	year         = 2014,
	journal      = {Computer Methods in Applied Mechanics and Engineering},
	publisher    = {Elsevier},
	volume       = 269,
	pages        = {515--537}
}

@article{meng2018uncertainty,
	title        = {Uncertainty quantification for subsurface flow and transport: Coping with nonlinearity/irregularity via polynomial chaos surrogate and machine learning},
	author       = {Meng, J and Li, H},
	year         = 2018,
	journal      = {Water Resources Research},
	publisher    = {Wiley Online Library},
	volume       = 54,
	number       = 10,
	pages        = {7733--7751}
}

@article{babaei2015robust,
	title        = {Robust optimization of subsurface flow using polynomial chaos and response surface surrogates},
	author       = {Babaei, Masoud and Alkhatib, Ali and Pan, Indranil},
	year         = 2015,
	journal      = {Computational Geosciences},
	publisher    = {Springer},
	volume       = 19,
	number       = 5,
	pages        = {979--998}
}

@book{hesthaven2016certified,
	title        = {Certified reduced basis methods for parametrized partial differential equations},
	author       = {Hesthaven, Jan S and Rozza, Gianluigi and Stamm, Benjamin and others},
	year         = 2016,
	publisher    = {Springer},
	volume       = 590
}

@article{asher2015review,
	title        = {A review of surrogate models and their application to groundwater modeling},
	author       = {Asher, Michael J and Croke, Barry FW and Jakeman, Anthony J and Peeters, Luk JM},
	year         = 2015,
	journal      = {Water Resources Research},
	publisher    = {Wiley Online Library},
	volume       = 51,
	number       = 8,
	pages        = {5957--5973}
}

@article{tadmor2010mean,
	title        = {Mean field representation of the natural and actuated cylinder wake},
	author       = {Tadmor, Gilead and Lehmann, Oliver and Noack, Bernd R and Morzy{\'n}ski, Marek},
	year         = 2010,
	journal      = {Physics of Fluids},
	publisher    = {AIP Publishing},
	volume       = 22,
	number       = 3
}

@article{semaan2016reduced,
	title        = {Reduced-order modelling of the flow around a high-lift configuration with unsteady Coanda blowing},
	author       = {Semaan, Richard and Kumar, Pradeep and Burnazzi, Marco and Tissot, Gilles and Cordier, Laurent and Noack, Bernd R},
	year         = 2016,
	journal      = {Journal of Fluid Mechanics},
	publisher    = {Cambridge University Press},
	volume       = 800,
	pages        = {72--110}
}

@article{buhmann2000radial,
	title        = {Radial basis functions},
	author       = {Buhmann, Martin Dietrich},
	year         = 2000,
	journal      = {Acta numerica},
	publisher    = {Cambridge university press},
	volume       = 9,
	pages        = {1--38}
}

@article{he2019data,
	title        = {Data-driven uncertainty quantification for predictive flow and transport modeling using support vector machines},
	author       = {He, Jiachuan and Mattis, Steven A and Butler, Troy D and Dawson, Clint N},
	year         = 2019,
	journal      = {Computational Geosciences},
	publisher    = {Springer},
	volume       = 23,
	number       = 4,
	pages        = {631--645}
}

@article{gao2025reduced,
	title        = {Reduced-Order Modeling for Fractured Reservoir Simulation by Use of Local Resolution Trajectory Piecewise Linearization},
	author       = {Gao, Ming and Jia, Yuyang and Cui, Jiawei and Zhang, Yongbin and Li, Junchao and Sun, Wenyue},
	year         = 2025,
	journal      = {SPE Journal},
	publisher    = {SPE},
	volume       = 30,
	number       = 12,
	pages        = {7822--7839}
}

@article{mo2019deep,
	title        = {Deep convolutional encoder-decoder networks for uncertainty quantification of dynamic multiphase flow in heterogeneous media},
	author       = {Mo, Shaoxing and Zhu, Yinhao and Zabaras, Nicholas and Shi, Xiaoqing and Wu, Jichun},
	year         = 2019,
	journal      = {Water Resources Research},
	publisher    = {Wiley Online Library},
	volume       = 55,
	number       = 1,
	pages        = {703--728}
}

@article{ma2022efficient,
	title        = {An efficient spatial-temporal convolution recurrent neural network surrogate model for history matching},
	author       = {Ma, Xiaopeng and Zhang, Kai and Wang, Jian and Yao, Chuanjin and Yang, Yongfei and Sun, Hai and Yao, Jun},
	year         = 2022,
	journal      = {SPE Journal},
	publisher    = {OnePetro},
	volume       = 27,
	number       = {02},
	pages        = {1160--1175}
}

@article{wang2023deep,
	title        = {Deep learning based closed-loop well control optimization of geothermal reservoir with uncertain permeability},
	author       = {Wang, Nanzhe and Chang, Haibin and Kong, Xiang-Zhao and Zhang, Dongxiao},
	year         = 2023,
	journal      = {Renewable Energy},
	publisher    = {Elsevier},
	volume       = 211,
	pages        = {379--394}
}

@article{tang2020deep,
	title        = {A deep-learning-based surrogate model for data assimilation in dynamic subsurface flow problems},
	author       = {Tang, Meng and Liu, Yimin and Durlofsky, Louis J},
	year         = 2020,
	journal      = {Journal of Computational Physics},
	publisher    = {Elsevier},
	volume       = 413,
	pages        = 109456
}

@article{wen2022u,
	title        = {U-FNO—An enhanced Fourier neural operator-based deep-learning model for multiphase flow},
	author       = {Wen, Gege and Li, Zongyi and Azizzadenesheli, Kamyar and Anandkumar, Anima and Benson, Sally M},
	year         = 2022,
	journal      = {Advances in Water Resources},
	publisher    = {Elsevier},
	volume       = 163,
	pages        = 104180
}

@article{pawar2026accelerated,
	title        = {Accelerated ccs site screening using fourier neural operator based surrogates for flow simulations},
	author       = {Pawar, Suraj and Panda, Aniruddha and Chandra, Anirban and Devarakota, Pandu and Alpak, Faruk O and Snippe, Jeroen and Hohl, Detlef},
	year         = 2026,
	journal      = {International Journal of Greenhouse Gas Control},
	publisher    = {Elsevier},
	volume       = 150,
	pages        = 104583
}

@article{tang2024graph,
	title        = {Graph network surrogate model for subsurface flow optimization},
	author       = {Tang, Haoyu and Durlofsky, Louis J},
	year         = 2024,
	journal      = {Journal of Computational Physics},
	publisher    = {Elsevier},
	volume       = 512,
	pages        = 113132
}

@article{tang2025graph,
	title        = {Graph network surrogate model for optimizing the placement of horizontal injection wells for CO2 storage},
	author       = {Tang, Haoyu and Durlofsky, Louis J},
	year         = 2025,
	journal      = {International Journal of Greenhouse Gas Control},
	publisher    = {Elsevier},
	volume       = 145,
	pages        = 104404
}

@article{wang2021theory,
	title        = {Theory-guided auto-encoder for surrogate construction and inverse modeling},
	author       = {Wang, Nanzhe and Chang, Haibin and Zhang, Dongxiao},
	year         = 2021,
	journal      = {Computer Methods in Applied Mechanics and Engineering},
	publisher    = {Elsevier},
	volume       = 385,
	pages        = 114037
}

@article{fu2025deep,
	title        = {Deep learning-based surrogate modeling for underground hydrogen storage},
	author       = {Fu, Shuojia and Mao, Shaowen and Carbonero, Alvaro and Srikishan, Bharat and Creasy, Neala and Chellal, Hichem and Mehana, Mohamed},
	year         = 2025,
	journal      = {Advances in Water Resources},
	publisher    = {Elsevier},
	volume       = 203,
	pages        = 105014
}

@article{han2026recurrent,
	title        = {Recurrent Transformer U-Net Surrogate for Flow Modeling and Data Assimilation in Subsurface Formations with Faults},
	author       = {Han, Yifu and Durlofsky, Louis J},
	year         = 2026,
	journal      = {Journal of Computational Physics},
	publisher    = {Elsevier},
	pages        = 114801
}

@article{wang2021efficient,
	title        = {Efficient uncertainty quantification for dynamic subsurface flow with surrogate by theory-guided neural network},
	author       = {Wang, Nanzhe and Chang, Haibin and Zhang, Dongxiao},
	year         = 2021,
	journal      = {Computer Methods in Applied Mechanics and Engineering},
	publisher    = {Elsevier},
	volume       = 373,
	pages        = 113492
}

@article{wang2022surrogate,
	title        = {Surrogate and inverse modeling for two-phase flow in porous media via theory-guided convolutional neural network},
	author       = {Wang, Nanzhe and Chang, Haibin and Zhang, Dongxiao},
	year         = 2022,
	journal      = {Journal of Computational Physics},
	publisher    = {Elsevier},
	volume       = 466,
	pages        = 111419
}

@article{karumuri2020simulator,
	title        = {Simulator-free solution of high-dimensional stochastic elliptic partial differential equations using deep neural networks},
	author       = {Karumuri, Sharmila and Tripathy, Rohit and Bilionis, Ilias and Panchal, Jitesh},
	year         = 2020,
	journal      = {Journal of Computational Physics},
	publisher    = {Elsevier},
	volume       = 404,
	pages        = 109120
}

@article{zhu2019physics,
	title        = {Physics-constrained deep learning for high-dimensional surrogate modeling and uncertainty quantification without labeled data},
	author       = {Zhu, Yinhao and Zabaras, Nicholas and Koutsourelakis, Phaedon-Stelios and Perdikaris, Paris},
	year         = 2019,
	journal      = {Journal of computational physics},
	publisher    = {Elsevier},
	volume       = 394,
	pages        = {56--81}
}

@article{wang2025deep,
	title        = {Deep learning framework for history matching CO2 storage with 4D seismic and monitoring well data},
	author       = {Wang, Nanzhe and Durlofsky, Louis J},
	year         = 2025,
	journal      = {Geoenergy Science and Engineering},
	publisher    = {Elsevier},
	volume       = 248,
	pages        = 213736
}

@article{naveed2025comprehensive,
	title        = {A comprehensive overview of large language models},
	author       = {Naveed, Humza and Khan, Asad Ullah and Qiu, Shi and Saqib, Muhammad and Anwar, Saeed and Usman, Muhammad and Akhtar, Naveed and Barnes, Nick and Mian, Ajmal},
	year         = 2025,
	journal      = {ACM Transactions on Intelligent Systems and Technology},
	publisher    = {ACM New York, NY},
	volume       = 16,
	number       = 5,
	pages        = {1--72}
}

@article{sapkota2025ai,
	title        = {Ai agents vs. agentic ai: A conceptual taxonomy, applications and challenges},
	author       = {Sapkota, Ranjan and Roumeliotis, Konstantinos I and Karkee, Manoj},
	year         = 2025,
	journal      = {Information Fusion},
	publisher    = {Elsevier},
	pages        = 103599
}

@article{masterman2024landscape,
	title        = {The landscape of emerging ai agent architectures for reasoning, planning, and tool calling: A survey},
	author       = {Masterman, Tula and Besen, Sandi and Sawtell, Mason and Chao, Alex},
	year         = 2024,
	journal      = {arXiv preprint arXiv:2404.11584}
}

@article{touvron2023llama,
	title        = {Llama: Open and efficient foundation language models},
	author       = {Touvron, Hugo and Lavril, Thibaut and Izacard, Gautier and Martinet, Xavier and Lachaux, Marie-Anne and Lacroix, Timoth{\'e}e and Rozi{\`e}re, Baptiste and Goyal, Naman and Hambro, Eric and Azhar, Faisal and others},
	year         = 2023,
	journal      = {arXiv preprint arXiv:2302.13971}
}

@article{achiam2023gpt,
	title        = {Gpt-4 technical report},
	author       = {Achiam, Josh and Adler, Steven and Agarwal, Sandhini and Ahmad, Lama and Akkaya, Ilge and Aleman, Florencia Leoni and Almeida, Diogo and Altenschmidt, Janko and Altman, Sam and Anadkat, Shyamal and others},
	year         = 2023,
	journal      = {arXiv preprint arXiv:2303.08774}
}

@article{guo2025deepseek,
	title        = {DeepSeek-R1 incentivizes reasoning in LLMs through reinforcement learning},
	author       = {Guo, Daya and Yang, Dejian and Zhang, Haowei and Song, Junxiao and Wang, Peiyi and Zhu, Qihao and Xu, Runxin and Zhang, Ruoyu and Ma, Shirong and Bi, Xiao and others},
	year         = 2025,
	journal      = {Nature},
	publisher    = {Nature Publishing Group UK London},
	volume       = 645,
	number       = 8081,
	pages        = {633--638}
}

@article{gao2024empowering,
	title        = {Empowering biomedical discovery with AI agents},
	author       = {Gao, Shanghua and Fang, Ada and Huang, Yepeng and Giunchiglia, Valentina and Noori, Ayush and Schwarz, Jonathan Richard and Ektefaie, Yasha and Kondic, Jovana and Zitnik, Marinka},
	year         = 2024,
	journal      = {Cell},
	publisher    = {Elsevier},
	volume       = 187,
	number       = 22,
	pages        = {6125--6151}
}

@article{moritz2025coordinated,
	title        = {Coordinated AI agents for advancing healthcare},
	author       = {Moritz, Michael and Topol, Eric and Rajpurkar, Pranav},
	year         = 2025,
	journal      = {Nature Biomedical Engineering},
	publisher    = {Nature Publishing Group UK London},
	volume       = 9,
	number       = 4,
	pages        = {432--438}
}

@article{fang2025comprehensive,
	title        = {A comprehensive survey of self-evolving ai agents: A new paradigm bridging foundation models and lifelong agentic systems},
	author       = {Fang, Jinyuan and Peng, Yanwen and Zhang, Xi and Wang, Yingxu and Yi, Xinhao and Zhang, Guibin and Xu, Yi and Wu, Bin and Liu, Siwei and Li, Zihao and others},
	year         = 2025,
	journal      = {arXiv preprint arXiv:2508.07407}
}

@article{wang2024openhands,
	title        = {Openhands: An open platform for ai software developers as generalist agents},
	author       = {Wang, Xingyao and Li, Boxuan and Song, Yufan and Xu, Frank F and Tang, Xiangru and Zhuge, Mingchen and Pan, Jiayi and Song, Yueqi and Li, Bowen and Singh, Jaskirat and others},
	year         = 2024,
	journal      = {arXiv preprint arXiv:2407.16741}
}

@article{han2025accelerated,
	title        = {Accelerated training of deep learning surrogate models for surface displacement and flow, with application to MCMC-based history matching of CO2 storage operations},
	author       = {Han, Yifu and Hamon, Francois P and Durlofsky, Louis J},
	year         = 2025,
	journal      = {Geoenergy Science and Engineering},
	publisher    = {Elsevier},
	volume       = 246,
	pages        = 213589
}

@misc{yaoReActSynergizingReasoning2023,
	title        = {{ReAct}: {Synergizing} {Reasoning} and {Acting} in {Language} {Models}},
	shorttitle   = {{ReAct}},
	author       = {Yao, Shunyu and Zhao, Jeffrey and Yu, Dian and Du, Nan and Shafran, Izhak and Narasimhan, Karthik and Cao, Yuan},
	year         = 2023,
	month        = mar,
	publisher    = {arXiv},
	doi          = {10.48550/arXiv.2210.03629}
}

@article{tangDeeplearningbasedSurrogateFlow2021,
	title        = {Deep-learning-based surrogate flow modeling and geological parameterization for data assimilation in {3D} subsurface flow},
	author       = {Tang, Meng and Liu, Yimin and Durlofsky, Louis J.},
	year         = 2021,
	month        = apr,
	journal      = {Computer Methods in Applied Mechanics and Engineering},
	volume       = 376,
	pages        = 113636,
	doi          = {10.1016/j.cma.2020.113636},
	issn         = {00457825},
	language     = {en}
}

@article{fengEncoderdecoderConvLSTMSurrogate2024,
	title        = {An encoder-decoder {ConvLSTM} surrogate model for simulating geological {CO2} sequestration with dynamic well controls},
	author       = {Feng, Zhao and Tariq, Zeeshan and Shen, Xianda and Yan, Bicheng and Tang, Xuhai and Zhang, Fengshou},
	year         = 2024,
	month        = may,
	journal      = {Gas Science and Engineering},
	volume       = 125,
	pages        = 205314,
	doi          = {10.1016/j.jgsce.2024.205314},
	issn         = {2949-9089}
}

@article{diabUDeepONetUNetEnhanced2024,
	title        = {U-{DeepONet}: {U}-{Net} enhanced deep operator network for geologic carbon sequestration},
	shorttitle   = {U-{DeepONet}},
	author       = {Diab, Waleed and Al Kobaisi, Mohammed},
	year         = 2024,
	month        = sep,
	journal      = {Scientific Reports},
	publisher    = {Nature Publishing Group},
	volume       = 14,
	number       = 1,
	pages        = 21298,
	doi          = {10.1038/s41598-024-72393-0},
	issn         = {2045-2322},
	copyright    = {2024 The Author(s)},
	language     = {en}
}

@misc{liFourierNeuralOperator2020,
	title        = {Fourier {Neural} {Operator} for {Parametric} {Partial} {Differential} {Equations}},
	author       = {Li, Zongyi and Kovachki, Nikola and Azizzadenesheli, Kamyar and Liu, Burigede and Bhattacharya, Kaushik and Stuart, Andrew and Anandkumar, Anima},
	year         = 2020,
	month        = oct,
	journal      = {arXiv.org},
	language     = {en}
}

@inproceedings{akibaOptunaNextgenerationHyperparameter2019,
	title        = {Optuna: {A} {Next}-generation {Hyperparameter} {Optimization} {Framework}},
	shorttitle   = {Optuna},
	author       = {Akiba, Takuya and Sano, Shotaro and Yanase, Toshihiko and Ohta, Takeru and Koyama, Masanori},
	year         = 2019,
	month        = jul,
	booktitle    = {Proceedings of the 25th {ACM} {SIGKDD} {International} {Conference} on {Knowledge} {Discovery} \& {Data} {Mining}},
	publisher    = {ACM},
	address      = {Anchorage AK USA},
	pages        = {2623--2631},
	doi          = {10.1145/3292500.3330701},
	isbn         = {978-1-4503-6201-6},
	language     = {en}
}

@article{feng2025hybrid,
	title        = {A hybrid CNN-transformer surrogate model for the multi-objective robust optimization of geological carbon sequestration},
	author       = {Feng, Zhao and Yan, Bicheng and Shen, Xianda and Zhang, Fengshou and Tariq, Zeeshan and Ouyang, Weiquan and Han, Zhilei},
	year         = 2025,
	journal      = {Advances in Water Resources},
	publisher    = {Elsevier},
	volume       = 196,
	pages        = 104897
}

@article{jiang2021deep,
	title        = {Deep residual U-net convolution neural networks with autoregressive strategy for fluid flow predictions in large-scale geosystems},
	author       = {Jiang, Zhihao and Tahmasebi, Pejman and Mao, Zhiqiang},
	year         = 2021,
	journal      = {Advances in Water Resources},
	publisher    = {Elsevier},
	volume       = 150,
	pages        = 103878
}

@book{remy2009applied,
	title        = {Applied geostatistics with SGeMS: A user's guide},
	author       = {Remy, Nicolas and Boucher, Alexandre and Wu, Jianbing},
	year         = 2009,
	publisher    = {Cambridge University Press}
}

@article{saadatpoor2010new,
	title        = {New trapping mechanism in carbon sequestration},
	author       = {Saadatpoor, Ehsan and Bryant, Steven L and Sepehrnoori, Kamy},
	year         = 2010,
	journal      = {Transport in porous media},
	publisher    = {Springer},
	volume       = 82,
	number       = 1,
	pages        = {3--17}
}

@article{settgast2024geos,
	title        = {GEOS: A performance portable multi-physics simulation framework for subsurface applications},
	author       = {Settgast, Randolph R and Aronson, Ryan M and Besset, Julien R and Borio, Andrea and Bui, Quan M and Byer, Thomas J and Castelletto, Nicola and Citrain, Aur{\'e}lien and Corbett, Benjamin C and Corbett, James and others},
	year         = 2024,
	journal      = {Journal of Open Source Software},
	publisher    = {Lawrence Livermore National Laboratory (LLNL), Livermore, CA (United States)},
	volume       = 9,
	number       = {LLNL--JRNL-864747}
}

@inproceedings{bergstraMakingScienceModel2013,
	title        = {Making a {Science} of {Model} {Search}: {Hyperparameter} {Optimization} in {Hundreds} of {Dimensions} for {Vision} {Architectures}},
	shorttitle   = {Making a {Science} of {Model} {Search}},
	author       = {Bergstra, James and Yamins, Daniel and Cox, David},
	year         = 2013,
	month        = feb,
	booktitle    = {Proceedings of the 30th {International} {Conference} on {Machine} {Learning}},
	publisher    = {PMLR},
	pages        = {115--123},
	issn         = {1938-7228},
	language     = {en}
}

@misc{liuVeriSureContractAwareMultiAgent2026,
	title        = {Veri-{Sure}: {A} {Contract}-{Aware} {Multi}-{Agent} {Framework} with {Temporal} {Tracing} and {Formal} {Verification} for {Correct} {RTL} {Code} {Generation}},
	shorttitle   = {Veri-{Sure}},
	author       = {Liu, Jiale and Zhou, Taiyu and Jiang, Tianqi},
	year         = 2026,
	month        = jan,
	publisher    = {arXiv},
	doi          = {10.48550/arXiv.2601.19747}
}

@article{liuInsightXAgentLMMBased2026,
	title        = {{InsightX} {Agent}: {An} {LMM}-{Based} {Agentic} {Framework} {With} {Integrated} {Tools} for {Reliable} {X}-{Ray} {NDT} {Analysis}},
	shorttitle   = {{InsightX} {Agent}},
	author       = {Liu, Jiale and Wang, Huan and Zhang, Yue and Luo, Xiaoyu and Hu, Jiaxiang and Liu, Zhiliang and Xie, Min},
	year         = 2026,
	journal      = {IEEE Transactions on Reliability},
	volume       = 75,
	pages        = {1455--1468},
	doi          = {10.1109/TR.2026.3668946},
	issn         = {0018-9529, 1558-1721},
	copyright    = {https://ieeexplore.ieee.org/Xplorehelp/downloads/license-information/IEEE.html}
}

@article{liuAeroGPTLeveragingLargeScale2026,
	title        = {{AeroGPT}: {Leveraging} {Large}-{Scale} {Audio} {Model} for {Aero}-{Engine} {Bearing} {Fault} {Diagnosis}},
	shorttitle   = {{AeroGPT}},
	author       = {Liu, Jiale and Peng, Dandan and Wang, Huan and Liu, Chenyu and Li, Yan-Fu and Xie, Min},
	year         = 2026,
	journal      = {IEEE Transactions on Cybernetics},
	pages        = {1--14},
	doi          = {10.1109/TCYB.2026.3668256},
	issn         = {2168-2267, 2168-2275},
	copyright    = {https://ieeexplore.ieee.org/Xplorehelp/downloads/license-information/IEEE.html}
}

@misc{guLargeLanguageModels2024,
	title        = {Large {Language} {Models} for {Constructing} and {Optimizing} {Machine} {Learning} {Workflows}: {A} {Survey}},
	shorttitle   = {Large {Language} {Models} for {Constructing} and {Optimizing} {Machine} {Learning} {Workflows}},
	author       = {Gu, Yang and You, Hengyu and Cao, Jian and Yu, Muran and Fan, Haoran and Qian, Shiyou},
	year         = 2024,
	month        = dec,
	publisher    = {arXiv},
	doi          = {10.48550/arXiv.2411.10478}
}

@article{yangHyperparameterOptimizationMachine2020,
	title        = {On hyperparameter optimization of machine learning algorithms: {Theory} and practice},
	shorttitle   = {On hyperparameter optimization of machine learning algorithms},
	author       = {Yang, Li and Shami, Abdallah},
	year         = 2020,
	month        = nov,
	journal      = {Neurocomputing},
	volume       = 415,
	pages        = {295--316},
	doi          = {10.1016/j.neucom.2020.07.061},
	issn         = {0925-2312}
}

@article{cuomoScientificMachineLearning2022,
	title        = {Scientific {Machine} {Learning} {Through} {Physics}–{Informed} {Neural} {Networks}: {Where} we are and {What}’s {Next}},
	shorttitle   = {Scientific {Machine} {Learning} {Through} {Physics}–{Informed} {Neural} {Networks}},
	author       = {Cuomo, Salvatore and Di Cola, Vincenzo Schiano and Giampaolo, Fabio and Rozza, Gianluigi and Raissi, Maziar and Piccialli, Francesco},
	year         = 2022,
	month        = jul,
	journal      = {Journal of Scientific Computing},
	volume       = 92,
	number       = 3,
	pages        = 88,
	doi          = {10.1007/s10915-022-01939-z},
	issn         = {1573-7691},
	language     = {en}
}

@article{zhuDigitalTwinSurrogate2026,
	title        = {Digital twin surrogate modeling for real-time monitoring of gear transmissions using a dynamic graph attention network},
	author       = {Zhu, Benran and Chao, Qun and Wang, Zhongrui and Xia, Pengcheng and Liu, Chengliang},
	year         = 2026,
	month        = may,
	journal      = {Advanced Engineering Informatics},
	volume       = 72,
	pages        = 104509,
	doi          = {10.1016/j.aei.2026.104509},
	issn         = {1474-0346}
}

@article{wirthDatadrivenSurrogateMaterial2026,
	title        = {Data-driven surrogate material model for the mechanical simulation of additively manufactured architected weaves},
	author       = {Wirth, M. and Shea, K.},
	year         = 2026,
	month        = sep,
	journal      = {Advanced Engineering Informatics},
	volume       = 74,
	pages        = 104661,
	doi          = {10.1016/j.aei.2026.104661},
	issn         = {1474-0346}
}

@article{xieRapidGenerationMethod2025,
	title        = {Rapid generation method of process routes based on multi-agent collaboration with {LLMs}},
	author       = {Xie, Yanling and Liu, Jihong and Wang, Ruiwen and Wang, Zuoxu and Yu, Kai and Song, Ziming},
	year         = 2025,
	month        = nov,
	journal      = {Advanced Engineering Informatics},
	volume       = 68,
	pages        = 103733,
	doi          = {10.1016/j.aei.2025.103733},
	issn         = {1474-0346}
}

@article{rohit2023tracing,
  title={Tracing the evolution and charting the future of geothermal energy research and development},
  author={Rohit, RV and Kiplangat, Dennis C and Veena, R and Jose, Rajan and Pradeepkumar, AP and Kumar, K Satheesh and others},
  journal={Renewable and Sustainable Energy Reviews},
  volume={184},
  pages={113531},
  year={2023},
  publisher={Elsevier}
}

@article{hellerschmied2024hydrogen,
  title={Hydrogen storage and geo-methanation in a depleted underground hydrocarbon reservoir},
  author={Hellerschmied, Cathrine and Schritter, Johanna and Waldmann, Niels and Zaduryan, Artur B and Rachbauer, Lydia and Scherr, Kerstin E and Andiappan, Anitha and Bauer, Stephan and Pichler, Markus and Loibner, Andreas P},
  journal={Nature Energy},
  volume={9},
  number={3},
  pages={333--344},
  year={2024},
  publisher={Nature Publishing Group UK London}
}

@article{boot2014carbon,
  title={Carbon capture and storage update},
  author={Boot-Handford, Matthew E and Abanades, Juan C and Anthony, Edward J and Blunt, Martin J and Brandani, Stefano and Mac Dowell, Niall and Fern{\'a}ndez, Jos{\'e} R and Ferrari, Maria-Chiara and Gross, Robert and Hallett, Jason P and others},
  journal={Energy \& Environmental Science},
  volume={7},
  number={1},
  pages={130--189},
  year={2014},
  publisher={Royal Society of Chemistry}
}



\end{document}